\documentclass{article}

\PassOptionsToPackage{numbers, compress}{natbib}
\usepackage[final]{neurips_2022}

\usepackage[utf8]{inputenc} %
\usepackage[T1]{fontenc}    %
\usepackage{hyperref}       %
\usepackage{url}            %
\usepackage{booktabs}       %
\usepackage{amsfonts}       %
\usepackage{nicefrac}       %
\usepackage{microtype}      %
\usepackage{xcolor}         %

\usepackage{minitoc}

\usepackage{colortbl}
\usepackage{multirow}
\usepackage{mathcommand}
\usepackage{amsmath,amsthm,amssymb,mathtools}
\usepackage{physics}

\usepackage{stackengine}
\usepackage{rotating}
\usepackage{tabularx}
\usepackage{soul}
\usepackage{algpseudocode, algorithm}
\usepackage{graphicx}
\usepackage{subcaption}

\usepackage{cleveref}

\usepackage{color}
\usepackage{bm}
\usepackage{hyperref}
\usepackage{amsmath,amsfonts}
\usepackage{blindtext}
\usepackage{listings}
\usepackage{mathtools}

\DeclarePairedDelimiterX{\infdivx}[2]{(}{)}{%
	#1\;\delimsize\|\;#2%
}

\newcommand{\kl}{D_{\mathrm{KL}}\infdivx}

\newcommand{\vect}[1]{\bm{#1}}

\newcommand{\x}{\xv}

\newcommand{\dm}{\mathrm{d}}

\newcommand{\E}{\mathbb{E}}
\newcommand{\R}{\mathbb{R}}
\newcommand{\N}{\mathcal{N}}

\newcommand{\dxv}{\mathrm{d}\xv}

\newcommand{\dwv}{\mathrm{d}\wv}
\newcommand{\dt}{\mathrm{d}t}

\newcommand{\deltav}{\vect\delta}
\newcommand{\epsilonv}{\vect\epsilon}

\newcommand{\hv}{\vect h}

\newcommand{\uv}{\vect u}

\newcommand{\wv}{\vect w}
\newcommand{\xv}{\vect x}

\newcommand{\Dv}{\vect D}

\newcommand{\Fv}{\vect F}

\newcommand{\Iv}{\vect I}

\newcommand{\Nv}{\vect N}

\newcommand{\Lc}{\mathcal L}

\newcommand{\Oc}{\mathcal O}

\makeatletter
\newtheorem*{rep@theorem}{\rep@title}
\newcommand{\newreptheorem}[2]{%
\newenvironment{rep#1}[1]{%
 \def\rep@title{#2 \ref{##1}}%
 \begin{rep@theorem}}%
 {\end{rep@theorem}}}
\makeatother

\newreptheorem{proposition}{Proposition}
\newreptheorem{theorem}{Theorem}
\newreptheorem{lemma}{Lemma}

\numberwithin{equation}{section}

\theoremstyle{plain}
\newtheorem{theorem}{Theorem}[section]
\newtheorem{proposition}[theorem]{Proposition}

\theoremstyle{definition}

\newtheorem{assumption}[theorem]{Assumption}

\newenvironment{proof-sketch}{\noindent{\textit{Sketch of proof.}}\hspace*{1em}}{\qed\bigskip}

\title{DPM-Solver: A Fast ODE Solver for Diffusion Probabilistic Model Sampling in Around 10 Steps}

\author{Cheng Lu$^\dag$, Yuhao Zhou$^\dag$, Fan Bao$^\dag$, Jianfei Chen$^{\dag*}$, Chongxuan Li$^\ddag$, Jun Zhu$^\dag$\thanks{Corresponding Author.} \\
$^\dag$Dept. of Comp. Sci. \& Tech., Institute for AI, BNRist Center, THBI Lab \\
$^\dag$Tsinghua-Bosch Joint ML Center, Tsinghua University, Beijing, 100084 China \\
$^\ddag$Gaoling School of Artificial Intelligence, Renmin University of China, \\
$^\ddag$Beijing Key Laboratory of Big Data Management and Analysis Methods, Beijing, China \\
\texttt{\{lucheng.lc15, yuhaoz.cs\}@gmail.com};\quad \texttt{bf19@mails.tsinghua.edu.cn} \\
\texttt{chongxuanli@ruc.edu.cn};\quad \texttt{\{jianfeic, dcszj\}@tsinghua.edu.cn} \\
}

\begin{document}

\maketitle

\begin{abstract}
Diffusion probabilistic models (DPMs) are emerging powerful generative models. Despite their high-quality generation performance, DPMs still suffer from their slow sampling as they generally need hundreds or thousands of sequential function evaluations (steps) of large neural networks to draw a sample. Sampling from DPMs can be viewed alternatively as solving the corresponding diffusion ordinary differential equations (ODEs). In this work, we propose an exact formulation of the solution of diffusion ODEs. The formulation analytically computes the linear part of the solution, rather than leaving all terms to black-box ODE solvers as adopted in previous works. By applying change-of-variable, the solution can be equivalently simplified to an exponentially weighted integral of the neural network. Based on our formulation, we propose \textit{DPM-Solver}, a fast dedicated high-order solver for diffusion ODEs with the convergence order guarantee. DPM-Solver is suitable for both discrete-time and continuous-time DPMs without any further training. Experimental results show that DPM-Solver can generate high-quality samples in only 10 to 20 function evaluations on various datasets. We achieve 4.70 FID in 10 function evaluations and 2.87 FID in 20 function evaluations on the CIFAR10 dataset, and a $4\sim 16\times$ speedup compared with previous state-of-the-art training-free samplers on various datasets.\footnote{Code is available at \url{https://github.com/LuChengTHU/dpm-solver}}

\end{abstract}

\doparttoc
\faketableofcontents

\section{Introduction}
Diffusion probabilistic models (DPMs)~\citep{sohl2015deep,ho2020denoising,song2020score} are emerging powerful generative models with promising performance on many tasks, such as image generation~\citep{dhariwal2021diffusion,meng2021sdedit}, video generation~\citep{ho2022video}, text-to-image generation~\citep{ramesh2022hierarchical}, speech synthesis~\citep{chen2020wavegrad,chen2021wavegrad} and lossless compression~\citep{kingma2021variational}.
DPMs are defined by discrete-time random processes~\citep{sohl2015deep,ho2020denoising} or continuous-time stochastic differential equations (SDEs)~\citep{song2020score}, which learn to gradually remove the noise added to the data points.
Compared with the widely-used generative adversarial networks (GANs)~\citep{goodfellow2014generative} and variational auto-encoders (VAEs)~\citep{kingma2013auto}, DPMs can not only compute exact likelihood~\citep{song2020score}, but also achieve even better sample quality for image generation~\citep{dhariwal2021diffusion}. 
However, to obtain high-quality samples, DPMs usually need hundreds or thousands of sequential steps of large neural network evaluations, thereby resulting in a much slower sampling speed than the single-step GANs or VAEs. Such inefficiency is becoming a critical bottleneck for the adoption of DPMs in downstream tasks, leading to an urgent request to design fast samplers for DPMs.

Existing fast samplers for DPMs can be divided into two categories.
The first category includes knowledge distillation~\citep{salimans2022progressive,luhman2021knowledge} and noise level or sample trajectory learning~\citep{san2021noise,nichol2021improved,lam2021bilateral,watson2021learning}. Such methods require a possibly expensive training stage before they can be used for efficient sampling. Furthermore, their applicability and flexibility might be limited. It might require nontrivial effort to adapt the method to different models, datasets, and number of sampling steps. 
The second category consists of training-free~\citep{song2020denoising,jolicoeur2021gotta,bao2022analytic} samplers, which are suitable for all pre-trained DPMs in a simple plug-and-play manner. 
Training-free samplers include adopting implicit~\citep{song2020denoising} or analytical~\citep{bao2022analytic} generation process, advanced differential equation (DE) solvers~\citep{song2020score,jolicoeur2021gotta,liu2022pseudo,popov2021diffusion,tachibana2021taylor} and dynamic programming~\citep{watson2021learning}.
However, these methods still require $\sim$ 50 function evaluations~\citep{bao2022analytic} to generate high-quality samples (comparable to those generated by plain samplers in about 1000 function evaluations), thereby are still time-consuming.

In this work, we bring the efficiency of training-free samplers to a new level to produce high-quality samples in the ``\textit{few-step sampling}'' regime, where the sampling can be done within around 10 steps of sequential function evaluations. We tackle the alternative problem of sampling from DPMs as solving the corresponding diffusion ordinary differential equations (ODEs) of DPMs, and carefully examine the structure of diffusion ODEs. 
Diffusion ODEs have a semi-linear structure --- they consist of a linear function of the data variable and a  nonlinear function  parameterized by neural networks. 
Such structure is omitted in previous training-free samplers~\cite{song2020score,jolicoeur2021gotta}, which directly use black-box DE solvers. 
To utilize the semi-linear structure, we derive an exact formulation of the solutions of diffusion ODEs by analytically computing the linear part of the solutions, avoiding the corresponding discretization error. 
Furthermore, by applying change-of-variable, the solutions can be equivalently simplified to an exponentially weighted integral of the neural network. Such integral is very special and can be efficiently approximated by the numerical methods for exponential integrators~\citep{hochbruck2010exponential}.

\begin{figure}[t]
	\begin{minipage}{0.18\linewidth}
		\centering
			\includegraphics[width=\linewidth]{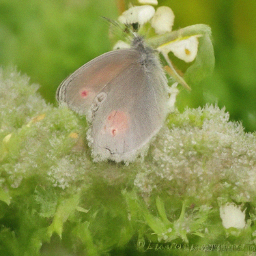}\\
			\includegraphics[width=\linewidth]{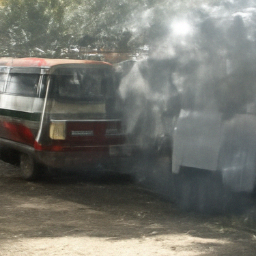}\\
			\small{NFE = 10} \\
	\end{minipage}
	\hspace{-0.15cm}
	\begin{minipage}{0.18\linewidth}
		\centering
			\includegraphics[width=\linewidth]{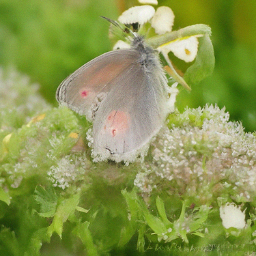}\\
			\includegraphics[width=\linewidth]{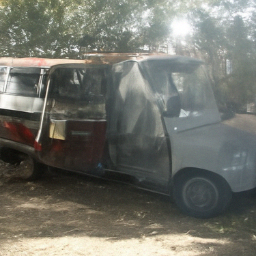}\\
		    \small{NFE = 15} \\
	\end{minipage}
	\hspace{-0.15cm}
	\begin{minipage}{0.18\linewidth}
		\centering
			\includegraphics[width=\linewidth]{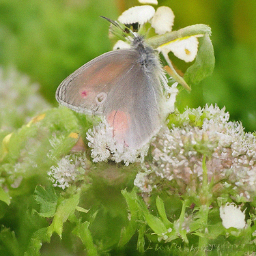}\\
			\includegraphics[width=\linewidth]{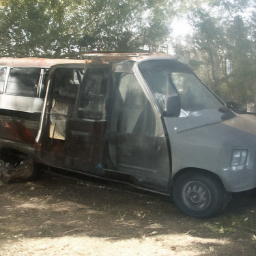}\\
		    \small{NFE = 20} \\
	\end{minipage}
	\hspace{-0.15cm}
	\begin{minipage}{0.18\linewidth}
		\centering
			\includegraphics[width=\linewidth]{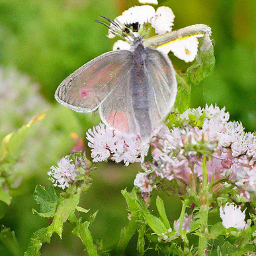}\\
			\includegraphics[width=\linewidth]{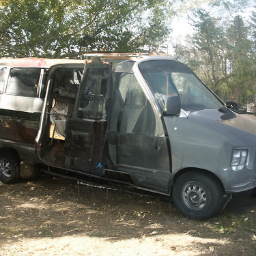}\\
		    \small{NFE = 100} \\
	\end{minipage}
	\hspace{-0.15cm}
	\hspace{0.05\linewidth}
	\begin{minipage}{0.18\linewidth}
		\centering
			\includegraphics[width=\linewidth]{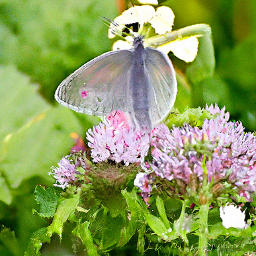}\\
			\includegraphics[width=\linewidth]{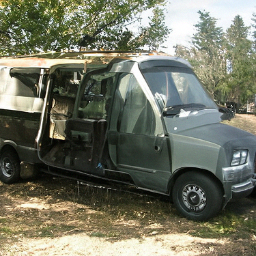}\\
		    \small{NFE = 10} \\
	\end{minipage}
	
	\vspace{0.3cm}
	\begin{minipage}{\linewidth}
    \hspace{0.28\linewidth}
    \small{(a) DDIM~\citep{song2020denoising}}
    \hspace{0.34\linewidth}
    \small{(b) DPM-Solver (ours)}
	\end{minipage}
	
	\caption{\small{Samples by DDIM~\citep{song2020denoising} with 10, 15, 20, 100 number of function evaluations (NFE), and DPM-Solver (ours) with only 10 NFE, using the pre-trained DPMs on ImageNet 256$\times$256 with classifier guidance~\citep{dhariwal2021diffusion}.}
	\label{fig:intro}
}
\vspace{-.2in}
\end{figure}

Based on our formulation of solutions, we propose \textit{DPM-Solver}, a fast dedicated solver for diffusion ODEs by approximating the above integral. Specifically, we propose first-order, second-order and third-order versions of DPM-Solver with convergence order guarantees.
We further propose an adaptive step size schedule for DPM-Solver.
In general, DPM-Solver is applicable to both continuous-time and discrete-time DPMs, and also conditional sampling with classifier guidance~\citep{dhariwal2021diffusion}.
Fig.~\ref{fig:intro} demonstrates the speedup performance of a Denoising Diffusion Implicit Models (DDIM)~\citep{song2020denoising} baseline and DPM-Solver, which shows that DPM-Solver can generate high-quality samples with as few as 10 function evaluations and is much faster than DDIM on the ImageNet 256x256 dataset~\citep{deng2009imagenet}. Our additional experimental results show that DPM-Solver can greatly improve the sampling speed of both discrete-time and continuous-time DPMs, and it can achieve excellent sample quality in around 10 function evaluations, which is much faster than all previous training-free samplers of DPMs.

\section{Diffusion Probabilistic Models}
\label{sec:bg}
We review diffusion probabilistic models and  their associated differential equations in this section.
\subsection{Forward Process and Diffusion SDEs}\label{sec:diffusion-sde}
Assume that we have a $D$-dimensional random variable $\x_0\in\R^{D}$ with an unknown distribution $q_0(\x_0)$. Diffusion Probabilistic Models (DPMs)~\citep{sohl2015deep,ho2020denoising,song2020score,kingma2021variational} define  a \textit{forward process} $\{\x_t\}_{t\in[0,T]}$ with $T>0$ starting with $\x_0$, such that for any $t\in [0,T]$, the distribution of $\x_t$ conditioned on $\x_0$ satisfies
\begin{equation}
\label{eqn:q0t}
    q_{0t}(\x_t|\x_0)=\N(\x_t | \alpha(t)\x_0, \sigma^2(t)\Iv),
\end{equation}
where $\alpha(t),\sigma(t) \in \R^{+}$ are differentiable functions of $t$ with bounded derivatives, and we denote them as $\alpha_t,\sigma_t$ for simplicity. The choice for $\alpha_t$ and $\sigma_t$ is referred to as the \textit{noise schedule} of a DPM.
Let $q_t(\x_t)$ denote the marginal distribution of $\x_t$, DPMs choose noise schedules to ensure that $q_T(\x_T)\approx \N(\x_T|\vect{0},\tilde\sigma^2\Iv)$ for some $\tilde\sigma>0$, and the \textit{signal-to-noise-ratio} (SNR) $\alpha_t^2 / \sigma_t^2$ is strictly decreasing w.r.t. $t$~\citep{kingma2021variational}.  Moreover, \citet{kingma2021variational} prove that the following stochastic differential equation (SDE) has the same transition distribution $q_{0t}(\x_t|\x_0)$ as in Eq.~\eqref{eqn:q0t} for any $t\in[0,T]$:
\begin{equation}
\label{eqn:forward_sde}
    \dxv_t = f(t)\x_t \dt + g(t) \dwv_t, \quad \x_0\sim q_0(\x_0),
\end{equation}
where $\wv_t\in\R^D$ is the standard Wiener process, and
\begin{equation}
\label{eqn:ft_gt}
    f(t)= \frac{\dd \log \alpha_t}{\dt}, \quad g^2(t)=\frac{\dd\sigma_t^2}{\dt} - 2\frac{\dd\log\alpha_t}{\dt}\sigma_t^2.
\end{equation}
Under some regularity conditions, \citet{song2020score} show that the forward process in Eq.~\eqref{eqn:forward_sde} has an equivalent \textit{reverse process} from time $T$ to $0$, starting with the marginal distribution $q_T(\x_T)$:
\begin{equation}
\label{eqn:reverse_sde_true}
    \dxv_t = [f(t)\x_t - g^2(t) \nabla_{\x}\log q_t(\x_t)] \dt + g(t) \dd \bar{\wv}_t, \quad \x_T\sim q_T(\x_T),
\end{equation}
where $\bar{\wv}_t$ is a standard Wiener process in the reverse time. The only unknown term in Eq.~\eqref{eqn:reverse_sde_true} is the \textit{score function} $\nabla_{\x}\log q_t(\x_t)$ at each time $t$. In practice, DPMs use a neural network $\epsilonv_\theta(\x_t,t)$ parameterized by $\theta$ to estimate the scaled score function: $-\sigma_t\nabla_{\x}\log q_t(\x_t)$. The parameter $\theta$ is optimized by minimizing the following objective~\citep{ho2020denoising, song2020score}:
\begin{align}
    \Lc(\theta;\omega(t))&\coloneqq \frac{1}{2}\int_0^T \omega(t) \E_{q_t(\x_t)}\Big[\|\epsilonv_\theta(\x_t,t)+\sigma_t\nabla_{\x}\log q_t(\x_t)\|_2^2\Big]\dt \nonumber \\
    &= \frac{1}{2}\int_0^T \omega(t) \E_{q_0(\x_0)}\E_{q(\epsilonv)}\Big[\|\epsilonv_\theta(\x_t,t)-\epsilonv\|_2^2\Big]\dt + C, \nonumber
\end{align}
where $\omega(t)$ is a weighting function, $\epsilonv\sim q(\epsilonv)=\N(\epsilonv|\vect{0},\Iv)$, $\x_t=\alpha_t\x_0+\sigma_t\epsilonv$, and $C$ is a constant independent of $\theta$. As $\epsilonv_\theta(\x_t,t)$ can also be regarded as predicting the Gaussian noise added to $\x_t$, it is usually called the \textit{noise prediction model}. 
Since the ground truth of $\epsilonv_\theta(\x_t,t)$ is $-\sigma_t\nabla_{\x}\log q_t(\x_t)$, DPMs replace the score function in Eq.~\eqref{eqn:reverse_sde_true} by $-\epsilonv_\theta(\x_t,t) / \sigma_t$ and define a parameterized reverse process (\textit{diffusion SDE}) from time $T$ to $0$, starting with $\x_T\sim \N(\vect{0},\tilde\sigma^2\Iv)$:
\begin{equation}
\label{eqn:reverse_sde_theta}
    \dxv_t = \left [f(t)\x_t + \frac{g^2(t)}{\sigma_t} \epsilonv_\theta(\x_t,t)\right] \dt + g(t) \dd \bar{\wv}_t, \quad \x_T\sim \N(\vect{0},\tilde\sigma^2\Iv).
\end{equation}
Samples can be generated from DPMs by solving the diffusion SDE in Eq.~(\ref{eqn:reverse_sde_theta}) with numerical solvers, which discretize the SDE from $T$ to $0$. \citet{song2020score} proved that the traditional ancestral sampling method for DPMs~\citep{ho2020denoising} can be viewed as a first-order SDE solver for Eq.~\eqref{eqn:reverse_sde_theta}. However, these first-order methods usually need hundreds of or thousands of function evaluations to converge~\citep{song2020score}, leading to extremely slow sampling speed.

\subsection{Diffusion (Probability Flow) ODEs}
\label{sec:bg_ode}
When discretizing SDEs, the step size is limited by the randomness of the Wiener process~\citep[Chap. 11]{kloeden1992Numerical}. A large step size (small number of steps) often causes non-convergence, especially in high dimensional spaces. 
For faster sampling, one can consider the 
associated \textit{probability flow ODE}~\citep{song2020score}, which has the same marginal distribution at each time $t$ as that of the SDE. 
Specifically, for DPMs, \citet{song2020score} proved that the probability flow ODE of Eq.~\eqref{eqn:reverse_sde_true} is
\begin{equation}
    \frac{\dxv_t}{\dt} = f(t)\x_t - \frac{1}{2}g^2(t)\nabla_{\x}\log q_t(\x_t), \quad \x_T\sim q_T(\x_T),
\end{equation}
where the marginal distribution of $\x_t$ is also $q_t(\x_t)$. By replacing the score function with the noise prediction model, \citet{song2020score} defined the following parameterized ODE (\textit{diffusion ODE}):
\begin{equation}
\label{eqn:diffusion_ode}
    \frac{\dxv_t}{\dt} = \hv_\theta(\x_t,t)\coloneqq f(t)\x_t + \frac{g^2(t)}{2\sigma_t}\epsilonv_\theta(\x_t,t), \quad \x_T\sim \N(\vect{0},\tilde\sigma^2\Iv).
\end{equation}
Samples can be drawn by solving the ODE from $T$ to $0$.
Comparing with SDEs, ODEs can be solved with larger step sizes as they have no randomness. Furthermore, we can take advantage of efficient numerical ODE solvers to accelerate the sampling. 
\citet{song2020score} used the RK45 ODE solver~\citep{dormand1980family}  for the diffusion ODEs, which generates samples in $\sim$ 60 function evaluations to reach comparable quality with a 1000-step SDE solver for Eq.~\eqref{eqn:reverse_sde_theta} on the CIFAR-10 dataset~\citep{Krizhevsky09learningmultiple}. However, existing general-purpose ODE solvers still cannot generate satisfactory samples in the few-step ($\sim$ 10 steps) sampling regime. To the best of our knowledge, there is still a lack of training-free samplers for DPMs in the few-step sampling regime, and the sampling speed of DPMs is still a critical issue.

\section{Customized  Fast Solvers for Diffusion ODEs }
\label{sec:solver}
As highlighted in Sec.~\ref{sec:bg_ode}, discretizing SDEs is generally difficult in high dimensions~\citep[Chap. 11]{kloeden1992Numerical} and it is hard to converge within few steps. In contrast, ODEs are easier to solve, yielding a potential for fast samplers.
However, as mentioned in Sec.~\ref{sec:bg_ode}, the general black-box ODE solver used in previous work~\citep{song2020score} empirically fails to converge in few steps. This motivates us to design a dedicated solver for diffusion ODEs to enable fast and high-quality few-step sampling. 
We start with a detailed investigation of the specific structure of diffusion ODEs.

\subsection{Simplified Formulation of Exact Solutions of Diffusion ODEs}
\label{sec:exact_solution}
The key insight of this work is that given an initial value $\x_s$ at time $s>0$, the solution $\x_t$ at each time $t < s$ of diffusion ODEs in Eq.~\eqref{eqn:diffusion_ode} can be simplified into a very special exact formulation which can be efficiently approximated.

Our first key observation is that a part of the solution $\x_t$ can be exactly computed by considering the particular structure of diffusion ODEs.
The r.h.s. of diffusion ODEs in Eq.~\eqref{eqn:diffusion_ode} consists of two parts: the part $f(t)\x_t$ is a linear function of $\x_t$, and the other part $\frac{g^2(t)}{2\sigma_t} \epsilonv_\theta(\x_t,t)$ is generally a nonlinear function of $\x_t$ because of the neural network $\epsilonv_\theta(\x_t,t)$. This type of ODE is referred to as \textit{semi-linear} ODE. The black-box ODE solvers adopted by previous work~\cite{song2020score} are ignorant of this semi-linear structure as they take the whole $\hv_\theta(\x_t,t)$ in Eq.~\eqref{eqn:diffusion_ode} as the input, which causes discretization errors of both the linear and nonlinear term. 
We note that for semi-linear ODEs, the solution at time $t$ can be exactly formulated by the \textit{``variation of constants''} formula~\citep{atkinson2011numerical}:
\begin{equation}
\label{eqn:variation_of_constants}
    \x_t = e^{\int_s^t f(\tau)\dd\tau}\x_s
        + \int_s^t \left(e^{\int_\tau^t f(r)\dd r}\frac{g^2(\tau)}{2\sigma_\tau} \epsilonv_\theta(\x_\tau,\tau)\right)\dm\tau.
\end{equation}
This formulation decouples the linear part and the nonlinear part. In contrast to black-box ODE solvers, the linear part is now exactly computed, which eliminates the approximation error of the linear term. However, the integral of the nonlinear part is still complicated because it couples the coefficients about the noise schedule (i.e., $f(\tau),g(\tau),\sigma_\tau$) and the complex neural network $\epsilonv_\theta$, which is still hard to approximate.

Our second key observation is that the integral of the nonlinear part can be greatly simplified by introducing a special variable. Let $\lambda_t \coloneqq \log(\alpha_t / \sigma_t)$ (one half of the log-SNR), then $\lambda_t$ is a strictly decreasing function of $t$ (due to the definition of DPMs as discussed in Sec.~\ref{sec:diffusion-sde}). We can rewrite $g(t)$ in Eq.~\eqref{eqn:ft_gt} as
\begin{equation}
\begin{aligned}
    g^2(t) = \frac{\dd\sigma_t^2}{\dt} - 2\frac{\dd\log\alpha_t}{\dt}\sigma_t^2
    = 2\sigma_t^2\left(\frac{\dd\log\sigma_t}{\dt} - \frac{\dd\log\alpha_t}{\dt}\right) = -2\sigma_t^2\frac{\dd\lambda_t}{\dt}.
\end{aligned}
\end{equation}
Combining with $f(t)=\dd \log \alpha_t / \dt$ in Eq.~\eqref{eqn:ft_gt}, we can rewrite Eq.~\eqref{eqn:variation_of_constants} as
\begin{align}
    \x_t = \frac{\alpha_t}{\alpha_s}\x_s - \alpha_t \int_s^t \left(\frac{\dd\lambda_\tau}{\dd\tau}\right) \frac{\sigma_\tau}{\alpha_\tau} \epsilonv_\theta(\x_\tau,\tau)\dd\tau.
    \label{eqn:analytical_solution_of_time}
\end{align}
As $\lambda(t)=\lambda_t$ is a strictly decreasing function of $t$, it has an inverse function $t_\lambda(\cdot)$ satisfying $t=t_\lambda(\lambda(t))$. We further change the subscripts of $\x$ and $\epsilonv_\theta$ from $t$ to $\lambda$ and denote $\hat\x_{\lambda}\coloneqq \x_{t_\lambda(\lambda)}$, $\hat\epsilonv_\theta(\hat\x_{\lambda},\lambda)\coloneqq \epsilonv_\theta(\x_{t_\lambda(\lambda)}, t_\lambda(\lambda))$. Rewrite Eq.~\eqref{eqn:analytical_solution_of_time} by \textit{``change-of-variable''} for $\lambda$, then we have:

\begin{proposition}[Exact solution of diffusion ODEs]
\label{proposition:exact_solution}
Given an initial value $\x_s$ at time $s>0$, the solution $\x_t$ at time $t\in[0,s]$ of diffusion ODEs in Eq.~\eqref{eqn:diffusion_ode} is:
\begin{equation}
\label{eqn:analytic_solution}
    \x_t = \frac{\alpha_t}{\alpha_s}\x_s - \alpha_t \int_{\lambda_s}^{\lambda_t} e^{-\lambda} \hat\epsilonv_\theta(\hat\x_\lambda,\lambda)\dd\lambda.
\end{equation}
\end{proposition}
We call the integral $\int e^{-\lambda} \hat\epsilonv_\theta(\hat\x_\lambda,\lambda)\dd\lambda$ the \textit{exponentially weighted integral} of $\hat\epsilonv_\theta$, which is very special and highly related to the \textit{exponential integrators} in the literature of ODE solvers~\citep{hochbruck2010exponential}.
To the best of our knowledge, such formulation has not been revealed in prior work of diffusion models.

Eq.~\eqref{eqn:analytic_solution} provides a new perspective for approximating the solutions of diffusion ODEs.
Specifically, given $\x_s$ at time $s$, According to Eq.~\eqref{eqn:analytic_solution}, approximating the solution at time $t$ is equivalent to directly approximating the exponentially weighted integral of $\hat\epsilonv_\theta$ from $\lambda_s$ to $\lambda_t$, which avoids the error of the linear terms and is well-studied in the literature of exponential integrators~\citep{hochbruck2010exponential,hochbruck2005explicit}. Based on this insight, we propose fast solvers for diffusion ODEs, as detailed in the following sections.

\subsection{High-Order Solvers for Diffusion ODEs}
\label{sec:high-order}
In this section, we propose high-order solvers for diffusion ODEs with convergence order guarantee by leveraging our proposed solution formulation Eq.~\eqref{eqn:analytic_solution}.~
The proposed solvers and analysis are highly motivated by the methods of exponential integrators~\citep{hochbruck2010exponential,hochbruck2005explicit} in the ODE literature.

Specifically, given an initial value $\x_T$ at time $T$ and $M+1$ time steps $\{t_i\}_{i=0}^{M}$ decreasing from $t_0=T$ to $t_{M}=0$. Let $\tilde\x_{t_0}=\x_T$ be the initial value. 
The proposed solvers use $M$ steps to iteratively compute a sequence $\{\tilde\x_{t_i}\}_{i=0}^{M}$
to approximate the true solutions at time steps $\{ t_{i} \}_{i=0}^M$. In particular, the last iterate $\tilde\x_{t_M}$ approximates the true solution at time $0$.

In order to reduce the approximation error between $\tilde\x_{t_{M}}$ and the true solution at time $0$, we need to reduce the approximation error for each $\tilde\x_{t_i}$ at every step~\citep{atkinson2011numerical}. Starting with the previous value $\tilde\x_{t_{i-1}}$ at time $t_{i-1}$, according to Eq.~\eqref{eqn:analytic_solution}, the exact solution $\x_{t_{i-1} \to t_i} $ at time $t_{i}$ is given by
\begin{equation}
\label{eqn:analytic_solution_each_step}
    \x_{t_{i-1} \to t_i}  = \frac{\alpha_{t_i}}{\alpha_{t_{i-1}}} \tilde\x_{t_{i-1}} - \alpha_{t_i} \int_{\lambda_{t_{i-1}}}^{\lambda_{t_i}} e^{-\lambda} \hat\epsilonv_\theta(\hat\x_{\lambda}, \lambda)\dd\lambda.
\end{equation}
Therefore, to compute the value $\tilde\x_{t_i}$ for approximating $\x_{t_{i-1} \to t_i} $, we need to approximate the exponentially weighted integral of $\hat\epsilonv_\theta$ from $\lambda_{t_{i-1}}$ to $\lambda_{t_i}$.
~Denote $h_i\coloneqq \lambda_{t_i} - \lambda_{t_{i-1}}$, and $\hat\epsilonv_\theta^{(n)}(\hat\x_\lambda,\lambda)\coloneqq \frac{\dd^n \hat\epsilonv_\theta(\hat\x_\lambda,\lambda)}{\dd\lambda^n}$ as the $n$-th order total derivative of $\hat\epsilonv_\theta(\hat\x_\lambda, \lambda)$ w.r.t. $\lambda$. For $k\geq 1$, the $(k-1)$-th order Taylor expansion of $\hat\epsilonv_\theta(\hat\x_{\lambda},\lambda)$ w.r.t. $\lambda$ at $\lambda_{t_{i-1}}$ is
\begin{equation*}
    \hat\epsilonv_\theta(\hat\x_\lambda, \lambda) = \sum_{n=0}^{k-1} \frac{(\lambda - \lambda_{t_{i-1}})^n}{n!} \hat\epsilonv_\theta^{(n)}(\hat\x_{\lambda_{t_{i-1}}},\lambda_{t_{i-1}})+\Oc((\lambda - \lambda_{t_{i-1}})^{k}),
\end{equation*}
Substituting the above Taylor expansion into Eq.~\eqref{eqn:analytic_solution_each_step} yields
\begin{equation}
    \label{eqn:k-th-expansion}
    \x_{t_{i-1} \to t_i}\! =\! \frac{\alpha_{t_i}}{\alpha_{t_{i-1}}} \tilde\x_{t_{i-1}} - \alpha_{t_i} \sum_{n=0}^{k-1} \hat\epsilonv_\theta^{(n)}(\hat\x_{\lambda_{t_{i-1}}},\lambda_{t_{i-1}}) \!\int_{\lambda_{t_{i-1}}}^{\lambda_{t_i}}\!\! e^{-\lambda}  \frac{(\lambda - \lambda_{t_{i-1}})^n}{n!} \dd\lambda + \Oc(h_i^{k+1}),
\end{equation}
where the integral $\int e^{-\lambda}  \frac{(\lambda - \lambda_{t_{i-1}})^n}{n!} \dd\lambda$ can be \textbf{analytically} computed by repeatedly applying $n$ times of integration-by-parts (see Appendix~\ref{appendix:general_expansion}). Therefore, to approximate $\x_{t_{i-1} \to t_i}$, we only need to approximate the $n$-th order total derivatives $\hat\epsilonv_\theta^{(n)}(\hat\x_\lambda,\lambda)$ for $n\leq k-1$, which is a well-studied problem in the ODE literature~\citep{hochbruck2005explicit,luan2021efficient}.
By dropping the $\Oc(h_i^{k+1})$  error term and approximating the first $(k-1)$-th total derivatives with the ``stiff order conditions''~\citep{hochbruck2005explicit,luan2021efficient},
we can derive $k$-th-order ODE solvers for diffusion ODEs. We name such solvers as \textit{DPM-Solver} overall, and \textit{DPM-Solver-$k$} for a specific order $k$. %
Here we take $k=1$ for demonstration. In this case, Eq.~\eqref{eqn:k-th-expansion} becomes
\begin{align}
    \x_{t_{i-1} \to t_i} &= \frac{\alpha_{t_i}}{\alpha_{t_{i-1}}} \tilde\x_{t_{i-1}} - \alpha_{t_i} \epsilonv_\theta(\tilde\x_{t_{i-1}},t_{i-1}) \int_{\lambda_{t_{i-1}}}^{\lambda_{t_i}} e^{-\lambda}\dd\lambda + \Oc(h_i^2) \nonumber \\
    &= \frac{\alpha_{t_i}}{\alpha_{t_{i-1}}} \tilde\x_{t_{i-1}} - \sigma_{t_i} (e^{h_i} - 1)\epsilonv_\theta(\tilde\x_{t_{i-1}},t_{i-1}) + \Oc(h_i^2).   \nonumber
\end{align}
By dropping the high-order error term $\Oc(h_i^2)$, we can obtain an approximation for $\x_{t_{i-1} \to t_i} $. As $k=1$ here, we call this solver \textit{DPM-Solver-1}, and the detailed algorithm is as following.

\textbf{DPM-Solver-1.}\quad Given an initial value $\x_T$ and $M+1$ time steps $\{t_i\}_{i=0}^M$ decreasing from $t_0=T$ to $t_M=0$. Starting with $\tilde\x_{t_0}=\x_T$, the sequence $\{\tilde\x_{t_i}\}_{i=1}^M$ is computed iteratively as follows:
\begin{equation}
\label{eqn:1st}
    \tilde\x_{t_i} = \frac{\alpha_{t_i}}{\alpha_{t_{i-1}}} \tilde\x_{t_{i-1}} - \sigma_{t_i} (e^{h_i} - 1)\epsilonv_\theta(\tilde\x_{t_{i-1}},t_{i-1}),~~~~\mbox{where }h_i=\lambda_{t_i} - \lambda_{t_{i-1}}.
\end{equation}

For $k\geq 2$, approximating the first $k$ terms of the Taylor expansion needs additional intermediate points between $t$ and $s$~\citep{hochbruck2005explicit}.
The derivation is more technical so we defer it to Appendix~\ref{appendix:proof}.
Below we propose algorithms for $k=2,3$ and name them as \textit{DPM-Solver-2} and \textit{DPM-Solver-3}, respectively.

\begin{algorithm}[H]
    \centering
    \caption{DPM-Solver-2.}\label{alg:dpm-solver-2}
    \begin{algorithmic}[1]
    \Require initial value $\x_T$, time steps $\{t_i\}_{i=0}^M$, model $\epsilonv_\theta$
        \State $\tilde\x_{t_0}\gets\x_T$
        \For{$i\gets 1$ to $M$}
        \State $s_{i} \gets t_{\lambda}\left(\frac{\lambda_{t_{i-1}} + \lambda_{t_i}}{2}\right)$
        \State 
    $\uv_{i} \gets \frac{\alpha_{s_i}}{\alpha_{t_{i-1}}} \tilde\x_{t_{i-1}} - \sigma_{s_i}\left(e^{\frac{h_i}{2}} - 1\right)\epsilonv_\theta(\tilde \x_{t_{i-1}},t_{i-1})$
    \State  $\tilde\x_{t_{i}} \gets \frac{\alpha_{t_{i}}}{\alpha_{t_{i-1}}} \tilde\x_{t_{i-1}} - \sigma_{t_{i}}\left(e^{h_i} - 1\right)\epsilonv_\theta(\uv_{i},s_i)$
        \EndFor
        \State \Return $\tilde\x_{t_M}$
    \end{algorithmic}
\end{algorithm}
\begin{algorithm}[H]
    \centering
    \caption{DPM-Solver-3.}\label{alg:dpm-solver-3}
    \begin{algorithmic}[1]
    \Require initial value $\x_T$, time steps $\{t_i\}_{i=0}^M$, model $\epsilonv_\theta$
    \State $\tilde\x_{t_0}\gets\x_T$, $r_1\gets\frac{1}{3}$, $r_2\gets\frac{2}{3}$
    \For{$i\gets 1$ to $M$}
\State 
$s_{2i-1} \gets t_\lambda\left(\lambda_{t_{i-1}} + r_1 h_i\right),\quad s_{2i} \gets t_\lambda\left(\lambda_{t_{i-1}} + r_2 h_i\right)$ 
\State 
    $\uv_{2i-1} \gets \frac{\alpha_{s_{2i-1}}}{\alpha_{t_{i-1}}} \tilde\x_{t_{i-1}} - \sigma_{s_{2i-1}}\left(e^{r_1 h_i} - 1\right)\epsilonv_\theta(\tilde \x_{t_{i-1}},t_{i-1})$
\State 
    $\Dv_{2i-1} \gets \epsilonv_\theta(\uv_{2i-1},s_{2i-1}) - \epsilonv_\theta(\tilde\x_{t_{i-1}},t_{i-1})$
\State
    $\uv_{2i} \gets \frac{\alpha_{s_{2i}}}{\alpha_{t_{i-1}}} \tilde\x_{t_{i-1}} 
    - \sigma_{s_{2i}}\left(e^{r_2 h_i} - 1\right)\epsilonv_\theta(\tilde\x_{t_{i-1}},t_{i-1}) - \frac{\sigma_{s_{2i}}r_2}{r_1}\left( \frac{e^{r_2 h_i} - 1}{r_2 h_i} - 1\right)\Dv_{2i-1}$
\State 
    $\Dv_{2i} \gets \epsilonv_\theta(\uv_{2i},s_{2i}) - \epsilonv_\theta(\tilde\x_{t_{i-1}},t_{i-1})$
\State $\tilde\x_{t_{i}} \gets \frac{\alpha_{t_{i}}}{\alpha_{t_{i-1}}} \tilde\x_{t_{i-1}}
        - \sigma_{t_{i}}\left(e^{h_i} - 1\right)\epsilonv_\theta(\tilde\x_{t_{i-1}},t_{i-1}) - \frac{\sigma_{t_{i}}}{r_2}\left( \frac{e^{h_i} - 1}{h} - 1\right)\Dv_{2i}$
        \EndFor
        \State  \Return $\tilde\x_{t_M}$
    \end{algorithmic}
\end{algorithm}

Here, $t_\lambda(\cdot)$ is the inverse function of $\lambda(t)$, which has an analytical formulation for the practical noise schedule used in~\citep{ho2020denoising,nichol2021improved}, as shown in Appendix~\ref{appendix:implementation}. The chosen intermediate points are ($s_i$, $\uv_{i}$) for DPM-Solver-2 and $(s_{2i-1}, \uv_{2i-1})$ and $(s_{2i}, \uv_{2i})$ for DPM-Solver-3. 
As shown in the algorithm, DPM-Solver-$k$ requires $k$ function evaluations per step for $k=1, 2, 3$. Despite the more expensive steps, higher-order solvers ($k=2, 3$) are usually more efficient since they require much fewer steps to converge, due to their higher convergence order. 
We show that DPM-Solver-$k$ is $k$-th-order solver, as stated in the following theorem. The proof is in Appendix~\ref{appendix:proof}.

\begin{theorem}[DPM-Solver-$k$ as a $k$-th-order solver]
\label{thrm:order}
Assume $\epsilonv_\theta(\x_t,t)$ follows the regularity conditions detailed in Appendix~\ref{appendix:assumptions}, then for $k=1,2,3$, DPM-Solver-$k$ is a $k$-th order solver for diffusion ODEs, i.e., for the sequence $\{\tilde\x_{t_i}\}_{i=1}^M$ computed by DPM-Solver-$k$, the approximation error at time $0$ satisfies $\tilde \x_{t_M} - \x_0 =\Oc(h_{\max}^k)$, where $h_{\text{max}}=\max_{1\leq i \leq M}(\lambda_{t_i}-\lambda_{t_{i-1}})$.
\end{theorem}

Finally, solvers with $k\geq 4$ need much more intermediate points as shown by previous work~\citep{hochbruck2005explicit,luan2021efficient} for  exponential integrators. Therefore, we only consider $k$ from $1$ to $3$ in this work, while leaving the solvers with higher $k$ for future study.

\subsection{Step Size Schedule}
\label{sec:adaptive}
The proposed solvers in Sec.~\ref{sec:high-order} need to specify the time steps $\{t_i\}_{i=0}^M$ in advance. We propose two choices of the time step schedule. One choice is handcrafted, which is to uniformly split the interval $[\lambda_T$, $\lambda_0$], i.e.
$\lambda_{t_i} = \lambda_T + \frac{i}{M}(\lambda_0 - \lambda_T)$, $i=0,\dots,M$. 
Note that this is different from previous work~\citep{ho2020denoising,song2020score} which chooses uniform steps for $t_i$.
Empirically, DPM-Solver with uniform time steps $\lambda_{t_i}$ can already generate quite good samples in few steps, where results are listed in Appendix~\ref{appendix:experiment}.
As the other choice, we propose an adaptive step size algorithm, which dynamically adjusts the step size by combining different orders of DPM-Solver. The adaptive algorithm is inspired by~\citep{jolicoeur2021gotta} and we defer its implementation details to Appendix~\ref{appendix:algorithm}.

For few-step sampling, we need to use up all the number of function evaluations (NFE). When the NFE is not divisible by $3$, we firstly apply DPM-Solver-3 as much as possible, and then add a single step of DPM-Solver-1 or DPM-Solver-2 (dependent on the reminder of $K$ divided by $3$), as detailed in Appendix~\ref{appendix:implementation}. In the subsequent experiments, we use such combination of solvers with the uniform step size schedule for NFE $\leq 20$, and otherwise the adaptive step size schedule.

\subsection{Sampling from Discrete-Time DPMs}
\label{sec:discrete-time}
Discrete-time DPMs~\citep{ho2020denoising} train the noise prediction model at $N$ fixed time steps $\{t_n\}_{n=1}^N$, and the noise prediction model is parameterized by $\tilde\epsilonv_\theta(\x_n, n)$ for $n=0,\dots,N-1$, where each $\x_n$ is corresponding to the value at time $t_{n+1}$. We can transform the discrete-time noise prediction model to the continuous version by letting
$\epsilonv_\theta(\x, t)\coloneqq \tilde\epsilonv_\theta(\x, \frac{(N-1)t}{T})$, for all $\x\in\R^d, t\in[0,T]$.
Note that the input time of $\tilde\epsilonv_\theta$ may not be integers, but we find that the noise prediction model can still work well, and we hypothesize that it is because of the smooth time embeddings (e.g., position embeddings~\citep{ho2020denoising}). By such reparameterization, the noise prediction model can adopt the continuous-time steps as input, and thus we can also use DPM-Solver for fast sampling.
\section{Comparison with Existing Fast Sampling Methods}

Here, we discuss the relationship and highlight the difference between DPM-Solver and existing ODE-based fast sampling methods for DPMs. We further briefly discuss the advantage of training-free samplers over those training-based ones. 

\subsection{DDIM as DPM-Solver-1}
\label{sec:ddim}

Denoising Diffusion Implicit Models (DDIM)~\citep{song2020denoising} design a deterministic method for fast sampling from DPMs. For two adjacent time steps $t_{i-1}$ and $t_{i}$, assume that we have a solution $\tilde\x_{t_{i-1}}$ at time $t_{i-1}$, then a single step of DDIM from time $t_{i-1}$ to time $t_i$ is
\begin{equation}
\label{eqn:ddim}
    \tilde\x_{t_i} = \frac{\alpha_{t_i}}{\alpha_{t_{i-1}}}\tilde\x_{t_{i-1}} - \alpha_{t_i}\left(\frac{\sigma_{t_{i-1}}}{\alpha_{t_{i-1}}} - \frac{\sigma_{t_i}}{\alpha_{t_i}}\right)\epsilonv_\theta(\tilde\x_{t_{i-1}},t_{i-1}).
\end{equation}
Although motivated by entirely different perspectives, we show that the updates of DPM-Solver-1 and Denoising Diffusion Implicit Models (DDIM)~\citep{song2020denoising} are identical. 
By the definition of $\lambda$, we have $\frac{\sigma_{t_{i-1}}}{\alpha_{t_{i-1}}}=e^{-\lambda_{t_{i-1}}}$ and $\frac{\sigma_{t_{i}}}{\alpha_{t_{i}}}=e^{-\lambda_{t_{i}}}$. Plugging these and $h_i= \lambda_{t_i} - \lambda_{t_{i-1}}$ to Eq.~\eqref{eqn:ddim}
results in exactly a step of DPM-Solver-1 in Eq.~\eqref{eqn:1st}.
However, the semi-linear ODE formulation of DPM-Solver allows for principled  generalization to higher-order solvers and convergence order analysis.

Recent work~\citep{salimans2022progressive} also show that DDIM is a first-order discretization of diffusion ODEs by differentiating  both sides of Eq.~\eqref{eqn:ddim}. However, they cannot explain the difference between DDIM and the first-order Euler discretization of diffusion ODEs. 
In contrast, by showing that DDIM is a special case of DPM-Solver, we reveal that DDIM makes full use of the semi-linearity of diffusion ODEs, which explains its superiority over traditional Euler methods.

\subsection{Comparison with Traditional Runge-Kutta Methods}

One can obtain a high-order solver by directly applying traditional explicit Runge-Kutta (RK) methods to the diffusion ODE in Eq.~\eqref{eqn:diffusion_ode}.
Specifically, RK methods write the solution of Eq.~\eqref{eqn:diffusion_ode} in the following integral form:
\begin{equation}
    \x_t = \x_s + \int_s^t \hv_\theta(\x_\tau,\tau)\dd\tau = \x_s + \int_s^t \left( f(\tau)\x_{\tau} + \frac{g^2(\tau)}{2\sigma_{\tau}}\epsilonv_\theta(\x_{\tau},\tau) \right) \dd\tau,
\end{equation}
and use some intermediate time steps between $[t, s]$ and combine the evaluations of $\hv_\theta$ at these time steps to approximate the whole integral. 
The approximation error of explicit RK methods depends on $\hv_\theta$, which consists of the error corresponding to both the linear term $f(\tau)\x_\tau$ and the nonlinear noise prediction model $\epsilonv_\theta$. However, the error of the linear term may increase exponentially because the exact solution of the linear term has an exponential coefficient (as shown in Eq.~\eqref{eqn:variation_of_constants}). There are many empirical evidence~\citep{hochbruck2010exponential,hochbruck2005explicit} showing that directly using explicit RK methods for semi-linear ODEs may suffer from unstable numerical issues for large step size. We also demonstrate the empirical difference of the proposed DPM-Solver and the traditional explicit RK methods in Sec.~\ref{sec:exp_continuous}, which shows that DPM-Solver have smaller discretization errors than the RK methods with the same order.

\subsection{Training-based Fast Sampling Methods for DPMs}
Samplers that need extra training or optimization include knowledge distillation~\citep{salimans2022progressive,luhman2021knowledge}, learning the noise level or variance~\citep{san2021noise,nichol2021improved,bao2022estimating}, and learning the noise schedule or sample trajectory~\citep{lam2021bilateral,watson2021learning}. Although the progressive distillation method~\citep{salimans2022progressive} can obtain a fast sampler within 4 steps, it needs further training costs and loses part of the information in the original DPM (e.g., after distillation, the noise prediction model cannot predict the noise (score function) at every time step between $[0,T]$). In contrast, training-free samplers can keep all the information of the original model, and thereby can be directly extended to the conditional sampling by combining the original model and an external classifier~\citep{dhariwal2021diffusion} (e.g. see Appendix~\ref{appendix:implementation} for the conditional sampling with classifier guidance).

Beyond directly designing fast samplers for DPMs, several works also propose novel types of DPMs which supports faster sampling. For instance, defining a low-dimensional latent variable for DPMs~\citep{vahdat2021score}; designing special diffusion processes with bounded score functions~\citep{dockhorn2021score}; combining GANs with the reverse process of DPMs~\citep{xiao2021tackling}. The proposed DPM-Solver may also be suitable for accelerating the sampling of these DPMs, and we leave them for future work.

\begin{table}[t]
    \centering
    \caption{\small{FID $\downarrow$ on CIFAR-10 for different orders of Runge-Kutta (RK) methods and DPM-Solvers, varying the number of function evaluations (NFE). For RK methods, we evaluate diffusion ODEs w.r.t. both $t$ (Eq.~\eqref{eqn:diffusion_ode}) and $\lambda$ (Eq.~\eqref{eqn:diffusion_ode_logSNR}). We use uniform step size in $t$ for RK ($t$), and uniform step size in $\lambda$ for RK ($\lambda$) and DPM-Solvers.}}
    \label{tab:RK_compare}
        \small{
    \begin{tabular}{lrrrrrrr}
    \toprule
      Sampling method $\backslash$ NFE & 12 & 18 & 24 & 30 & 36 & 42 & 48\\
    \midrule
      RK2 ($t$) & 16.40 & 7.25 & 3.90 & 3.63 & 3.58 & 3.59 & 3.54 \\
      RK2 ($\lambda$) & 107.81 & 42.04 & 17.71 & 7.65 & 4.62 & 3.58 & 3.17 \\
      DPM-Solver-2 & \textbf{5.28} & \textbf{3.43} & \textbf{3.02} & \textbf{2.85} & \textbf{2.78} & \textbf{2.72} & \textbf{2.69} \\
    \arrayrulecolor{black!30}\midrule
      RK3 ($t$) & 48.75 & 21.86 & 10.90 & 6.96 & 5.22 & 4.56 & 4.12 \\
      RK3 ($\lambda$) & 34.29 & 4.90 & 3.50 & 3.03 & 2.85 & 2.74 & 2.69 \\
      DPM-Solver-3 & \textbf{6.03} & \textbf{2.90} & \textbf{2.75} & \textbf{2.70} & \textbf{2.67} & \textbf{2.65} & \textbf{2.65} \\
    \arrayrulecolor{black}\bottomrule
    \end{tabular}
    }
\end{table}

\begin{figure}[t]
\centering
	\begin{subfigure}{0.32\linewidth}
		\centering
			\includegraphics[width=\linewidth]{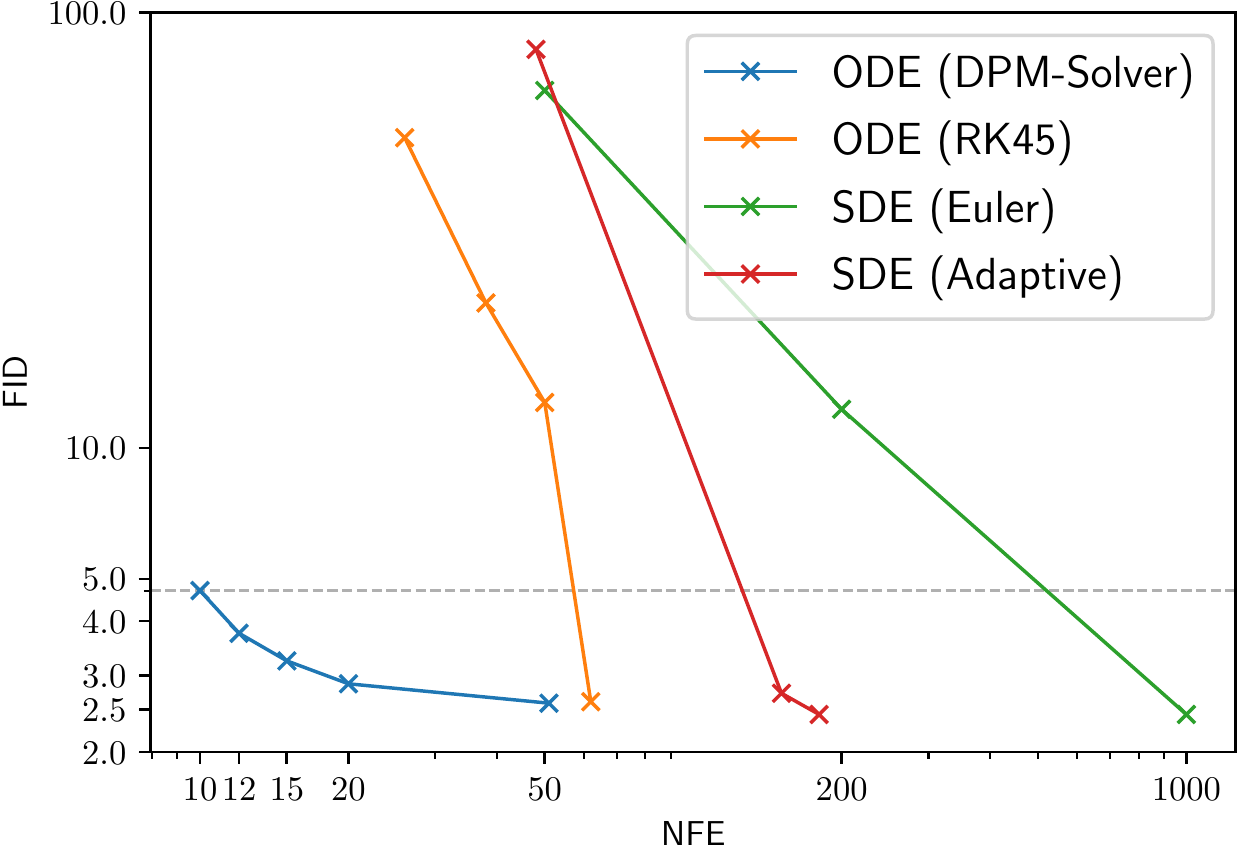}
			\caption{\label{fig:cifar10_fid_continuous}CIFAR-10 (continuous)}
	\end{subfigure}
	\begin{subfigure}{0.32\linewidth}
		\centering
			\includegraphics[width=\linewidth]{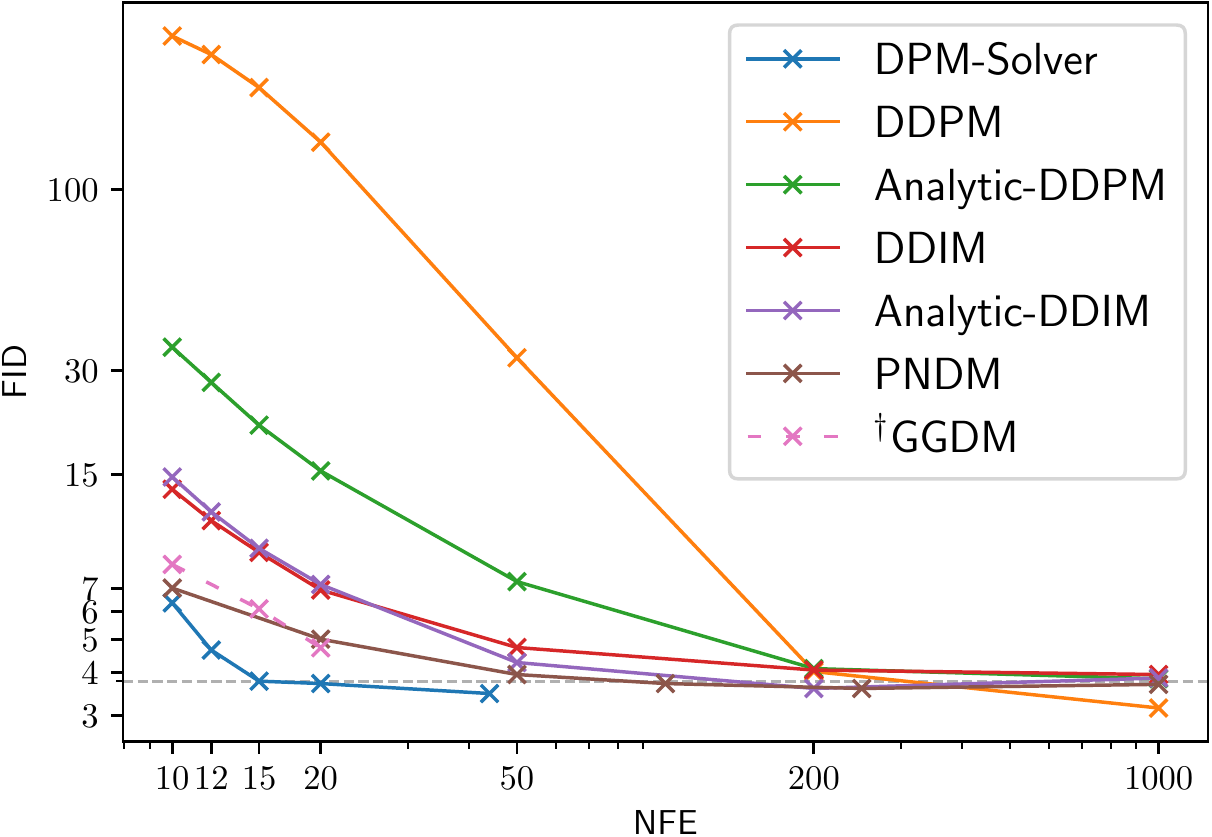}
			\caption{CIFAR-10 (discrete)}
	\end{subfigure}
	\begin{subfigure}{0.32\linewidth}
		\centering
			\includegraphics[width=\linewidth]{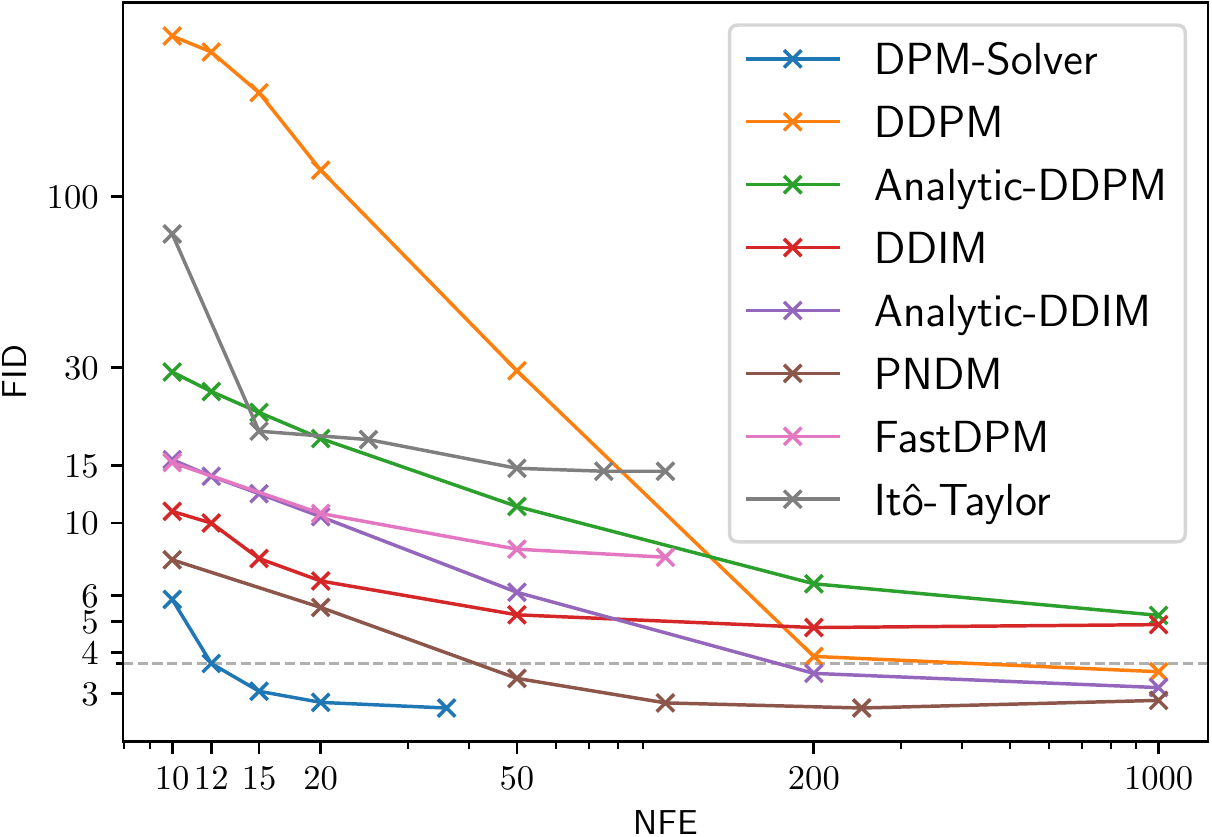}
			\caption{CelebA 64x64 (discrete)} 
	\end{subfigure} \\
	
	\begin{subfigure}{0.32\linewidth}
		\centering
			\includegraphics[width=\linewidth]{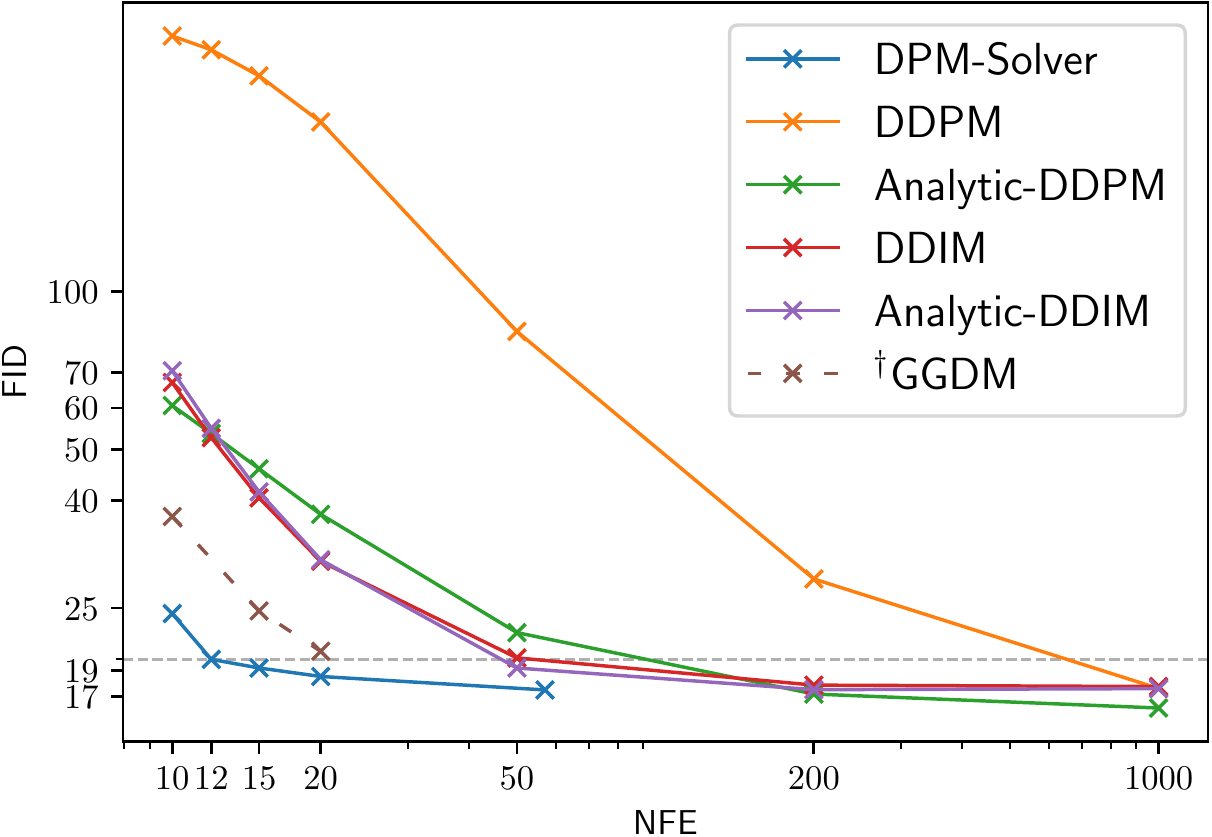}
			\caption{ImageNet 64x64 (discrete)} 
	\end{subfigure}
	\begin{subfigure}{0.32\linewidth}
		\centering
			\includegraphics[width=\linewidth]{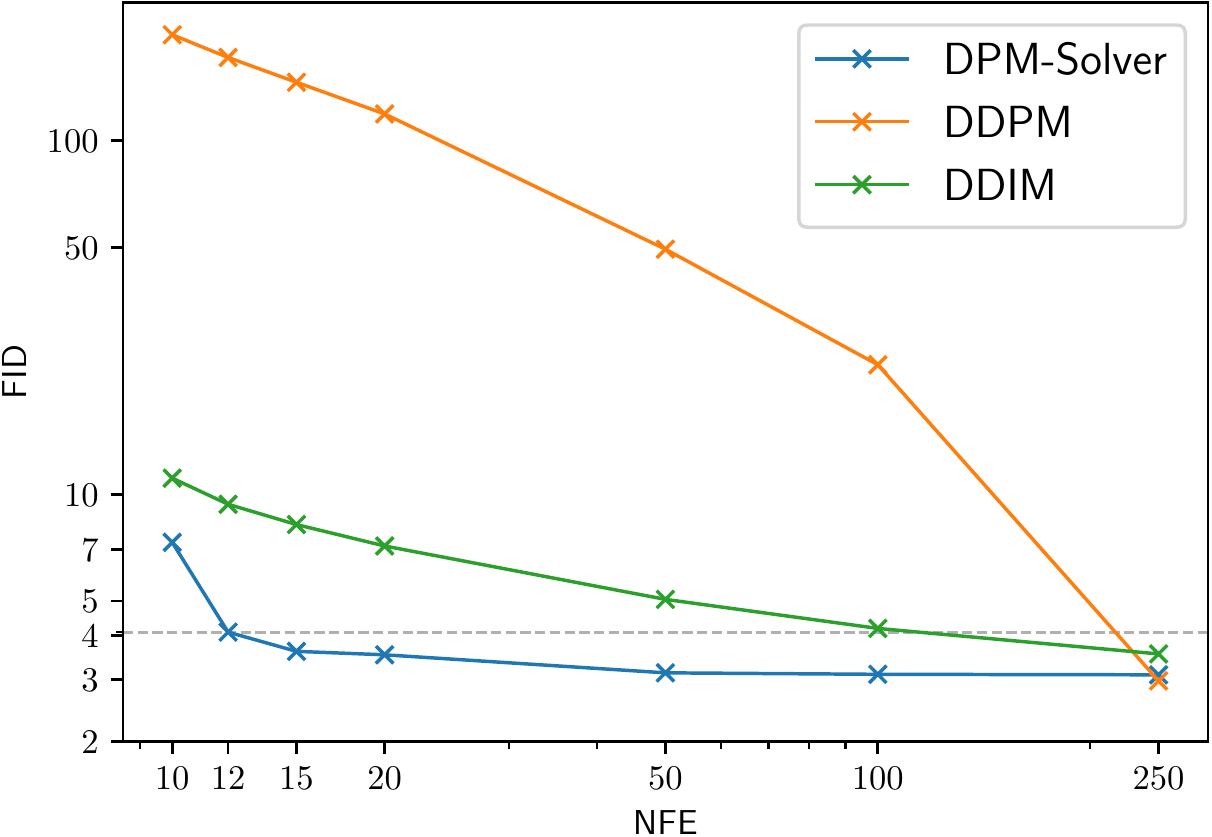}
			\caption{ImageNet 128x128 (discrete)} 
	\end{subfigure}
	\begin{subfigure}{0.32\linewidth}
		\centering
			\includegraphics[width=\linewidth]{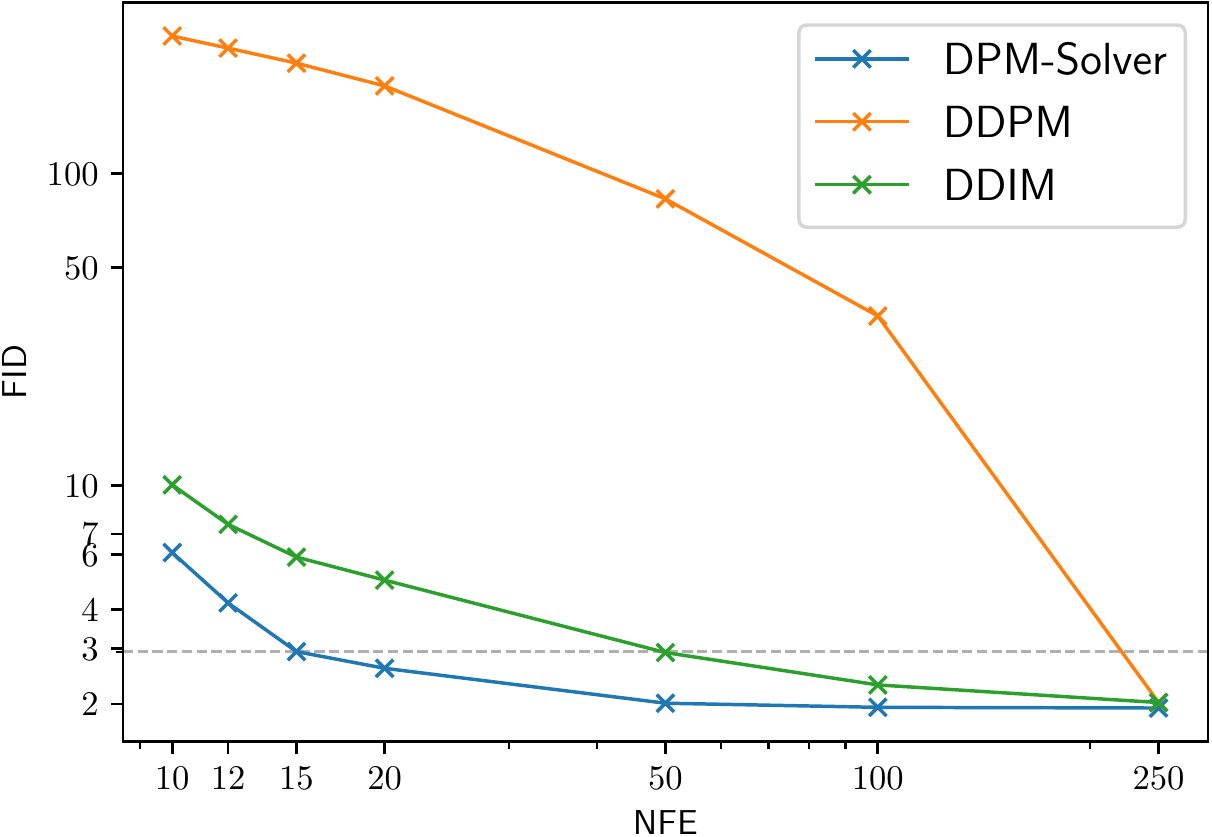}
			\caption{LSUN bedroom 256x256 (discrete)} 
	\end{subfigure}
	\caption{\label{fig:fid}\small{Sample quality measured by FID $\downarrow$ of different sampling methods for DPMs on CIFAR-10 with both continuous-time and discrete-time models, CelebA 64x64, ImageNet 64x64, ImageNet 128x128 and LSUN bedroom 256x256 with discrete-time models, varying the number of function evaluations (NFE). The method $^\dagger$GGDM~\citep{watson2021learning} needs extra training to optimize the sample trajectory, while other methods are training-free. To get the strongest baseline, we use the quadratic step size for DDIM on CelebA, which has a better FID than that of the uniform step size in the original paper~\citep{song2020denoising}.}
}
\vspace{-.2in}
\end{figure}

\section{Experiments}
\label{sec:experiments}
In this section, we show that as a training-free sampler, DPM-Solver can greatly speedup the sampling of existing pre-trained DPMs, including both continuous-time and discrete-time ones, with both linear noise schedule~\citep{ho2020denoising,song2020denoising} and cosine noise schedule~\citep{nichol2021improved}. We vary different number of function evaluations (NFE) which is the number of calls to the noise prediction model $\epsilonv_\theta(\x_t,t)$, and compare the sample quality between DPM-Solver and other methods. For each experiment, We draw 50K samples and use the widely adopted FID score~\citep{heusel2017gans} to evaluate the sample quality, where lower FID usually implies better sample quality.

Unless explicitly mentioned, we always use the solver combination with the uniform step size schedule in Sec.~\ref{sec:adaptive} if the NFE budget is less than 20, and otherwise the DPM-Solver-3 with the adaptive step size schedule in Sec.~\ref{sec:adaptive}. We refer to Appendix~\ref{appendix:implementation} for other implementation details of DPM-Solver and Appendix~\ref{appendix:experiment} for detailed settings.

\subsection{Comparison with Continuous-Time Sampling Methods}
\label{sec:exp_continuous}
We firstly compare DPM-Solver with other continuous-time sampling methods for DPMs. The compared methods include the Euler-Maruyama discretization for diffusion SDEs~\citep{song2020score}, the adaptive step size solver for diffusion SDEs~\citep{jolicoeur2021gotta} and the RK methods for diffusion ODEs~\citep{song2020score,dormand1980family} in Eq.~\eqref{eqn:diffusion_ode}. We compare these methods for sampling from a pre-trained continuous-time ``VP deep'' model~\citep{song2020score} on the CIFAR-10 dataset~\citep{Krizhevsky09learningmultiple} with the linear noise schedule.

Fig.~\ref{fig:cifar10_fid_continuous} shows the efficiency of compared solvers. We use uniform time steps with 50, 200, 1000 NFE for the diffusion SDE with Euler discretization, and vary the tolerance hyperparameter~\citep{song2020score,jolicoeur2021gotta} for the adaptive step size SDE solver~\citep{jolicoeur2021gotta} and RK45 ODE solver~\citep{dormand1980family} to control the NFE. DPM-Solver can generate good sample quality within around 10 NFE, while other solvers have large discretization error even in 50 NFE, which shows that DPM-Solver can achieve $\sim$5 speedup of the previous best solver. In particular, we achieve 4.70 FID with 10 NFE, 3.75 FID with 12 NFE, 3.24 FID with 15 NFE, and 2.87 FID with 20 NFE, which is the fastest sampler on CIFAR-10.

As an ablation study, 
we also compare the second-order and third-order DPM-Solver and RK methods, as shown in Table~\ref{tab:RK_compare}. We compare RK methods for diffusion ODEs w.r.t. both time $t$ in Eq.~\eqref{eqn:diffusion_ode} and half-log-SNR $\lambda$ by applying change-of-variable (see detailed formulations in Appendix~\ref{appendix:RK}). The results show that given the same NFE, the sample quality of DPM-Solver is consistently better than RK methods with the same order. The superior efficiency of DPM-Solver is particularly evident in the few-step regime under 15 NFE, where RK methods have rather large discretization errors. 
This is mainly because DPM-Solver analytically computes the linear term, avoiding the corresponding discretization error. 
Besides, the higher order DPM-Solver-3 converges faster than DPM-Solver-2, which matches  the order analysis in Theorem~\ref{thrm:order}.

\subsection{Comparison with Discrete-Time Sampling Methods}
We use the method in Sec.~\ref{sec:discrete-time} for using DPM-Solver in discrete-time DPMs, and then compare DPM-Solver with other discrete-time training-free samplers, including DDPM~\citep{ho2020denoising}, DDIM~\citep{song2020denoising}, Analytic-DDPM~\citep{bao2022analytic}, Analytic-DDIM~\citep{bao2022analytic}, PNDM~\citep{liu2022pseudo}, FastDPM~\citep{kong2021fast} and It\^o-{T}aylor~\citep{tachibana2021taylor}. We also compare with GGDM~\citep{watson2021learning}, which uses the same pre-trained model but needs further training for the sampling trajectory.
We compare the sample quality by varying NFE from 10 to 1000.

Specifically, we use the discrete-time model trained by $L_{\text{simple}}$ in~\citep{ho2020denoising} on the CIFAR-10 dataset with linear noise schedule; the discrete-time model in~\citep{song2020denoising} on CelebA 64x64~\citep{liu2015deep} with linear noise schedule; the discrete-time model trained by $L_{\text{hybrid}}$ in~\citep{nichol2021improved} on ImageNet 64x64~\citep{deng2009imagenet} with cosine noise schedule; the discrete-time model with classifier guidance in~\citep{dhariwal2021diffusion} on ImageNet 128x128~\citep{deng2009imagenet} with linear noise schedule; the discrete-time model in~\citep{dhariwal2021diffusion} on LSUN bedroom 256x256~\citep{yu2015lsun} with linear noise schedule. For the models trained on ImageNet, we only use their ``mean'' model and omit the ``variance'' model. As shown in Fig.~\ref{fig:fid}, 
on all  datasets, DPM-Solver can obtain reasonable samples within 12 steps (FID 4.65 on CIFAR-10, FID 3.71 on CelebA 64x64 and FID 19.97 on ImageNet 64x64, FID 4.08 on ImageNet 128x128), which is $4\sim 16 \times$ faster than the previous fastest training-free sampler.
DPM-Solver even outperforms GGDM, which requires additional training.

\section{Conclusions}
\label{sec:conclusion}
We tackle the problem of fast and training-free sampling from DPMs. 
We propose DPM-Solver, a fast dedicated training-free solver of diffusion ODEs for fast sampling of DPMs in around 10 steps of function evaluations. DPM-Solver leverages the semi-linearity of diffusion ODEs and it directly approximates a simplified formulation of exact solutions of diffusion ODEs, which consists of an exponentially weighted integral of the noise prediction model. Inspired by numerical methods for exponential integrators, we propose first-order, second-order and third-order DPM-Solver to approximate the exponentially weighted integral of noise prediction models with theoretical convergence guarantee. We propose both handcrafted and adaptive step size schedule, and apply DPM-Solver for both continuous-time and discrete-time DPMs. Our experimental results show that DPM-Solver can generate high-quality samples in around 10 function evaluations on various datasets, and it can achieve $4\sim 16\times$ speedup  compared with previous state-of-the-art training-free samplers.

\textbf{Limitations and broader impact}\quad Despite the promising speedup performance, DPM-Solver is designed for fast sampling, which may be not suitable for accelerating the likelihood evaluations of DPMs. Besides, compared to the commonly-used GANs, diffusion models with DPM-Solver are still not fast enough for real-time applications. In addition, like other deep generative models, DPMs may be used to generate adverse fake contents, and the proposed solver may further amplify the potential undesirable influence of deep generative models for malicious applications.

\section*{Acknowledgements}

This work was supported by National Key Research and Development Project of China (No. 2021ZD0110502); NSF of China Projects (Nos. 62061136001, 61620106010, 62076145, U19B2034, U1811461, U19A2081, 6197222, 62106120); Beijing NSF Project (No. JQ19016); Beijing Outstanding Young Scientist Program NO. BJJWZYJH012019100020098; a grant from Tsinghua Institute for Guo Qiang; the NVIDIA NVAIL Program with GPU/DGX Acceleration; the High Performance Computing Center, Tsinghua University; the Fundamental Research Funds for the Central Universities, and the Research Funds of Renmin University of China (22XNKJ13). J.Z is also supported by the XPlorer Prize.

\bibliography{ref}
\bibliographystyle{IEEEtranN}

\section*{Checklist}

\begin{enumerate}
\item For all authors...
\begin{enumerate}
  \item Do the main claims made in the abstract and introduction accurately reflect the paper's contributions and scope?
    \answerYes{}
  \item Did you describe the limitations of your work?
    \answerYes{See section~\ref{sec:conclusion}.}
  \item Did you discuss any potential negative societal impacts of your work?
    \answerYes{See section~\ref{sec:conclusion}.}
  \item Have you read the ethics review guidelines and ensured that your paper conforms to them?
    \answerYes{}
\end{enumerate}

\item If you are including theoretical results...
\begin{enumerate}
  \item Did you state the full set of assumptions of all theoretical results?
    \answerYes{See Appendix~\ref{appendix:proof}.}
        \item Did you include complete proofs of all theoretical results?
    \answerYes{See Appendix~\ref{appendix:proof}.}
\end{enumerate}

\item If you ran experiments...
\begin{enumerate}
  \item Did you include the code, data, and instructions needed to reproduce the main experimental results (either in the supplemental material or as a URL)?
    \answerYes{Code is attached in the supplemental materials, with the appendix.}
  \item Did you specify all the training details (e.g., data splits, hyperparameters, how they were chosen)?
    \answerYes{Our method is training-free. But we also report the hyperparameters for evaluations used in our proposed solver.}
        \item Did you report error bars (e.g., with respect to the random seed after running experiments multiple times)?
    \answerNo{We observe that the standard deviation of the FID evaluations of DPM-Solver are rather small (mainly less than 0.01) because the FID is already averaged over 50K samples, following existing work~\citep{watson2021learning,jolicoeur2021gotta,bao2022analytic}. The small standard deviation does not change the conclusion. }
        \item Did you include the total amount of compute and the type of resources used (e.g., type of GPUs, internal cluster, or cloud provider)?
    \answerYes{The GPU type and amount is detailed in Appendix~\ref{appendix:experiment}.}
\end{enumerate}

\item If you are using existing assets (e.g., code, data, models) or curating/releasing new assets...
\begin{enumerate}
  \item If your work uses existing assets, did you cite the creators?
    \answerYes{}
  \item Did you mention the license of the assets?
    \answerYes{See Appendix~\ref{appendix:experiment}}
  \item Did you include any new assets either in the supplemental material or as a URL?
    \answerYes{We include our code in the supplemental materials.}
  \item Did you discuss whether and how consent was obtained from people whose data you're using/curating?
    \answerNo{All of the datasets used in the experiments are publicly available.}
  \item Did you discuss whether the data you are using/curating contains personally identifiable information or offensive content?
    \answerYes{We mentioned the human privacy issues of the ImageNet dataset in Appendix~\ref{appendix:experiment}.}
\end{enumerate}

\item If you used crowdsourcing or conducted research with human subjects...
\begin{enumerate}
  \item Did you include the full text of instructions given to participants and screenshots, if applicable?
    \answerNA{}
  \item Did you describe any potential participant risks, with links to Institutional Review Board (IRB) approvals, if applicable?
    \answerNA{}
  \item Did you include the estimated hourly wage paid to participants and the total amount spent on participant compensation?
    \answerNA{}
\end{enumerate}

\end{enumerate}

\newpage
\appendix

\section{Sampling with Invariance to the Noise Schedule}
\label{appendix:discuss}
\begin{table}
    \centering
    \caption{Formulations that are invariant to the choice of the noise schedules. The maximum likelihood training loss w.r.t. $\lambda$ is equivalent to the objectives in~\citep{kingma2021variational,song2021maximum}, and the exact solution of the diffusion ODEs are proposed in Proposition~\ref{proposition:exact_solution}.}
    \vskip 0.10in
    \begin{tabular}{c|c}
    \toprule
    Method & Invariance Formulation\\
    \midrule
    Maximum likelihood training & \(\displaystyle \int_{\lambda_T}^{\lambda_0} \E_{q_0(\x_0)}\E_{\epsilonv\sim\N(\vect{0},\Iv)}\Big[\|\epsilonv_\theta(\hat\x_{\lambda},\lambda)-\epsilonv\|_2^2\Big]\dd\lambda \) \\
    Sampling by diffusion ODEs & \(\displaystyle \hat\x_{\lambda_t} = \frac{\hat\alpha_{\lambda_t}}{\hat\alpha_{\lambda_s}}\hat\x_{\lambda_s} - \hat\alpha_{\lambda_t} \int_{\lambda_s}^{\lambda_t} e^{-\lambda} \hat\epsilonv_\theta(\hat\x_\lambda,\lambda)\dd\lambda \) \\
    \bottomrule
    \end{tabular}
    \label{tab:invariance}
\end{table}

In this section, we discuss more about the exact solution in Proposition~\ref{proposition:exact_solution} and give some insights about the formulation. Below we firstly restate the proposition w.r.t. $\lambda$ (i.e. the half-logSNR).
\begin{repproposition}{proposition:exact_solution}[Exact solution of diffusion ODEs]
Given an initial value $\hat\x_{\lambda_s}$ at time $s$ with the corresponding half-logSNR $\lambda_s$, the solution $\hat\x_{\lambda_t}$ at time $t$ of diffusion ODEs in Eq.~\eqref{eqn:diffusion_ode} with the corresponding half-logSNR $\lambda_t$ is:
\begin{equation}
    \hat\x_{\lambda_t} = \frac{\alpha_t}{\alpha_s}\hat\x_{\lambda_s} - \alpha_t \int_{\lambda_s}^{\lambda_t} e^{-\lambda} \hat\epsilonv_\theta(\hat\x_\lambda,\lambda)\dd\lambda.
\end{equation}
\end{repproposition}
In the following subsections, we will show that such formulation decouples the model $\epsilonv_\theta$ from the specific noise schedule, and thus is invariant to the noise schedule. Moreover, such \textit{change-of-variable} for $\lambda$ in Proposition~\ref{proposition:exact_solution} is highly related to the maximum likelihood training of diffusion models~\citep{kingma2021variational,song2021maximum}. We show that both the maximum likelihood training and the sampling of diffusion models have invariance formulations that are independent of the noise schedule.

\subsection{Decoupling the Sampling Solution from the Noise Schedule}
\label{sec:decoupling}
In this section, we show that Proposition~\ref{proposition:exact_solution} can decouples the exact solutions of the diffusion ODEs from the specific noise schedules (i.e. choice of the functions $\alpha_t=\alpha(t)$ and $\sigma_t=\sigma(t)$). Namely, given a starting point $\lambda_s$, a ending point $\lambda_t$, an initial value $\hat\x_{\lambda_s}$ at $\lambda_s$ and a noise prediction model $\hat\epsilonv_\theta$, \textbf{the solution of $\hat\x_{\lambda_t}$ is invariant of the noise schedule between $\lambda_s$ and $\lambda_t$}.

We firstly consider the VP type diffusion models, which is equivalent to the original DDPM~\citep{ho2020denoising,song2020score}. For VP type diffusion models, we always have $\alpha_t^2+\sigma_t^2=1$, so defining the noise schedule is equivalent to defining the function $\alpha_t=\alpha(t)$ (For example, DDPM~\citep{ho2020denoising} uses a noise schedule such that $\beta(t)=\frac{\dd\log\alpha_t}{\dd t}$ is a linear function of $t$, and i-DDPM~\citep{nichol2021improved} uses a noise schedule such that $\beta(t)=\frac{\dd\log\alpha_t}{\dd t}$ is a cosine function of $t$). As $\lambda_t=\log\alpha_t-\log\sigma_t$, we have $\alpha_t = \sqrt{\frac{1}{1+e^{-2\lambda_t}}}$ and $\sigma_t=\sqrt{\frac{1}{1+e^{2\lambda_t}}}$. Thus, we can directly compute the $\alpha_t$ and $\sigma_t$ for a given $\lambda_t$. Denote $\hat\alpha_{\lambda}\coloneqq \sqrt{\frac{1}{1+e^{-2\lambda}}}$, we have
\begin{equation}
    \hat\x_{\lambda_t} = \frac{\hat\alpha_{\lambda_t}}{\hat\alpha_{\lambda_s}}\hat\x_{\lambda_s} - \hat\alpha_{\lambda_t} \int_{\lambda_s}^{\lambda_t} e^{-\lambda} \hat\epsilonv_\theta(\hat\x_\lambda,\lambda)\dd\lambda.
\end{equation}
We should notice that the integrand $e^{-\lambda}\hat\epsilonv_\theta(\hat\x_{\lambda},\lambda)$ is a function of $\lambda$, so its integral from $\lambda_s$ to $\lambda_t$ is only dependent on the starting point $\lambda_s$, the ending point $\lambda_t$ and the function $\hat\epsilonv_\theta$, which is independent of the intermediate values. As other coefficients ($\hat\alpha_{\lambda_s}$ and $\hat\alpha_{\lambda_t}$) are also only dependent on the starting point $\lambda_s$ and the ending point $\lambda_t$, we can conclude that $\hat\x_{\lambda_t}$ is invariant of the specific choice of the noise schedules. Intuitively, this is because we converts the original integral of time $t$ in Eq.~\eqref{eqn:variation_of_constants} to the integral of $\lambda$, and the functions $f(t)$ and $g(t)$ are converted to an \textbf{analytical} formulation $e^{-\lambda}$, which is invariant to the specific choices of $f(t)$ and $g(t)$. Finally, for other types of diffusion models (such as the VE type and the subVP type), they are all equivalent to the VP type by equivalently rescaling the noise prediction models, as proved in~\citep{kingma2021variational}. Therefore, the solutions of these types also have such property.

In summary, Proposition~\ref{proposition:exact_solution} decouples the solution of diffusion ODEs from the noise schedules, which gives us an opportunity to design tailor-made samplers for DPMs. In fact, as shown in Sec.~\ref{sec:high-order}, the only approximation of the proposed DPM-Solver is about the Taylor expansion of the neural network $\hat\epsilonv_\theta$ w.r.t. $\lambda$, and DPM-Solver \textbf{analytically} computes other coefficients (which are corresponding to the specific noise schedules). Intuitively, DPM-Solver keeps the known information as much as possible, and only approximates the intractable integral of the neural network, so it can generate comparable samples within much fewer steps.

\subsection{Choosing Time Steps for $\lambda$ is Invariant to the Noise Schedule}
As mentioned in Appendix~\ref{sec:decoupling}, the formulation of Proposition~\ref{proposition:exact_solution} decouples the sampling solution from the noise schedule. The solution depends on the starting point $\lambda_s$ and the ending point $\lambda_t$, and is invariant to the intermediate noise schedule. Similarly, the updating equations of the algorithm of DPM-Solver are also invariant to the intermediate noise schedule. Therefore, if we have chosen the time steps $\{\lambda_i\}_{i=0}^M$, then the solution of DPM-Solver is also determined and is invariant to the intermediate noise schedule.

A simple way for choosing time steps for $\lambda$ is uniformly splitting $[\lambda_T, \lambda_{\epsilon}]$, which is the setting in our experiments. However, we believe that there exists more precise ways for choosing the time steps, and we leave it for future work.

\subsection{Relationship with the Maximum Likelihood Training of Diffusion Models}
Interestingly, the maximum likelihood training of diffusion SDEs in continuous time also has such invariance property~\citep{kingma2021variational}. Below we briefly review the maximum likelihood training loss of diffusion SDEs, and then propose a new insight for understanding diffusion models.

Denote the data distribution as $q_0(\x_0)$, the distribution of the forward process at each time $t$ as $q_t(\x_t)$, the distribution of the reverse process at each time $t$ as $p_t(\x_t)$ with $p_T=\N(\vect{0},\Iv)$. In~\citep{song2020score}, it is proved that the KL-divergence between $q_0$ and $p_0$ can be bounded by a weighted score matching loss:
\begin{equation}
    \kl{q_0}{p_0}\leq \kl{q_T}{p_T} + \frac{1}{2}\int_0^T \frac{g^2(t)}{\sigma_t^2}\E_{q_0(\x_0)}\E_{\epsilonv\sim\N(\vect{0},\Iv)}\Big[\|\epsilonv_\theta(\x_t,t)-\epsilonv\|_2^2\Big]\dt + C,
\end{equation}
where $\x_t=\alpha_t\x_0+\sigma_t\epsilonv$ and $C$ is a constant independent of $\theta$. As shown in Sec.~\ref{sec:exact_solution}, we have
\begin{equation}
\begin{aligned}
    g^2(t) = \frac{\dd\sigma_t^2}{\dt} - 2\frac{\dd\log\alpha_t}{\dt}\sigma_t^2
    = 2\sigma_t^2\left(\frac{\dd\log\sigma_t}{\dt} - \frac{\dd\log\alpha_t}{\dt}\right) = -2\sigma_t^2\frac{\dd\lambda_t}{\dt},
\end{aligned}
\end{equation}
so by applying \textit{change-of-variable} w.r.t. $\lambda$, we have
\begin{equation}
    \kl{q_0}{p_0}\leq \kl{q_T}{p_T} + \int_{\lambda_T}^{\lambda_0} \E_{q_0(\x_0)}\E_{\epsilonv\sim\N(\vect{0},\Iv)}\Big[\|\epsilonv_\theta(\hat\x_{\lambda},\lambda)-\epsilonv\|_2^2\Big]\dd\lambda + C,
\end{equation}
which is equivalent to the importance sampling trick in~\citep[Sec. 5.1]{song2021maximum} and the continuous-time diffusion loss in~\citep[Eq. (22)]{kingma2021variational}. Compared to Proposition~\ref{proposition:exact_solution}, we can find that the sampling and the maximum likelihood training of diffusion models can both be converted to an integral w.r.t. $\lambda$, such that the formulation is invariant to the specific noise schedules, and we summarize it in Table~\ref{tab:invariance}. Such invariance property for both training and sampling brings a new insight for understanding diffusion models. For instance, we can directly define the noise prediction model $\epsilonv_\theta$ w.r.t. the (half-)logSNR $\lambda$ instead of the time $t$, then the training and sampling for diffusion models can be done \textbf{without further choosing any ad-hoc noise schedules}. Such finding may unify the different ways of the training and the inference of diffusion models, and we leave it for future study.

\section{Proof of Theorem~\ref{thrm:order}}
\label{appendix:proof}
\subsection{Assumptions}
\label{appendix:assumptions}
Throughout this section, we denote $\x_s$ as the solution of the diffusion ODE Eq.~\eqref{eqn:diffusion_ode} starting from $\x_T$.
For DPM-Solver-$k$ we make the following assumptions:
\begin{assumption}
\label{ass:taylor}
The total derivatives $\frac{\dd^j \hat\epsilonv_\theta(\hat\x_\lambda,\lambda)}{\dd\lambda^j}$ (as a function of $\lambda$) exist and are continuous for $0 \leq j \leq k + 1$.
\end{assumption}
\begin{assumption}
\label{ass:lip}
The function $\epsilonv_\theta(\x, s)$ is Lipschitz w.r.t. to its first parameter $\x$.
\end{assumption}
\begin{assumption}
\label{ass:tech-step}
$h_{max} = \Oc(1 / M)$.
\end{assumption}

We note that the first assumption is required by Taylor's theorem Eq.~\eqref{eqn:k-th-expansion}, and the second assumption is used to replace $\epsilon_\theta(\tilde \x_s, s)$ with $\epsilon_\theta(\x_s, s) + \Oc( \x_s - \tilde \x_s )$ so that the Taylor expansion w.r.t. $\lambda_s$ is applicable.
The last one is a technical assumption to exclude a significantly large step-size.

\subsection{General Expansion of the Exponentially Weighted Integral}
\label{appendix:general_expansion}
Firstly, we derive the Taylor expansion of the exponentially weighted integral. Let $t<s$ and then $\lambda_t>\lambda_s$. Denote $h\coloneqq \lambda_t-\lambda_s$, and the $k$-th order total derivative $\hat\epsilonv_\theta^{(k)}(\hat\x_\lambda,\lambda)\coloneqq \frac{\dd^k \hat\epsilonv_\theta(\hat\x_\lambda,\lambda)}{\dd\lambda^k}$. For $n\geq 0$, the $n$-th order Taylor expansion of $\hat\epsilonv_\theta(\hat\x_{\lambda},\lambda)$ w.r.t. $\lambda$ is
\begin{equation}
    \label{eqn:taylor-epsilonv}
    \hat\epsilonv_\theta(\hat\x_\lambda, \lambda) = \sum_{k=0}^n \frac{(\lambda-\lambda_s)^k}{k!} \hat\epsilonv_\theta^{(k)}(\hat\x_{\lambda_s},\lambda_s)+\Oc(h^{n+1}).
\end{equation}
To expand the exponential integrator, we further define~\citep{hochbruck2005explicit}:
\begin{equation}
    \varphi_k(z)\coloneqq\int_0^1 e^{(1-\delta)z}\frac{\delta^{k-1}}{(k-1)!}\dm\delta,\quad\quad \varphi_0(z)=e^z
\end{equation}
and it satisfies $\varphi_k(0)=\frac{1}{k!}$ and a recurrence relation $\varphi_{k+1}(z)=\frac{\varphi_k(z)-\varphi_k(0)}{z}$. By taking the Taylor expansion of $\hat\epsilonv_\theta(\hat\x_\lambda,\lambda)$, the exponential integrator can be rewritten as
\begin{equation}
    \int_{\lambda_s}^{\lambda_t} e^{-\lambda}\hat\epsilonv_\theta(\hat\x_{\lambda},\lambda)\dd\lambda = \frac{\sigma_t}{\alpha_t}\sum_{k=0}^n h^{k+1}\varphi_{k+1}(h)\hat\epsilonv_\theta^{(k)}(\hat\x_{\lambda_s},\lambda_s)+\Oc(h^{n+2}).
\end{equation}
So the solution of $\x_t$ in Eq.~\eqref{eqn:analytic_solution} can be expanded as
\begin{equation}
\label{eqn:analytic_expansion}
    \x_t = \frac{\alpha_t}{\alpha_s}\x_s - \sigma_t\sum_{k=0}^n h^{k+1}\varphi_{k+1}(h)\hat\epsilonv_\theta^{(k)}(\hat\x_{\lambda_s},\lambda_s)+\Oc(h^{n+2}).
\end{equation}

Finally, we list the closed-forms of $\varphi_k$ for $k = 1, 2, 3$:
\begin{align}
\varphi_1(h) &= \frac{e^h - 1}{h}, \\
\varphi_2(h) &= \frac{e^h - h - 1}{h^2}, \\
\varphi_3(h) &= \frac{e^h - \nicefrac{h^2}{2} - h - 1}{h^3}.
\end{align}

\subsection{Proof of Theorem~\ref{thrm:order} when $k=1$}
\label{app:proof-k1}
\begin{proof}
Taking $n = 0, t = t_i, s = t_{i-1}$ in Eq.~\eqref{eqn:analytic_expansion}, we obtain
\begin{equation}
\label{eqn:analytic_expansion_n0}
    \x_{t_i} = \frac{\alpha_{t_i}}{\alpha_{t_{i-1}}}\x_{t_{i-1}} 
    - \sigma_t (e^{h_i} - 1)\epsilonv_\theta(\x_{t_{i-1}},{t_{i-1}})
    +\Oc(h_i^{2}).
\end{equation}
By Assumption~\ref{ass:lip} and Eq.~\eqref{eqn:1st}, it holds that
\[ \begin{aligned}
    \tilde\x_{t_i} 
    &= \frac{\alpha_{t_i}}{\alpha_{t_{i-1}}} \tilde\x_{t_{i-1}} - \sigma_{t_i} (e^{h_i} - 1)\epsilonv_\theta(\tilde\x_{t_{i-1}},t_{i-1}) \\
    &= \frac{\alpha_{t_i}}{\alpha_{t_{i-1}}} \tilde\x_{t_{i-1}} - \sigma_{t_i} (e^{h_i} - 1) \left ( \epsilonv_\theta(\x_{t_{i-1}},t_{i-1}) + \Oc(\tilde \x_{t_{i-1}} - \x_{t_{i-1}} ) \right ) \\
    &= \frac{\alpha_{t_i}}{\alpha_{t_{i-1}}} \x_{t_{i-1}} - \sigma_{t_i} (e^{h_i} - 1) \epsilonv_\theta(\x_{t_{i-1}},t_{i-1}) + \Oc(\tilde \x_{t_{i-1}} - \x_{t_{i-1}} )  \\
    &= \x_{t_i} + \Oc(h_{max}^2) + \Oc(\tilde \x_{t_{i-1}} - \x_{t_{i-1}} ).
\end{aligned} \]
Repeat this argument, we find that
\[ \tilde \x_{t_M} = \x_{t_0} + \Oc( M h_{max}^2 ) = \x_{t_0} + \Oc(h_{max}), \]
and thus completes the proof.
\end{proof}

\subsection{Proof of Theorem~\ref{thrm:order} when $k=2$}
\label{appendix:proof_2nd}
We prove the discretization error of the general form of DPM-Solver-2 in Algorithm~\ref{alg:dpm-solver-2-appendix}.
\begin{proof}
First, we consider the following update for $0 < t < s < T, h := \lambda_t - \lambda_s$.
\begin{subequations}
\begin{align}
    s_{1} &= t_{\lambda}\left(\lambda_s + r_1 h\right), \\
    \bar \uv &= \frac{\alpha_{s_1}}{\alpha_{s}} \x_{s} - \sigma_{s_1}\left(e^{r_1h} - 1\right)\epsilonv_\theta(\x_s, s), \\
    \bar \x_{t} &= \frac{\alpha_{t}}{\alpha_{s}} \x_s - \sigma_{t}\left(e^{h} - 1\right)\epsilonv_\theta(\x_s,s)
    - \frac{\sigma_t}{2r_1} (e^{h} - 1) (\epsilonv_\theta(\bar \uv, s_1) - \epsilonv_\theta(\x_s, s)).
\end{align}
\end{subequations}
Note that the above update is the same as a single step of DPM-Solver-2 with $s = t_{i-1}$ and $t = t_i$, except that $\tilde x_{t_{i-1}}$ is replaced with the exact solution $\x_{t_{i-1}}$.
Once we have proven that $\bar \x_t = \x_t + \Oc(h^3)$, we can show that $\tilde \x_{t_i} = \x_{t_i} + \Oc(h_{max}^3) + \Oc(\tilde \x_{t_{i-1}} - \x_{t_{i-1}})$ by a similar argument as in Appendix~\ref{app:proof-k1}, and therefore completes the proof.

In this remaining part we prove that $\bar \x_t = \x_t + \Oc(h^3)$.

Taking $n = 1$ in Eq.~\eqref{eqn:analytic_expansion}, we obtain
\begin{equation}
\label{eqn:analytic_expansion_n1}
    \x_t = \frac{\alpha_t}{\alpha_s}\x_s 
    - \sigma_t h\varphi_{1}(h)\epsilonv_\theta(\x_s,s)
    - \sigma_t h^{2}\varphi_{2}(h)\hat\epsilonv_\theta^{(1)}(\hat\x_{\lambda_s},\lambda_s)
    +\Oc(h^{3}).
\end{equation}

From Eq.~\eqref{eqn:taylor-epsilonv}, we have
\[
\begin{aligned}
\bar \x_{t} &= 
    \frac{\alpha_t}{\alpha_s} \x_s 
       - \sigma_{t} \left ( e^{h} - 1\right ) \epsilonv_\theta(\x_s, s) 
     - \frac{\sigma_t}{2r_1} (e^{h} - 1) (\epsilonv_\theta(\bar \uv, s_1) - \epsilonv_\theta(\x_s, s))  \\
    &= 
    \frac{\alpha_t}{\alpha_s} \x_s 
       - \sigma_{t} \left ( e^{h} - 1\right ) \epsilonv_\theta(\x_s, s) 
          -   \frac{\sigma_t}{2r_1}  \left ( e^{h} - 1\right )  \left [ 
       \epsilonv_\theta(\bar \uv, s_1) 
       - \epsilonv_\theta(\x_{s_1}, s_1) 
      \right ] \\
     &\phantom{{}={}} - \frac{\sigma_{t}}{2r_1} \left ( e^{h} - 1\right ) 
      \left [ (\lambda_{s_1} - \lambda_s)\hat\epsilonv_\theta^{(1)}(\hat\x_{\lambda_s},\lambda_s)
       + \Oc(h^2)
      \right ].
\end{aligned}
\]
Note that by the Lipschitzness of $\epsilonv_\theta$ w.r.t. $\x$ (Assumption~\ref{ass:lip}),
\[
       \| \epsilonv_\theta(\bar \uv, s_1) 
       - \epsilonv_\theta(\x_{s_1}, s_1)  \|
       =
       \Oc(\| \bar \uv - \x_{s_1} \|) = \Oc(h^2),
\]
where the last equation follows from a similar argument in the proof of $k=1$. Since $e^h - 1 = \Oc(h)$, the second term of the above display is $\Oc(h^3)$.

As $\lambda_{s_1} - \lambda_s = r_1h$, $\varphi_i(h) = (e^h - 1) / h$ and $\varphi_2(h) =  (e^h - h - 1) / h^2$, we find
\[
\begin{aligned}
\x_t - \bar \x_t
&= \sigma_t \left [ h^2 \varphi_2(h) - (e^h - 1)\frac{ \lambda_{s_1} - \lambda_s}{2r_1} \right ] \hat\epsilonv_\theta^{(1)} (\hat \x_{\lambda_s}, \lambda_s)
+ \Oc(h^3).
\end{aligned}
\]
Then, the proof is completed by noticing that 
\[ h^2\varphi_2(h) - (e^h - 1) \frac{\lambda_{s_1} - \lambda_s}{2r_1} = (2e^h - h - 2 - he^h) / 2 = \Oc(h^3). \]
\end{proof}

\subsection{Proof of Theorem~\ref{thrm:order} when $k=3$}
\label{appendix:proof_3rd}
\begin{proof}
As in Appendix~\ref{appendix:proof_2nd}, it suffices to show that the following update has error $\bar \x_t = \x_t + \Oc(h^4)$ for $0 < t < s < T$ and $h = \lambda_s - \lambda_t$.
\begin{subequations}
\begin{align}
    s_{1} &= t_\lambda\left(\lambda_s + r_1 h\right),\quad s_{2} = t_\lambda\left(\lambda_{s} + r_2 h\right), \\
    \bar \uv_{1} &=\frac{\alpha_{s_{1}}}{\alpha_{s}} \x_{s} - \sigma_{s_{1}}\left(e^{r_1 h} - 1\right)\epsilonv_\theta( \x_{s},s), \\
    \Dv_{1} &= \epsilonv_\theta(\bar \uv_{1},s_{1}) - \epsilonv_\theta(\x_{s},s), \\
    \bar \uv_{2} &= \frac{\alpha_{s_{2}}}{\alpha_{s}} \x_{s}  
    - \sigma_{s_{2}}\left(e^{r_2 h} - 1\right)\epsilonv_\theta(\x_{s},s) - \frac{\sigma_{s_{2}}r_2}{r_1}\left( \frac{e^{r_2 h} - 1}{r_2 h} - 1\right)\Dv_{1}, \\
    \Dv_{2} &= \epsilonv_\theta(\bar \uv_{2},s_{2}) - \epsilonv_\theta(\x_{s},s), \\
    \bar \x_{t} &= \frac{\alpha_{t}}{\alpha_{s}} \x_{s}
        - \sigma_{t}\left(e^{h} - 1\right)\epsilonv_\theta(\x_{s},s) - \frac{\sigma_{t}}{r_2}\left( \frac{e^{h} - 1}{h} - 1\right)\Dv_{2}.
\end{align}
\end{subequations}

First, we prove that
\begin{equation}
    \bar \uv_2 = \x_{s_2} + \Oc(h^3).
\end{equation}
Similar to the proof in Appendix~\ref{appendix:proof_2nd}, since $\frac{e^{r_2h - 1}}{r_2h} - 1 = \Oc(h)$ and $\bar \uv_1 =\x_{s_1} + \Oc(h^2)$, then
\[ \begin{aligned}
\bar \uv_2
&= \frac{\alpha_{s_2}}{\alpha_s} \x_s - \sigma_{s_2} \left ( e^{r_2h} - 1 \right ) \epsilonv_\theta(\x_s, s)
\\
&\phantom{{}={}} - \sigma_{s_2} \frac{r_2}{r_1} \left (\frac{e^{r_2h} - 1}{r_2h} - 1 \right ) \left ( \epsilonv_\theta(\x_{s_1}, s_1) - \epsilonv_\theta(\x_s, s) \right ) + \Oc(h^3) \\
&= \frac{\alpha_{s_2}}{\alpha_s} \x_s - \sigma_{s_2} \left ( e^{r_2h} - 1 \right ) \epsilonv_\theta(\x_s, s)
\\ 
& \phantom{{}={}} - \sigma_{s_2} \frac{r_2}{r_1} \left (\frac{e^{r_2h} - 1}{r_2h} - 1 \right ) \epsilonv^{(1)}_\theta(\x_{s}, s) (\lambda_{s_1} - \lambda_s) + \Oc(h^3).
\end{aligned} \]
Let $h_2 = r_2h$, then following the same line of arguments in the proof of Appendix~\ref{appendix:proof_2nd}, it suffices to check that 
\[
\begin{aligned}
   \varphi_1(h_2) h_2 &= e^{h_2} - 1, \\
   \varphi_2(h_2) h_2^2 &= 
  \frac{r_2}{r_1} \left (\frac{e^{h_2} - 1}{h_2} - 1 \right ) (\lambda_{s_1} - \lambda_s) + \Oc(h^3), \\
\end{aligned}
\]
which holds by applying Taylor expansion.

Using $\bar \uv_2 = \x_{s_2} + \Oc(h^3)$ and $\lambda_{s_2} - \lambda_s = r_2h = \frac{2}{3}h$, we find that
\[ \begin{aligned}
    \bar\x_{t} 
    &= \frac{\alpha_{t}}{\alpha_s} \x_s
        - \sigma_{t}\left(e^h - 1\right)\epsilonv_\theta(\x_s,s)
        - \sigma_{t}\frac{1}{r_2}\left( \frac{e^{h} - 1}{h} - 1\right)
          \big(\epsilonv_\theta(\bar\uv_2,s_2) - \epsilonv_\theta(\x_{s},s)\big) \\
    &= \frac{\alpha_{t}}{\alpha_s} \x_s
        - \sigma_{t}\left(e^h - 1\right)\epsilonv_\theta(\x_s,s)
        - \sigma_{t}\frac{1}{r_2}\left( \frac{e^{h} - 1}{h} - 1\right)
          \big(\epsilonv_\theta(\x_{s_2},s_2) - \epsilonv_\theta(\x_{s},s)\big) + \Oc(h^4) \\
    &= \frac{\alpha_{t}}{\alpha_s} \x_s
        - \sigma_{t}\left(e^h - 1\right)\epsilonv_\theta(\x_s,s)
   \\
   &\phantom{{}={}} - \sigma_{t}\frac{1}{r_2}\left( \frac{e^{h} - 1}{h} - 1\right)
          \big(\epsilonv^{(1)}_\theta(\x_{s},s) r_2h + \frac{1}{2}\epsilonv^{(2)}_\theta(\x_{s},s) r_2^2h^2 \big) + \Oc(h^4).
  \end{aligned} \]

Comparing with the Taylor expansion in Eq.~\eqref{eqn:analytic_expansion} with $n = 2$:
    \[ \x_t = \frac{\alpha_t}{\alpha_s}\x_s 
    - \sigma_t h\varphi_{1}(h)\epsilonv_\theta(\x_s,s)
    - \sigma_t h^{2}\varphi_{2}(h)\epsilonv_\theta^{(1)}(\x_{s},s)
    - \sigma_t h^{3}\varphi_{3}(h)\epsilonv_\theta^{(2)}(\x_{s},s)
    +\Oc(h^{4}), \]
we need to check the following conditions:
\[
\begin{aligned}
h\varphi_1(h) &= e^h - 1, \\
h^2\varphi_2(h) &= \left ( \frac{e^h - 1}{h} - 1\right ) h, \\
h^3\varphi_3(h) &= \left ( \frac{e^h - 1}{h} - 1\right ) \frac{r_2h^2}{2} + \Oc(h^4).
\end{aligned}
\]
The first two conditions are clear. The last condition follows from
\[ \begin{aligned}
h^3 \varphi_3(h) &= e^h  - 1 - h - \frac{h^2}{2} = \frac{h^3}{6} + \Oc(h^4) = 
 \left ( \frac{e^h - 1}{h} - 1\right ) \frac{r_2h^2}{2}.
\end{aligned}
\]
Therefore, $\bar \x_t = \x_t + \Oc(h^4)$.
\end{proof}

\subsection{Connections to Explicit Exponential Runge-Kutta (expRK) Methods}
\label{appendix:expRK}
Assume we have an ODE with the following form:
\begin{equation*}
    \frac{\dd\x_{t}}{\dd t} = \alpha \x_t + \Nv(\x_t,t),
\end{equation*}
where $\alpha\in \R$ and $\Nv(\x_t,t)\in\R^D$ is a non-linear function of $\x_t$. Given an initial value $\x_t$ at time $t$, for $h>0$, the true solution at time $t+h$ is
\begin{equation*}
    \x_{t+h} = e^{\alpha h}\x_t
        + e^{\alpha h}\int_0^h e^{-\alpha\tau}\Nv(\x_{t+\tau},t+\tau)\dm\tau.
\end{equation*}
The exponential Runge-Kutta methods~\cite{hochbruck2010exponential,hochbruck2005explicit} use some intermediate points to approximate the integral $\int e^{-\alpha\tau}\Nv(\x_{t+\tau},t+\tau)\dm\tau$. Our proposed DPM-Solver is inspired by the same technique for approximating the same integral with $\alpha=1$ and $\Nv=\tilde\epsilonv_\theta$. 
However, DPM-Solver is different from the expRK methods, because their linear term $e^{\alpha h}\x_t$ is different from our linear term $\frac{\alpha_{t+h}}{\alpha_t}\x_t$. In summary, DPM-Solver is inspired by the same technique of expRK for deriving high-order approximations of the exponentially weighted integral, but the formulation of DPM-Solver is different from expRK, and DPM-Solver is customized for the specific formulation of diffusion ODEs.
\section{Algorithms of DPM-Solvers}
\label{appendix:algorithm}
We firstly list the detailed DPM-Solver-1, 2, 3 in Algorithms~\ref{alg:dpm-solver-1-appendix}, \ref{alg:dpm-solver-2-appendix}, \ref{alg:dpm-solver-3-appendix}. Note that DPM-Solver-2 is the general case with $r_1\in(0,1)$, and we usually set $r_1=0.5$ for DPM-Solver-2, as in Sec.~\ref{sec:solver}.

\begin{algorithm}[htbp]
    \centering
    \caption{DPM-Solver-1.}\label{alg:dpm-solver-1-appendix}
    \begin{algorithmic}[1]
    \Require initial value $\x_T$, time steps $\{t_i\}_{i=0}^M$, model $\epsilonv_\theta$
        \State \textbf{def} dpm-solver-1$(\tilde\x_{t_{i-1}}, t_{i-1}, t_i)$:
            \State\quad $h_i \gets \lambda_{t_{i}} - \lambda_{t_{i-1}}$
            \State\quad $\tilde\x_{t_i} \gets \frac{\alpha_{t_i}}{\alpha_{t_{i-1}}} \tilde\x_{t_{i-1}} - \sigma_{t_i}\left(e^{h_i} - 1\right)\epsilonv_\theta(\tilde \x_{t_{i-1}},t_{i-1})$
            \State\quad \Return $\tilde\x_{t_i}$
        \State $\tilde\x_{t_0}\gets\x_T$
        \For{$i\gets 1$ to $M$}
        \State $\tilde\x_{t_i} \gets \text{dpm-solver-1}(\tilde\x_{t_{i-1}}, t_{i-1}, t_i)$
        \EndFor
    \State \Return $\tilde\x_{t_M}$
    \end{algorithmic}
\end{algorithm}

\begin{algorithm}[htbp]
    \centering
    \caption{DPM-Solver-2 (general version).}\label{alg:dpm-solver-2-appendix}
    \begin{algorithmic}[1]
    \Require initial value $\x_T$, time steps $\{t_i\}_{i=0}^M$, model $\epsilonv_\theta$, $r_1=0.5$
        \State \textbf{def} dpm-solver-2$(\tilde\x_{t_{i-1}}, t_{i-1}, t_i, r_1)$:
            \State\quad $h_i \gets \lambda_{t_{i}} - \lambda_{t_{i-1}}$
            \State\quad $s_{i} \gets t_{\lambda}\left(\lambda_{t_{i-1}} + r_1 h_i\right)$
            \State\quad $\uv_{i} \gets \frac{\alpha_{s_i}}{\alpha_{t_{i-1}}} \tilde\x_{t_{i-1}} - \sigma_{s_i}\left(e^{r_1 h_i} - 1\right)\epsilonv_\theta(\tilde \x_{t_{i-1}},t_{i-1})$
            \State\quad $\tilde\x_{t_{i}} \gets \frac{\alpha_{t_{i}}}{\alpha_{t_{i-1}}} \tilde\x_{t_{i-1}} - \sigma_{t_{i}}(e^{h_i} - 1)\epsilonv_\theta(\tilde\x_{t_{i-1}},t_{i-1}) - \frac{\sigma_{t_{i}}}{2r_1}(e^{h_i} - 1)(\epsilonv_\theta(\uv_{i},s_i) - \epsilonv_\theta(\tilde\x_{t_{i-1}}, t_{i-1}))$
            \State\quad \Return $\tilde\x_{t_i}$
        \State $\tilde\x_{t_0}\gets\x_T$
        \For{$i\gets 1$ to $M$}
        \State $\tilde\x_{t_i} \gets \text{dpm-solver-2}(\tilde\x_{t_{i-1}}, t_{i-1}, t_i, r_1)$
        \EndFor
    \State \Return $\tilde\x_{t_M}$
    \end{algorithmic}
\end{algorithm}

\begin{algorithm}[htbp]
    \centering
    \caption{DPM-Solver-3.}\label{alg:dpm-solver-3-appendix}
    \begin{algorithmic}[1]
    \Require initial value $\x_T$, time steps $\{t_i\}_{i=0}^M$, model $\epsilonv_\theta$, $r_1=\frac{1}{3}$, $r_2=\frac{2}{3}$
    \State \textbf{def} dpm-solver-3$(\tilde\x_{t_{i-1}}, t_{i-1}, t_i, r_1, r_2)$:
        \State\quad $h_i \gets \lambda_{t_{i}} - \lambda_{t_{i-1}}$
    \State
        \quad $s_{2i-1} \gets t_\lambda\left(\lambda_{t_{i-1}} + r_1 h_i\right),\quad s_{2i} \gets t_\lambda\left(\lambda_{t_{i-1}} + r_2 h_i\right)$ 
    \State 
        \quad $\uv_{2i-1} \gets \frac{\alpha_{s_{2i-1}}}{\alpha_{t_{i-1}}} \tilde\x_{t_{i-1}} - \sigma_{s_{2i-1}}\left(e^{r_1 h_i} - 1\right)\epsilonv_\theta(\tilde \x_{t_{i-1}},t_{i-1})$
    \State 
        \quad$\Dv_{2i-1} \gets \epsilonv_\theta(\uv_{2i-1},s_{2i-1}) - \epsilonv_\theta(\tilde\x_{t_{i-1}},t_{i-1})$
    \State
        \quad$\uv_{2i} \gets \frac{\alpha_{s_{2i}}}{\alpha_{t_{i-1}}} \tilde\x_{t_{i-1}} 
        - \sigma_{s_{2i}}\left(e^{r_2 h_i} - 1\right)\epsilonv_\theta(\tilde\x_{t_{i-1}},t_{i-1}) - \frac{\sigma_{s_{2i}}r_2}{r_1}\left( \frac{e^{r_2 h_i} - 1}{r_2 h_i} - 1\right)\Dv_{2i-1}$
    \State 
        \quad$\Dv_{2i} \gets \epsilonv_\theta(\uv_{2i},s_{2i}) - \epsilonv_\theta(\tilde\x_{t_{i-1}},t_{i-1})$
    \State
        \quad$\tilde\x_{t_{i}} \gets \frac{\alpha_{t_{i}}}{\alpha_{t_{i-1}}} \tilde\x_{t_{i-1}}
            - \sigma_{t_{i}}\left(e^{h_i} - 1\right)\epsilonv_\theta(\tilde\x_{t_{i-1}},t_{i-1}) - \frac{\sigma_{t_{i}}}{r_2}\left( \frac{e^{h_i} - 1}{h} - 1\right)\Dv_{2i}$
    \State\quad \Return $\tilde\x_{t_{i}}$
    
    \State $\tilde\x_{t_0}\gets\x_T$
    \For{$i\gets 1$ to $M$}
        \State $\tilde\x_{t_i} \gets \text{dpm-solver-3}(\tilde\x_{t_{i-1}}, t_{i-1}, t_i, r_1, r_2)$
        \EndFor
        \State  \Return $\tilde\x_{t_M}$
    \end{algorithmic}
\end{algorithm}

Then we list the adaptive step size algorithms, named as \textit{DPM-Solver-12} (combining 1 and 2; Algorithm~\ref{algorithm:dpm_solver_12}) and \textit{DPM-Solver-23} (combining 2 and 3; Algorithm~\ref{algorithm:dpm_solver_23}). We follow~\citep{jolicoeur2021gotta} to let the absolute tolerance $\epsilon_{\text{atol}}=\frac{\x_{\text{max}} - \x_{\text{min}}}{256}$ for image data, which is $0.0078$ for VP type DPMs. We can tune the relative tolerance $\epsilon_{\text{rtol}}$ to balance the accuracy and NFE, and we find that $\epsilon_{\text{rtol}}=0.05$ is good enough and can converge quickly.

In practice, the inputs of the adaptive step size solvers are batch data. We simply choose $E_2$ and $E_3$ as the maximum value of all the batch data. Besides, we implement the comparison $s > \epsilon$ by $|s-\epsilon|>10^{-5}$ to avoid numerical issues.

\begin{algorithm}[htbp]
\centering
\caption{(\textbf{DPM-Solver-12}) Adaptive step size algorithm by combining DPM-Solver-1 and 2.}\label{algorithm:dpm_solver_12}
\begin{algorithmic}[1]
\Require start time $T$, end time $\epsilon$, initial value $\x_T$, model $\epsilonv_\theta$, data dimension $D$, hyperparameters $\epsilon_{\text{rtol}}=0.05$, $\epsilon_{\text{atol}}=0.0078$, $h_{\text{init}}=0.05$, $\theta=0.9$
\Ensure the approximated solution $\x_{\epsilon}$ at time $\epsilon$
\State $s\gets T$, $h\gets h_{\text{init}}$, $\x\gets \x_T$, $\x_{\text{prev}}\gets \x_T$, $r_1\gets \frac{1}{2}$, $\text{NFE}\gets 0$
\While{$s > \epsilon$}
    \State $t\gets t_{\lambda}(\lambda_s + h)$
    \State $\x_1\gets \text{dpm-solver-1}(\x, s, t)$
    \State $\x_2\gets \text{dpm-solver-2}(\x, s, t, r_1)$ (Share the same function value $\epsilonv_\theta(\x, s)$ with dpm-solver-1.)
    \State $\deltav \gets \max(\epsilon_{\text{atol}}, \epsilon_{\text{rtol}}\max(|\x_1|, |\x_{\text{prev}}|))$
    \State $E_2\gets \frac{1}{\sqrt{D}}\|\frac{\x_1-\x_2}{\deltav}\|_2$
    \If{$E_2\leq 1$}
        \State $\x_{\text{prev}}\gets \x_1$, $\x\gets \x_2$, $s\gets t$
    \EndIf
    \State $h\gets \min(\theta h E_2^{-\frac{1}{2}}, \lambda_{\epsilon} - \lambda_s)$
    \State $\text{NFE}\gets \text{NFE} + 2$
\EndWhile
\State \Return $\x$, $\text{NFE}$

\end{algorithmic}
\end{algorithm}

\begin{algorithm}[htbp]
\centering
\caption{(\textbf{DPM-Solver-23}) Adaptive step size algorithm by combining DPM-Solver-2 and 3.}\label{algorithm:dpm_solver_23}
\begin{algorithmic}[1]
\Require start time $T$, end time $\epsilon$, initial value $\x_T$, model $\epsilonv_\theta$, data dimension $D$, hyperparameters $\epsilon_{\text{rtol}}=0.05$, $\epsilon_{\text{atol}}=0.0078$, $h_{\text{init}}=0.05$, $\theta=0.9$
\Ensure the approximated solution $\x_{\epsilon}$ at time $\epsilon$
\State $s\gets T$, $h\gets h_{\text{init}}$, $\x\gets \x_T$, $\x_{\text{prev}}\gets \x_T$, $r_1\gets \frac{1}{3}$, $r_2\gets \frac{2}{3}$, $\text{NFE}\gets 0$
\While{$s > \epsilon$}
    \State $t\gets t_{\lambda}(\lambda_s + h)$
    \State $\x_2\gets \text{dpm-solver-2}(\x, s, t, r_1)$
    \State $\x_3\gets \text{dpm-solver-3}(\x, s, t, r_1, r_2)$ (Share the same function values with dpm-solver-2.)
    \State $\deltav \gets \max(\epsilon_{\text{atol}}, \epsilon_{\text{rtol}}\max(|\x_2|, |\x_{\text{prev}}|))$
    \State $E_3\gets \frac{1}{\sqrt{D}}\|\frac{\x_2-\x_3}{\deltav}\|_2$
    \If{$E_3\leq 1$}
        \State $\x_{\text{prev}}\gets \x_2$, $\x\gets \x_3$, $s\gets t$
    \EndIf
    \State $h\gets \min(\theta h E_3^{-\frac{1}{3}}, \lambda_{\epsilon} - \lambda_s)$
    \State $\text{NFE}\gets \text{NFE} + 3$
\EndWhile
\State \Return $\x$, $\text{NFE}$

\end{algorithmic}
\end{algorithm}

\section{Implementation Details of DPM-Solver}
\label{appendix:implementation}

\subsection{End Time of Sampling}
Theoretically, we need to solve diffusion ODEs from time $T$ to time $0$ to generate samples. Practically, the training and evaluation for the noise prediction model $\epsilonv_\theta(\x_t,t)$ usually start from time $T$ to time $\epsilon$ to avoid numerical issues for $t$ near to $0$, where $\epsilon>0$ is a hyperparameter~\citep{song2020score}.

In contrast to the sampling methods based on diffusion SDEs~\citep{ho2020denoising,song2020score}, We don't add the ``denoising'' trick at the final step at time $\epsilon$ (which is to set the noise variance to zero), and we just solve diffusion ODEs from $T$ to $\epsilon$ by DPM-Solver, since we find it performs well enough.

For discrete-time DPMs, we firstly convert the model to continuous time (see Appendix~\ref{appendix:discrete}), and then solver it from time $T$ to time $t$.

\subsection{Sampling from Discrete-Time DPMs}
\label{appendix:discrete}
In this section, we discuss the more general case for discrete-time DPMs, in which we consider the 1000-step DPMs~\citep{ho2020denoising} and the 4000-step DPMs~\citep{nichol2021improved}, and we also consider the end time $\epsilon$ for sampling.

Discrete-time DPMs~\citep{ho2020denoising} train the noise prediction model at $N$ fixed time steps $\{t_n\}_{n=1}^N$. In practice, $N=1000$ or $N=4000$, and the implementation of the 4000-step DPMs~\citep{nichol2021improved} converts the time steps of 4000-step DPMs to the range of 1000-step DPMs. Specifically, the noise prediction model is parameterized by $\tilde\epsilonv_\theta(\x_n, \frac{1000n}{N})$ for $n=0,\dots,N-1$, where each $\x_n$ is corresponding to the value at time $t_{n+1}$. In practice, these discrete-time DPMs usually choose uniform time steps between $[0,T]$, thus $t_{n}=\frac{nT}{N}$, for $n=1,\dots,N$.

However, the discrete-time noise prediction model cannot predict the noise at time less than the smallest time $t_1$. As the smallest time step $t_1=\frac{T}{N}$ and the corresponding discrete-time noise prediction model at time $t_1$ is $\tilde\epsilonv_\theta(\x_0, 0)$, we need to ``scale'' the discrete time steps $[t_1, t_N]=[\frac{T}{N}, T]$ to the continuous time range $[\epsilon, T]$. We propose two types of scaling as following.

\textbf{Type-1.}\quad Scale the discrete time steps $[t_1, t_N]=[\frac{T}{N}, T]$ to the continuous time range $[\frac{T}{N}, T]$, and let $\epsilonv_\theta(\cdot, t)=\epsilonv_\theta(\cdot, \frac{T}{N})$ for $t\in[\epsilon, \frac{T}{N}]$. In this case, we can define the continuous-time noise prediction model by
\begin{equation}
    \epsilonv_\theta(\x, t) = \tilde\epsilonv_\theta\left(\x, 1000 \cdot \max\left(t - \frac{T}{N}, 0\right)\right),
\end{equation}
where the continuous time $t\in [\epsilon, \frac{T}{N}]$ maps to the discrete input $0$, and the continuous time $T$ maps to the discrete input $\frac{1000(N-1)}{N}$.

\textbf{Type-2.}\quad Scale the discrete time steps $[t_1, t_N]=[\frac{T}{N}, T]$ to the continuous time range $[0, T]$. In this case, we can define the continuous-time noise prediction model by
\begin{equation}
    \epsilonv_\theta(\x, t) = \tilde\epsilonv_\theta\left(\x, 1000 \cdot \frac{(N-1)t}{NT}\right),
\end{equation}
where the continuous time $0$ maps to the discrete input $0$, and the continuous time $T$ maps to the discrete input $\frac{1000(N-1)}{N}$.

Note that the input time of $\tilde\epsilonv_\theta$ may not be integers, but we find that the noise prediction model can still work well, and we hypothesize that it is because of the smooth time embeddings (e.g., position embeddings~\citep{ho2020denoising}). By such reparameterization, the noise prediction model can adopt the continuous-time steps as input, and thus we can also use DPM-Solver for fast sampling.

In practice, we have $T=1$, and the smallest discrete time $t_1=10^{-3}$. For fixed $K$ number of function evaluations, we empirically find that for small $K$, the Type-1 with $\epsilon=10^{-3}$ may have better sample quality, and for large $K$, the Type-2 with $\epsilon=10^{-4}$ may have better sample quality. We refer to Appendix~\ref{appendix:experiment} for detailed results.

\subsection{DPM-Solver in 20 Function Evaluations}
\label{appendix:dpm-solver-fast}
Given a fixed budget $K\leq 20$ of the number of function evaluations, we uniformly divide the interval $[\lambda_T, \lambda_\epsilon]$ into $M = (\lfloor K / 3\rfloor + 1)$ segments, and take $M$ steps to generate samples. The $M$ steps are dependent on the remainder $R$ of $K$ mod $3$ to make sure the total number of function evaluations is exactly $K$.

\begin{itemize}
\item 
If $R = 0$, we firstly take $M - 2$ steps of DPM-Solver-3, and then take $1$ step of DPM-Solver-2 and 1 step of DPM-Solver-1. The total number of function evaluations is $3\cdot(\frac{K}{3} - 1) + 2 + 1 = K$.

    \item 
If $R = 1$, we firstly take $M - 1$ steps of DPM-Solver-3 and then take $1$ step of DPM-Solver-1. The total number of function evaluations is $3\cdot(\frac{K-1}{3}) + 1=K$.

    \item 
If $R = 2$, we firstly take $M - 1$ steps of DPM-Solver-3 and then take $1$ step of DPM-Solver-2. The total number of function evaluations is $3\cdot(\frac{K-2}{3}) + 2 = K$.
\end{itemize}

We empirically find that this design of time steps can greatly improve the generation quality, and DPM-Solver can generate comparable samples in 10 steps and high-quality samples in 20 steps.

\subsection{Analytical Formulation of the function $t_\lambda(\cdot)$ (the inverse function of $\lambda(t)$)}
The costs of computing $t_\lambda(\cdot)$ is negligible, because for the noise schedules of $\alpha_t$ and $\sigma_t$ used in previous DPMs (``linear'' and ``cosine'')~\citep{ho2020denoising,nichol2021improved}, both $\lambda(t)$ and its inverse function $t_\lambda(\cdot)$ have analytic formulations. We mainly consider the variance preserving type here, since it is the most widely-used type. The functions of other types (variance exploding and sub-variance preserving type) can be similarly derived.

\textbf{Linear Noise Schedule~\citep{ho2020denoising}.}\quad We have
\begin{equation*}
    \log\alpha_t = -\frac{(\beta_1-\beta_0)}{4}t^2 - \frac{\beta_0}{2}t,
\end{equation*}
where $\beta_0=0.1$ and $\beta_1=20$, following~\citep{song2020score}. As $\sigma_t=\sqrt{1-\alpha_t^2}$, we can compute $\lambda_t$ analytically. Moreover, the inverse function is
\begin{equation*}
    t_\lambda(\lambda) = \frac{1}{\beta_1-\beta_0}\left(\sqrt{\beta_0^2 + 2(\beta_1-\beta_0)\log\left(e^{-2\lambda}+1\right)} - \beta_0\right).
\end{equation*}
To reduce the influence of numerical issues, we can compute $t_\lambda$ by the following equivalent formulation:
\begin{equation*}
    t_\lambda(\lambda) = \frac{2\log\left(e^{-2\lambda}+1\right)}{\sqrt{\beta_0^2 + 2(\beta_1-\beta_0)\log\left(e^{-2\lambda}+1\right)} + \beta_0}.
\end{equation*}
And we solve diffusion ODEs between $[\epsilon, T]$, where $T=1$.

\textbf{Cosine Noise Schedule~\citep{nichol2021improved}.}\quad Denote
\begin{equation*}
    \log\alpha_t = \log\left(\cos\left( \frac{\pi}{2}\cdot\frac{t+s}{1+s} \right)\right) - \log\left(\cos\left( \frac{\pi}{2}\cdot\frac{s}{1+s} \right)\right),
\end{equation*}
where $s=0.008$, following~\citep{nichol2021improved}. As~\citep{nichol2021improved} clipped the derivatives to ensure the numerical stability, we also clip the maximum time $T=0.9946$.
As $\sigma_t=\sqrt{1-\alpha_t^2}$, we can compute $\lambda_t$ analytically. Moreover, given a fixed $\lambda$, let
\begin{equation*}
    f(\lambda) = -\frac{1}{2}\log\left(e^{-2\lambda} + 1\right),
\end{equation*}
which computes the corresponding $\log\alpha$ for $\lambda$. Then the inverse function is
\begin{equation*}
    t_{\lambda}(\lambda) = 
        \frac{2(1+s)}{\pi}\arccos\left(e^{f(\lambda) + \log\cos\left(\frac{\pi s}{2(1+s)}\right)}\right) - s.
\end{equation*}
And we solve diffusion ODEs between $[\epsilon, T]$, where $T=0.9946$.

\subsection{Conditional Sampling by DPM-Solver}
DPM-Solver can also be used for conditional sampling, with a simple modification. The conditional generation needs to sample from the conditional diffusion ODE~\citep{song2020score, dhariwal2021diffusion} which includes the conditional noise prediction model. We follow the classifier guidance method~\citep{dhariwal2021diffusion} to define the conditional noise prediction model as $\epsilonv_\theta(\x_t,t,y)\coloneqq \epsilonv_\theta(\x_t,t) - s \cdot \sigma_t\nabla_{\x}\log p_t(y|\x_t;\theta)$, where $p_t(y|\x_t;\theta)$ is a pre-trained classifier and $s$ is the classifier guidance scale (default is 1.0). Thus, we can use DPM-Solver to solve this diffusion ODE for fast conditional sampling, as shown in Fig.~\ref{fig:intro}.

\subsection{Numerical Stability}
As we need to compute $e^{h_i} - 1$ in the algorithm of DPM-Solver, we follow~\citep{kingma2021variational} to use \texttt{expm1($h_i$)} instead of \texttt{exp($h_i$)-1} to improve numerical stability.

\section{Experiment Details}
\label{appendix:experiment}

We test our method for sampling the most widely-used \textit{variance-preserving} (VP) type  DPMs~\citep{sohl2015deep,ho2020denoising}. 
In this case, we have $\alpha_t^2+\sigma_t^2=1$ for all $t\in[0,T]$ and $\tilde\sigma=1$. In spite of this, our method and theoretical results are general and independent of the choice of the noise schedule $\alpha_t$ and $\sigma_t$. 

For all experiments, we evaluate DPM-Solver on NVIDIA A40 GPUs. However, the computation resource can be other types of GPU, such as NVIDIA GeForce RTX 2080Ti, because we can tune the batch size for sampling.

\subsection{Diffusion ODEs w.r.t. $\lambda$}
\label{appendix:RK}
Alternatively, the diffusion ODE can be reparameterized to the $\lambda$ domain. In this section, we propose the formulation of diffusion ODEs w.r.t. $\lambda$ for VP type, and other types can be similarly derived.

For a given $\lambda$, denote $\hat\alpha_{\lambda}\coloneqq \alpha_{t(\lambda)}$, $\hat\sigma_{\lambda}\coloneqq \sigma_{t(\lambda)}$. As $\hat\alpha_{\lambda}^2 + \hat\sigma_{\lambda}^2 = 1$, we can prove that $\frac{\dd\lambda}{\dd\hat\alpha_{\lambda}}=\frac{1}{\hat\alpha_{\lambda}\hat\sigma^2_{\lambda}}$, so $\frac{\dd\log \hat\alpha_{\lambda}}{\dd \lambda}=\hat\sigma^2_{\lambda}$. Applying change-of-variable to Eq.~\eqref{eqn:diffusion_ode}, we have
\begin{equation}
\label{eqn:diffusion_ode_logSNR}
    \frac{\dd\hat\xv_\lambda}{\dd\lambda} = \hat\hv_\theta(\hat\x_\lambda,\lambda)\coloneqq \hat\sigma_\lambda^2 \hat\x_\lambda - \hat\sigma_\lambda \hat\epsilonv_\theta(\hat\x_\lambda,\lambda).
\end{equation}
The ODE Eq.~\eqref{eqn:diffusion_ode_logSNR} can be also solved directly by RK methods, and we use such formulation for the experiments of RK2 ($\lambda$) and RK3 ($\lambda$) in Table~\ref{tab:RK_compare}.

\subsection{Code Implementation}
We implement our code with both JAX (for continuous-time DPMs) and PyTorch (for discrete-time DPMs), and our code is released at \url{https://github.com/LuChengTHU/dpm-solver}.

\subsection{Sample Quality Comparison with Continuous-Time Sampling Methods}
\begin{table}[t]
    \centering
    \caption{\small{Sample quality measured by FID $\downarrow$ on CIFAR-10 dataset with continuous-time methods, varying the number of function evaluations (NFE).}}
    \label{tab:continuous_fid}
    \resizebox{\textwidth}{!}{%
    \begin{tabular}{lllrrrrrrr}
    \toprule
      \multicolumn{3}{l}{Sampling method $\backslash$ NFE} & 10 & 12 & 15 & 20 & 50 & 200 & 1000 \\
    \midrule
     \multicolumn{9}{l}{CIFAR-10 (continuous-time model (VP deep)~\citep{song2020score}, linear noise schedule)}  & \\
     \arrayrulecolor{black!30}\midrule
      \multirow{3}{*}{SDE} & \multirow{2}{*}{Euler (denoise)~\citep{song2020score}} & $\epsilon=10^{-3}$ & 304.73 & 278.87 & 248.13 & 193.94 & 66.32 & 12.27 & \textbf{2.44} \\
      & & $\epsilon=10^{-4}$ & 444.63 & 427.54 & 395.95 & 300.41 & 101.66 & 22.98 & 5.01 \\
      \arrayrulecolor{black!30}\cmidrule{2-10}
      & Improved Euler~\citep{jolicoeur2021gotta} & $\epsilon=10^{-3}$ & \multicolumn{7}{c}{82.42(NFE=48), 2.73(NFE=151), 2.44(NFE=180)}\\
      \arrayrulecolor{black!30}\midrule
      \multirow{4}{*}{ODE} & \multirow{2}{*}{RK45 Solver~\citep{dormand1980family,song2020score}} & $\epsilon=10^{-3}$ & \multicolumn{7}{c}{19.55(NFE=26), 17.81(NFE=38), 3.55(NFE=62)} \\
      & & $\epsilon=10^{-4}$ & \multicolumn{7}{c}{51.66(NFE=26), 21.54(NFE=38),  12.72(NFE=50), 2.61(NFE=62)} \\
      \arrayrulecolor{black!30}\cmidrule{2-10}
      & \multirow{2}{*}{\shortstack{DPM-Solver\\(\textbf{ours})}} & $\epsilon=10^{-3}$ & \textbf{4.70} & \textbf{3.75} & \textbf{3.24} & 3.99 & \multicolumn{3}{c}{3.84 (NFE = 42)} \\
      & & $\epsilon=10^{-4}$ & 6.96 & 4.93 & 3.35 & \textbf{2.87} & \multicolumn{3}{c}{\textbf{2.59 (NFE = 51)}}\\ %
    \arrayrulecolor{black}\bottomrule
    \end{tabular}%
    }
\end{table}

Table~\ref{tab:continuous_fid} shows the detailed FID results, which is corresponding to Fig.~\ref{fig:cifar10_fid_continuous}. We use the official code and checkpoint in~\citep{song2020score}, the code license is Apache License 2.0. We use their released ``checkpoint\_8'' of the ``VP deep'' type. We compare methods for $\epsilon=10^{-3}$ and $\epsilon=10^{-4}$. We find that the sampling methods based on diffusion SDEs can achieve better sample quality with $\epsilon=10^{-3}$; and that the sampling methods based on diffusion ODEs can achieve better sample quality with $\epsilon=10^{-4}$. For DPM-Solver, we find that DPM-Solver with less than 15 NFE can achieve better FID with $\epsilon=10^{-3}$ than $\epsilon=10^{-4}$, while DPM-Solver with more than 15 NFE can achieve better FID with $\epsilon=10^{-4}$ than $\epsilon=10^{-3}$.

For the diffusion SDEs with Euler discretization, we use the PC sampler in~\citep{song2020score} with ``euler\_maruyama'' predictor and no corrector, which uses uniform time steps between $T$ and $\epsilon$. We add the ``denoise'' trick at the final step, which can greatly improve the FID score for $\epsilon=10^{-3}$.

For the diffusion SDEs with Improved Euler discretization~\citep{jolicoeur2021gotta}, we follow the results in their original paper, which only includes the results with $\epsilon=10^{-3}$. The corresponding relative tolerance $\epsilon_{rel}$ are $0.50$, $0.10$ and $0.05$, respectively.

For the diffusion ODEs with RK45 Solver, we use the code in~\citep{song2020score}, and tune the \texttt{atol} and \texttt{rtol} of the solver. For the NFE from small to large, we use the same \texttt{atol} = \texttt{rtol} = $0.1$, $0.01$, $0.001$ for the results of $\epsilon=10^{-3}$, and the same \texttt{atol} = \texttt{rtol} = $0.1$, $0.05$, $0.02$, $0.01$, $0.001$ for the results of $\epsilon=10^{-4}$, respectively.

For the diffusion ODEs with DPM-Solver, we use the method in Appendix~\ref{appendix:dpm-solver-fast} for NFE $\leq 20$, and the adaptive step size solver in Appendix~\ref{appendix:algorithm}. For $\epsilon=10^{-3}$, we use DPM-Solver-12 with relative tolerance $\epsilon_{\text{rtol}}=0.05$. For $\epsilon=10^{-4}$, we use DPM-Solver-23 with relative tolerance $\epsilon_{\text{rtol}}=0.05$. 

\subsection{Sample Quality Comparison with RK Methods}
Table~\ref{tab:RK_compare} shows the different performance of RK methods and DPM-Solver-2 and 3. We list the detailed settings in this section.

Assume we have an ODE with
\begin{equation*}
    \frac{\dxv_t}{\dt} = \Fv(\x_t,t),
\end{equation*}
Starting with $\tilde\x_{t_{i-1}}$ at time $t_{i-1}$, we use RK2 to approximate the solution $\tilde\x_{t_i}$ at time $t_i$ in the following formulation (which is known as the explicit midpoint method):
\begin{equation*}
\begin{aligned}
    h_i &= t_{i} - t_{i-1}, \\
    s_i &= t_{i-1} + \frac{1}{2} h_i, \\
    \uv_i &= \tilde\x_{t_{i-1}} + \frac{h_i}{2}\Fv(\tilde\x_{t_{i-1}}, t_{i-1}),\\
    \tilde\x_{t_i} &= \tilde\x_{t_{i-1}} + h_i \Fv(\uv_i, s_i).
\end{aligned}
\end{equation*}
And we use the following RK3 to approximate the solution $\tilde\x_{t_i}$ at time $t_i$ (which is known as ``Heun's third-order method''), because it is very similar to our proposed DPM-Solver-3:
\begin{equation*}
\begin{aligned}
    h_i &= t_{i} - t_{i-1}, \quad r_1 = \frac{1}{3}, \quad r_2 = \frac{2}{3}, \\
    s_{2i-1} &= t_{i-1} + r_1 h_i, \quad s_{2i} = t_{i-1} + r_2 h_i, \\
    \uv_{2i-1} &= \tilde\x_{t_{i-1}} + r_1 h_i\Fv(\tilde\x_{t_{i-1}}, t_{i-1}),\\
    \uv_{2i} &= \tilde\x_{t_{i-1}} + r_2 h_i\Fv(\uv_{2i-1}, s_{2i-1}),\\
    \tilde\x_{t_i} &= \tilde\x_{t_{i-1}} + \frac{h_i}{4} \Fv(\tilde\x_{t_{i-1}}, t_{i-1}) + \frac{3 h_i}{4} \Fv(\uv_{2i}, s_{2i}).
\end{aligned}
\end{equation*}

We use $\Fv(\x_t,t)=\hv_\theta(\x_t,t)$ in Eq.~\eqref{eqn:diffusion_ode} for the results with RK2 ($t$) and RK3 ($t$), and $\Fv(\hat\x_\lambda, \lambda)=\hat\hv_\theta(\hat\x_\lambda,\lambda)$ in Eq.~\eqref{eqn:diffusion_ode_logSNR} for the results with RK2 ($\lambda$) and RK3 ($\lambda$). For all experiments, we use the uniform step size w.r.t. $t$ or $\lambda$.

\subsection{Sample Quality Comparison with Discrete-Time Sampling Methods}
\begin{table}[htbp]
    \centering
    \caption{\label{tab:fid-discrete}\small{Sample quality measured by FID $\downarrow$ on CIFAR-10, CelebA 64$\times$64 and ImageNet 64$\times$64 with discrete-time DPMs, varying the number of function evaluations (NFE). The method $^\dagger$GGDM needs extra training, and some results are missing in their original papers, which are replaced by ``$\backslash$''.}}
    \resizebox{\textwidth}{!}{%
    \begin{tabular}{llrrrrrrr}
    \toprule
      \multicolumn{2}{l}{Sampling method $\backslash$ NFE} & 10 & 12 & 15 & 20 & 50 & 200 & 1000 \\
    \midrule
     \multicolumn{8}{l}{CIFAR-10 (discrete-time model~\citep{ho2020denoising}, linear noise schedule)} \\
     \arrayrulecolor{black!30}\midrule
      \multicolumn{1}{l}{DDPM~\citep{ho2020denoising}} & Discrete & 278.67 & 246.29 & 197.63 & 137.34 & 32.63 & 4.03 & \textbf{3.16} \\
      \multicolumn{1}{l}{Analytic-DDPM~\citep{bao2022analytic}} & Discrete & 35.03 & 27.69 & 20.82 & 15.35 & 7.34 & 4.11 & 3.84 \\
      \multicolumn{1}{l}{Analytic-DDIM~\citep{bao2022analytic}} & Discrete & 14.74 & 11.68 & 9.16 & 7.20 & 4.28 & 3.60 & 3.86 \\
      \multicolumn{1}{l}{$^{\dagger}$GGDM~\citep{watson2021learning}} & Discrete & 8.23 & $\backslash$ & 6.12 & 4.72 & $\backslash$ & $\backslash$ & $\backslash$ \\
      \multicolumn{1}{l}{DDIM~\citep{song2020denoising}} & Discrete & 13.58 & 11.02 & 8.92 & 6.94 & 4.73 & 4.07 & 3.95 \\
      \arrayrulecolor{black!30}\midrule
      \multirow{2}{*}{\shortstack{DPM-Solver (Type-1 discrete)}} & $\epsilon=10^{-3}$ & \textbf{6.37} & \textbf{4.65} & \textbf{3.78} & 4.28 & \multicolumn{3}{c}{3.90 (NFE = 44)}\\
      & $\epsilon=10^{-4}$ & 11.32 & 7.31 & 4.75 & 3.80 & \multicolumn{3}{c}{3.57 (NFE = 46)}  \\
      \arrayrulecolor{black!30}\midrule
      \multirow{2}{*}{DPM-Solver (Type-2 discrete)} & $\epsilon=10^{-3}$ & 6.42 & 4.86 & 4.39 & 5.52 & \multicolumn{3}{c}{5.22 (NFE = 42)}\\
      & $\epsilon=10^{-4}$ & 10.16 & 6.26 & 4.17 & \textbf{3.72} & \multicolumn{3}{c}{\textbf{3.48 (NFE = 44)}}  \\    
      \arrayrulecolor{black}\midrule
     \multicolumn{7}{l}{CelebA 64$\times$64 (discrete-time model~\citep{song2020denoising}, linear noise schedule)}  & \\
     \arrayrulecolor{black!30}\midrule
      DDPM~\citep{ho2020denoising} & Discrete & 310.22 & 277.16 & 207.97 & 120.44 & 29.25 & 3.90 & 3.50 \\
      Analytic-DDPM~\citep{bao2022analytic} & Discrete & 28.99 & 25.27 & 21.80 & 18.14 & 11.23 & 6.51 & 5.21\\
      Analytic-DDIM~\citep{bao2022analytic} & Discrete & 15.62 & 13.90 & 12.29 & 10.45 & 6.13 & 3.46 & 3.13\\
      DDIM~\citep{song2020denoising} & Discrete & 10.85 & 9.99 & 7.78 & 6.64 & 5.23 & 4.78 & 4.88\\
      \arrayrulecolor{black!30}\midrule
      \multirow{2}{*}{DPM-Solver (Type-1 discrete)} & $\epsilon=10^{-3}$ & 7.15 & 5.51 & 4.28 & 4.40 & \multicolumn{3}{c}{4.23 (NFE = 36)}\\
      & $\epsilon=10^{-4}$ & 6.92 & 4.20 & \textbf{3.05} & \textbf{2.82} & \multicolumn{3}{c}{\textbf{2.71 (NFE = 36)}}  \\
      \arrayrulecolor{black!30}\midrule
      \multirow{2}{*}{DPM-Solver (Type-2 discrete)} & $\epsilon=10^{-3}$ & 7.33 & 6.23 & 5.85 & 6.87 & \multicolumn{3}{c}{6.68 (NFE = 36)}\\
      & $\epsilon=10^{-4}$ & \textbf{5.83} & \textbf{3.71} & 3.11 & 3.13 & \multicolumn{3}{c}{3.10 (NFE = 36)}  \\
     \arrayrulecolor{black}\midrule
    \multicolumn{7}{l}{ImageNet 64$\times$64 (discrete-time model~\citep{nichol2021improved}, cosine noise schedule)}  & \\
      \arrayrulecolor{black!30}\midrule
      DDPM~\citep{ho2020denoising} & Discrete & 305.43 & 287.66 & 256.69 & 209.73 & 83.86 & 28.39 & 17.58\\
      Analytic-DDPM~\citep{bao2022analytic} & Discrete & 60.65 & 53.66 & 45.98 & 37.67 & 22.45 & \textbf{17.16} & \textbf{16.14} \\
      Analytic-DDIM~\citep{bao2022analytic} & Discrete & 70.62 & 54.88 & 41.56 & 30.88 & 19.23 & 17.49 & 17.57 \\
      $^\dagger$GGDM~\citep{watson2021learning} & Discrete & 37.32 & $\backslash$ & 24.69 & 20.69 & $\backslash$ & $\backslash$ & $\backslash$ \\ 
      DDIM~\citep{song2020denoising} & Discrete & 67.07 & 52.69 & 40.49 & 30.67 & 20.10 & 17.84 & 17.73 \\
      \arrayrulecolor{black!30}\midrule
      \multirow{2}{*}{DPM-Solver (Type-1 discrete)} & $\epsilon=10^{-3}$ & 24.44 & 20.03 & 19.31 & 18.59 & \multicolumn{3}{c}{17.50 (NFE = 48)}\\
      & $\epsilon=10^{-4}$ & 27.74 & 23.66 & 20.09 & 19.06 & \multicolumn{3}{c}{17.56 (NFE = 51)}  \\
      \arrayrulecolor{black!30}\midrule
      \multirow{2}{*}{DPM-Solver (Type-2 discrete)} & $\epsilon=10^{-3}$ & \textbf{24.40} & \textbf{19.97} & \textbf{19.23} & \textbf{18.53} & \multicolumn{3}{c}{\textbf{17.47 (NFE = 57)}}\\
      & $\epsilon=10^{-4}$ & 27.72 & 23.75 & 20.02 & 19.08 & \multicolumn{3}{c}{17.62 (NFE = 48)} \\
    \arrayrulecolor{black}\bottomrule
    \end{tabular}%
    }
\end{table}

\begin{table}[htbp]
    \centering
    \caption{\label{tab:fid-discrete-highres}\small{Sample quality measured by FID $\downarrow$ on ImageNet 128$\times$128 with classifier guidance and on LSUN bedroom 256$\times$256, varying the number of function evaluations (NFE). For DDIM and DDPM, we use uniform time steps for all the experiments, except that the experiment$^\dagger$ uses the fine-tuned time steps by~\citep{dhariwal2021diffusion}. For DPM-Solver, we use the uniform logSNR steps as described in Appendix~\ref{appendix:dpm-solver-fast}.}}
    \resizebox{\textwidth}{!}{%
    \begin{tabular}{llrrrrrrr}
    \toprule
      \multicolumn{2}{l}{Sampling method $\backslash$ NFE} & 10 & 12 & 15 & 20 & 50 & 100 & 250 \\
    \midrule
     \multicolumn{8}{l}{ImageNet 128$\times$128 (discrete-time model~\citep{dhariwal2021diffusion}, linear noise schedule, classifier guidance scale: 1.25)} \\
     \arrayrulecolor{black!30}\midrule
      \multicolumn{1}{l}{DDPM~\citep{ho2020denoising}} & Discrete & 199.56 & 172.09 & 146.42 & 119.13 & 49.38 & 23.27 & \textbf{2.97} \\
      \multicolumn{1}{l}{DDIM~\citep{song2020denoising}} & Discrete & 11.12 & 9.38 & 8.22 & 7.15 & 5.05 & 4.18 & 3.54 \\
      \arrayrulecolor{black!30}\midrule
      \multirow{2}{*}{\shortstack{DPM-Solver (Type-1 discrete)}} & $\epsilon=10^{-3}$ & \textbf{7.32} & \textbf{4.08} & \textbf{3.60} & 3.89 & 3.63 & 3.62 & 3.63 \\
      & $\epsilon=10^{-4}$ & 13.91 & 5.84 & 4.00 & \textbf{3.52} & \textbf{3.13} & \textbf{3.10} & 3.09  \\
      \arrayrulecolor{black}\midrule
     \multicolumn{7}{l}{LSUN bedroom 256$\times$256 (discrete-time model~\citep{dhariwal2021diffusion}, linear noise schedule)}  & \\
     \arrayrulecolor{black!30}\midrule
      DDPM~\citep{ho2020denoising} & Discrete & 274.67 & 251.26 & 224.88 & 190.14 & 82.70 & 34.89 & $^\dagger$2.02 \\
      DDIM~\citep{song2020denoising} & Discrete & 10.05 & 7.51 & 5.90 & 4.98 & 2.92 & 2.30 & 2.02 \\
      \arrayrulecolor{black!30}\midrule
      \multirow{2}{*}{DPM-Solver (Type-1 discrete)} & $\epsilon=10^{-3}$ & \textbf{6.10} & 4.29 & 3.30 & 3.09 & 2.53 & 2.46 & 2.46 \\
      & $\epsilon=10^{-4}$ & 8.04 & \textbf{4.21} & \textbf{2.94} & \textbf{2.60} & \textbf{2.01} & \textbf{1.95} & \textbf{1.94}  \\
    \arrayrulecolor{black}\bottomrule
    \end{tabular}%
    }
\end{table}

We compare DPM-Solver with other discrete-time sampling methods for DPMs, as shown in Table~\ref{tab:fid-discrete} and Table~\ref{tab:fid-discrete-highres}. We use the code in~\citep{song2020denoising} for sampling with DDPM and DDIM, and the code license is MIT License. We use the code in~\citep{bao2022analytic} for sampling with Analytic-DDPM and Analytic-DDIM, whose 
license is unknown. We directly follow the best results in the original paper of GGDM~\citep{watson2021learning}.

For the CIFAR-10 experiments, we use the pretrained checkpoint by~\citep{ho2020denoising}, which is also provided in the released code in~\citep{song2020denoising}. We use quadratic time steps for DDPM and DDIM, which empirically has better FID performance than the uniform time steps~\citep{song2020denoising}. We use the uniform time steps for Analytic-DDPM and Analytic-DDIM. For DPM-Solver, we use both Type-1 discrete and Type-2 discrete methods to convert the discrete-time model to the continuous-time model. We use the method in Appendix~\ref{appendix:dpm-solver-fast} for NFE $\leq 20$, and the adaptive step size solver in Appendix~\ref{appendix:algorithm} for NFE $> 20$. For all the experiments, we use DPM-Solver-12 with relative tolerance $\epsilon_{\text{rtol}}=0.05$.

For the CelebA 64x64 experiments, we use the pretrained checkpoint by~\citep{song2020denoising}. We use quadratic time steps for DDPM and DDIM, which empirically has better FID performance than the uniform time steps~\citep{song2020denoising}. We use the uniform time steps for Analytic-DDPM and Analytic-DDIM. For DPM-Solver, we use both Type-1 discrete and Type-2 discrete methods to convert the discrete-time model to the continuous-time model. We use the method in Appendix~\ref{appendix:dpm-solver-fast} for NFE $\leq 20$, and the adaptive step size solver in Appendix~\ref{appendix:algorithm} for NFE $> 20$. For all the experiments, we use DPM-Solver-12 with relative tolerance $\epsilon_{\text{rtol}}=0.05$. Note that our best FID results on CelebA 64x64 is even better than the 1000-step DDPM (and all the other methods).

For the ImageNet 64x64 experiments, we use the pretrained checkpoint by~\citep{nichol2021improved}, and the code license is MIT License. We use the uniform time steps for DDPM and DDIM, following~\citep{song2020denoising}. We use the uniform time steps for Analytic-DDPM and Analytic-DDIM. For DPM-Solver, we use both Type-1 discrete and Type-2 discrete methods to convert the discrete-time model to the continuous-time model. We use the method in Appendix~\ref{appendix:dpm-solver-fast} for NFE $\leq 20$, and the adaptive step size solver in Appendix~\ref{appendix:algorithm} for NFE $> 20$. For all the experiments, we use DPM-Solver-23 with relative tolerance $\epsilon_{\text{rtol}}=0.05$. Note that the ImageNet dataset includes real human photos and it may have privacy issues, as discussed in~\citep{yang2021study}.

For the ImageNet 128x128 experiments, we use classifier guidance for sampling with the pretrained checkpoints (for both the diffusion model and the classifier model) by~\citep{dhariwal2021diffusion}, and the code license is MIT License. We use the uniform time steps for DDPM and DDIM, following~\citep{song2020denoising}. For DPM-Solver, we only use Type-1 discrete method to convert the discrete-time model to the continuous-time model. We use the method in Appendix~\ref{appendix:dpm-solver-fast} for NFE $\leq 20$, and the adaptive step size solver DPM-Solver-12 with relative tolerance $\epsilon_{\text{rtol}}=0.05$ (detailed in Appendix~\ref{appendix:algorithm}) for NFE $> 20$. For all the experiments, we set the classifier guidance scale $s=1.25$, which is the best setting for DDIM in~\citep{dhariwal2021diffusion} (we refer to their Table 14 for details).

For the LSUN bedroom 256x256 experiments, we use the unconditional pretrained checkpoint by~\citep{dhariwal2021diffusion}, and the code license is MIT License. We use the uniform time steps for DDPM and DDIM, following~\citep{song2020denoising}. For DPM-Solver, we only use Type-1 discrete method to convert the discrete-time model to the continuous-time model. We use the method in Appendix~\ref{appendix:dpm-solver-fast} for DPM-Solver.

\subsection{Comparing Different Orders of DPM-Solver}
We also compare the sample quality of the different orders of DPM-Solver, as shown in Table~\ref{tab:dpm-solver-123}. We use DPM-Solver-1,2,3 with uniform time steps w.r.t. $\lambda$, and the fast version in Appendix~\ref{appendix:dpm-solver-fast} for NFE less than 20, and we name it as \textit{DPM-Solver-fast}. For the discrete-time models, we only compare the Type-2 discrete method, and the results of Type-1 are similar.

As the actual NFE of DPM-Solver-2 is $2\times\lfloor\text{NFE}/2\rfloor$ and the actual NFE of DPM-Solver-3 is $3\times\lfloor\text{NFE}/3\rfloor$, which may be smaller than NFE, we use the notation $^\dagger$ to note that the actual NFE is less than the given NFE. We find that for NFE less than 20, the proposed fast version (DPM-Solver-fast) is usually better than the single order method, and for larger NFE, DPM-Solver-3 is better than DPM-Solver-2, and DPM-Solver-2 is better than DPM-Solver-1, which matches our proposed convergence rate analysis.

\begin{table}[t]
    \centering
    \caption{Sample quality measured by FID $\downarrow$ of different orders of DPM-Solver, varying the number of function evaluations (NFE). The results with $^\dagger$ means the actual NFE is smaller than the given NFE because the given NFE cannot be divided by $2$ or $3$. For DPM-Solver-fast, we only evaluate it for NFE less than 20, because it is almost the same as DPM-Solver-3 for large NFE.} %
    \label{tab:dpm-solver-123}
    
    \small{
    \begin{tabular}{llrrrrrrr}
    \toprule
      \multicolumn{2}{l}{Sampling method $\backslash$ NFE} & 10 & 12 & 15 & 20 & 50 & 200 & 1000 \\
    \midrule
     \multicolumn{8}{l}{CIFAR-10 (VP deep continuous-time model~\citep{song2020score})}  & \\
     \arrayrulecolor{black!30}\midrule
      \multirow{4}{*}{$\epsilon=10^{-3}$} & DPM-Solver-1 & 11.83 & 9.69 & 7.78 & 6.17 & 4.28 & 3.85 & 3.83  \\
      & DPM-Solver-2 & 5.94 & 4.88 & $^\dagger$4.30  & 3.94 & 3.78 & 3.74 & 3.74\\
      & DPM-Solver-3 & $^\dagger$18.37 & 5.53 & 4.08 & $^\dagger$4.04 & $^\dagger$3.81 & $^\dagger$3.78 & $^\dagger$3.78 \\
      & DPM-Solver-fast & \textbf{4.70} & \textbf{3.75} & \textbf{3.24} & 3.99 & $\backslash$ & $\backslash$ & $\backslash$\\
      \arrayrulecolor{black!30}\midrule
      \multirow{4}{*}{$\epsilon=10^{-4}$} & DPM-Solver-1 & 11.29 & 9.07 & 7.15 & 5.50 & 3.32 & 2.72 & 2.64 \\
      & DPM-Solver-2 & 7.30 & 5.28 & $^\dagger$4.23 & 3.26 & 2.69 & \textbf{2.60} & \textbf{2.59}\\
      & DPM-Solver-3 & $^\dagger$54.56 & 6.03 & 3.55 & $^\dagger$2.90 & $^{\dagger}$\textbf{2.65} & $^\dagger$2.62 & $^\dagger$2.62 \\
      & DPM-Solver-fast & 6.96 & 4.93 & 3.35 & \textbf{2.87} & $\backslash$ & $\backslash$ & $\backslash$\\
     \arrayrulecolor{black}\midrule
     \multicolumn{8}{l}{CIFAR-10 (DDPM discrete-time model~\citep{ho2020denoising}), DPM-Solver with Type-2 discrete}  & \\
     \arrayrulecolor{black!30}\midrule
      \multirow{4}{*}{$\epsilon=10^{-3}$} & DPM-Solver-1 & 16.69 & 13.63 & 11.08 & 8.90 & 6.24 & 5.44 & 5.29 \\
      & DPM-Solver-2 & 7.90 & 6.15 & $^\dagger$5.53 & 5.24 & 5.23 & 5.25 & 5.25\\
      & DPM-Solver-3 & $^\dagger$24.37 & 8.20 & 5.73 & $^\dagger$5.43 & $^\dagger$5.29 & $^\dagger$5.25 & $^\dagger$5.25\\
      & DPM-Solver-fast & \textbf{6.42} & \textbf{4.86} & 4.39 & 5.52 & $\backslash$ & $\backslash$ & $\backslash$\\
      \arrayrulecolor{black!30}\midrule
      \multirow{4}{*}{$\epsilon=10^{-4}$} & DPM-Solver-1 & 13.61 & 10.98 & 8.71 & 6.79 & 4.36 & 3.63 & 3.49\\
      & DPM-Solver-2 & 11.80 & 6.31 & $^\dagger$5.23 & 3.95 & 3.50 & 3.46 & 3.46\\
      & DPM-Solver-3 & $^\dagger$67.02 & 9.45 & 5.21 & $^\dagger$3.81 & $^\dagger$\textbf{3.49} & $^\dagger$\textbf{3.45} & $^\dagger$\textbf{3.45}\\
      & DPM-Solver-fast & 10.16 & 6.26 & \textbf{4.17} & \textbf{3.72} & $\backslash$ & $\backslash$ & $\backslash$\\
     \arrayrulecolor{black}\midrule
     \multicolumn{9}{l}{CelebA 64$\times$64 (discrete-time model~\citep{song2020denoising}, linear noise schedule), DPM-Solver with Type-2 discrete}\\
     \arrayrulecolor{black!30}\midrule
      \multirow{4}{*}{$\epsilon=10^{-3}$} & DPM-Solver-1 & 18.66 & 16.30 & 13.92 & 11.84 & 8.85 & 7.24 & 6.93 \\
      & DPM-Solver-2 & 5.89 & 5.83 & $^\dagger$6.08 & 6.38 & 6.78 & 6.84 & 6.85 \\
      & DPM-Solver-3 & $^\dagger$11.45 & 5.46 & 6.18 & $^\dagger$6.51 & $^\dagger$6.87 & $^\dagger$6.84 & $^\dagger$6.85 \\
      & DPM-Solver-fast & 7.33 & 6.23 & 5.85 & 6.87 & $\backslash$ & $\backslash$ & $\backslash$ \\
      \arrayrulecolor{black!30}\midrule
      \multirow{4}{*}{$\epsilon=10^{-4}$} & DPM-Solver-1 & 13.24 & 11.13 & 9.08 & 7.24 & 4.50 & 3.48 & 3.25 \\
      & DPM-Solver-2 & \textbf{4.28} & \textbf{3.40} & $^\dagger$3.30 & 3.17 & \textbf{3.19} & \textbf{3.20} & \textbf{3.20}\\
      & DPM-Solver-3 & $^\dagger$49.48 & 3.84 & \textbf{3.09} & $^\dagger$3.15 & $^\dagger$3.20 & $^\dagger$\textbf{3.20} & $^\dagger$\textbf{3.20}\\
      & DPM-Solver-fast & 5.83 & 3.71 & 3.11 & \textbf{3.13} & $\backslash$ & $\backslash$ & $\backslash$ \\
     \arrayrulecolor{black}\midrule
    \multicolumn{9}{l}{ImageNet 64$\times$64 (discrete-time model~\citep{nichol2021improved}, cosine noise schedule), DPM-Solver with Type-2 discrete} \\
     \arrayrulecolor{black!30}\midrule
      \multirow{4}{*}{$\epsilon=10^{-3}$} & DPM-Solver-1 & 32.84 & 28.54 & 24.79 & 21.71 & 18.30 & 17.45 & 17.18 \\
      & DPM-Solver-2 & 29.20 & 24.97 & $^\dagger$22.26 & 19.94 & 17.79 & 17.29 & \textbf{17.27} \\
      & DPM-Solver-3 & $^\dagger$57.48 & 24.62 & 19.76 & $^\dagger$18.95 & $^\dagger$\textbf{17.52} & \textbf{17.26} & \textbf{17.27} \\
      & DPM-Solver-fast & \textbf{24.40} & \textbf{19.97} & \textbf{19.23} & \textbf{18.53} & $\backslash$ & $\backslash$ & $\backslash$ \\
      \arrayrulecolor{black!30}\midrule
      \multirow{4}{*}{$\epsilon=10^{-4}$} & DPM-Solver-1 & 32.31 & 28.44 & 25.15 & 22.38 & 19.14 & 17.95 & 17.44 \\
      & DPM-Solver-2 & 33.16 & 27.28 & $^\dagger$24.26 & 20.58 & 18.04 & 17.46 & 17.41 \\
      & DPM-Solver-3 & $^\dagger$162.27 & 27.28 & 22.38 & $^\dagger$19.39 & $^\dagger$17.71 & $^\dagger$17.43 & $^\dagger$17.41\\
      & DPM-Solver-fast & 27.72 & 23.75 & 20.02 & 19.08 & $\backslash$ & $\backslash$ & $\backslash$ \\
    \arrayrulecolor{black}\bottomrule
    \end{tabular}
    }
\end{table}

\subsection{Runtime Comparison between DPM-Solver and DDIM}
Theoretically, for the same NFE, the runtime of DPM-Solver and DDIM are almost the same (linear to NFE) because the main computation costs are the serial evaluations of the large neural network $\epsilonv_\theta$ and the other coefficients are \textbf{analytically} computed with ignorable costs.

Table~\ref{tab:runtime} shows the runtime of DPM-Solver and DDIM on a single NVIDIA A40, varying different datasets and NFE. We use \texttt{torch.cuda.Event} and \texttt{torch.cuda.synchronize} for accurately computing the runtime. We use the discrete-time pretrained diffusion models for each dataset. We evaluate the runtime for 8 batches and computes the mean and std of the runtime. We use 64 batch size for LSUN bedroom 256x256 due to the GPU memory limitation, and 128 batch size for other datasets.

For DDIM, we use the official implementation\footnote{\url{https://github.com/ermongroup/ddim}}. We find that our implementation of DPM-Solver reduces some repetitive computation of the coefficients, so under the same NFE, DPM-Solver is slightly faster than DDIM of their implementation. Nevertheless, the runtime evaluation results show that the runtime of DPM-Solver and DDIM are almost the same for the same NFE, and the runtime is approximately linear to the NFE. Therefore, the speedup for the NFE is almost the actual speedup of the runtime, so the proposed DPM-Solver can greatly speedup the sampling of DPMs.

\begin{table}[t]
    \centering
    \caption{\small{Runtime of a single batch (second / batch, $\pm$std) on a single NVIDIA A40 of DDIM and DPM-Solver for sampling by discrete-time diffusion models, varying the number of function evaluations (NFE).}}
    \label{tab:runtime}
    \resizebox{\textwidth}{!}{%
    \begin{tabular}{lrrrr}
    \toprule
      Sampling method $\backslash$ NFE & 10 & 20 & 50 & 100 \\
    \midrule
     \multicolumn{4}{l}{CIFAR-10 32$\times$32 (batch size = 128)} \\
     \arrayrulecolor{black!30}\midrule
      DDIM & 0.956($\pm$0.011) & 1.924($\pm$0.016) & 4.838($\pm$0.024) & 9.668($\pm$0.013) \\
      DPM-Solver & 0.923($\pm$0.006) & 1.833($\pm$0.004) & 4.580($\pm$0.005) & 9.204($\pm$0.011) \\
      \arrayrulecolor{black}\midrule
      \multicolumn{4}{l}{CelebA 64$\times$64 (batch size = 128)} \\
     \arrayrulecolor{black!30}\midrule
      DDIM & 3.253($\pm$0.015) & 6.438($\pm$0.029) & 16.132($\pm$0.050) & 32.255($\pm$0.044)\\
      DPM-Solver & 3.126($\pm$0.003) & 6.272($\pm$0.006) & 15.676($\pm$0.008) & 31.269($\pm$0.012)  \\
      \arrayrulecolor{black}\midrule
     \multicolumn{4}{l}{ImageNet 64$\times$64 (batch size = 128)} \\
     \arrayrulecolor{black!30}\midrule
      DDIM & 5.084($\pm$0.018) & 10.194($\pm$0.022) & 25.440($\pm$0.044) & 50.926($\pm$0.042) \\
      DPM-Solver & 4.992($\pm$0.004) & 9.991($\pm$0.003) & 24.948($\pm$0.007) & 49.835($\pm$0.028) \\
      \arrayrulecolor{black}\midrule
     \multicolumn{4}{l}{ImageNet 128$\times$128 (batch size = 128, with classifier guidance)} \\
     \arrayrulecolor{black!30}\midrule
      DDIM & 29.082($\pm$0.015) & 58.159($\pm$0.012) & 145.427($\pm$0.011) & 290.874($\pm$0.134) \\
      DPM-Solver & 28.865($\pm$0.011) & 57.645($\pm$0.008) & 144.124($\pm$0.035) & 288.157($\pm$0.022) \\
      \arrayrulecolor{black}\midrule
     \multicolumn{4}{l}{LSUN bedroom 256$\times$256 (batch size = 64)} \\
     \arrayrulecolor{black!30}\midrule
      DDIM & 37.700($\pm$0.005) & 75.316($\pm$0.013) & 188.275($\pm$0.172) & 378.790($\pm$0.105)\\
      DPM-Solver & 36.996($\pm$0.039) & 73.873($\pm$0.023) & 184.590($\pm$0.010) & 369.090($\pm$0.076) \\
    \arrayrulecolor{black}\bottomrule
    \end{tabular}%
    }
\end{table}

\subsection{Conditional Sampling on ImageNet 256x256}
For the conditional sampling in Fig.~\ref{fig:intro}, we use the pretrained checkpoint in~\citep{dhariwal2021diffusion} with classifier guidance (ADM-G), and the classifier scale is $1.0$. The code license is MIT License. We use uniform time step for DDIM, and the fast version for DPM-Solver in Appendix~\ref{appendix:dpm-solver-fast} (DPM-Solver-fast) with 10, 15, 20 and 100 steps.

\begin{figure}[t]
	\hspace{0.16\linewidth}
	\begin{minipage}{0.20\linewidth}
	\centering
	NFE = 10
	\end{minipage}
	\begin{minipage}{0.19\linewidth}
	\centering
	NFE = 15
	\end{minipage}
	\begin{minipage}{0.19\linewidth}
	\centering
	NFE = 20
	\end{minipage}
	\begin{minipage}{0.20\linewidth}
	\centering
	NFE = 100
	\end{minipage}
	\vspace{0.2cm}
	\\
	\centering
	\begin{minipage}{0.15\linewidth}
	\centering
    DDIM \\
    \centering
    \citep{song2020denoising}
    \\
    \vspace{2.2cm}
	\centering
    DPM-Solver \\ 
	\centering
    (ours)
	\end{minipage}
	\begin{minipage}{0.8\linewidth}
		\centering
			\includegraphics[width=\linewidth]{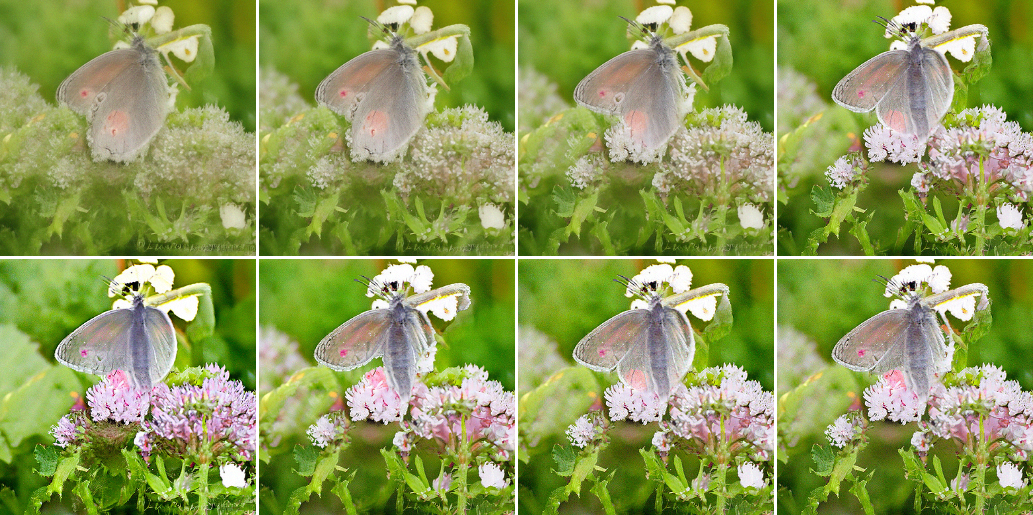}\\
	\end{minipage}
	\vspace{0.2cm}
	\\
	\centering
	\begin{minipage}{0.15\linewidth}
	\centering
    DDIM \\
    \centering
    \citep{song2020denoising}
    \\
    \vspace{2.2cm}
	\centering
    DPM-Solver \\ 
	\centering
    (ours)
	\end{minipage}
	\begin{minipage}{0.8\linewidth}
		\centering
			\includegraphics[width=\linewidth]{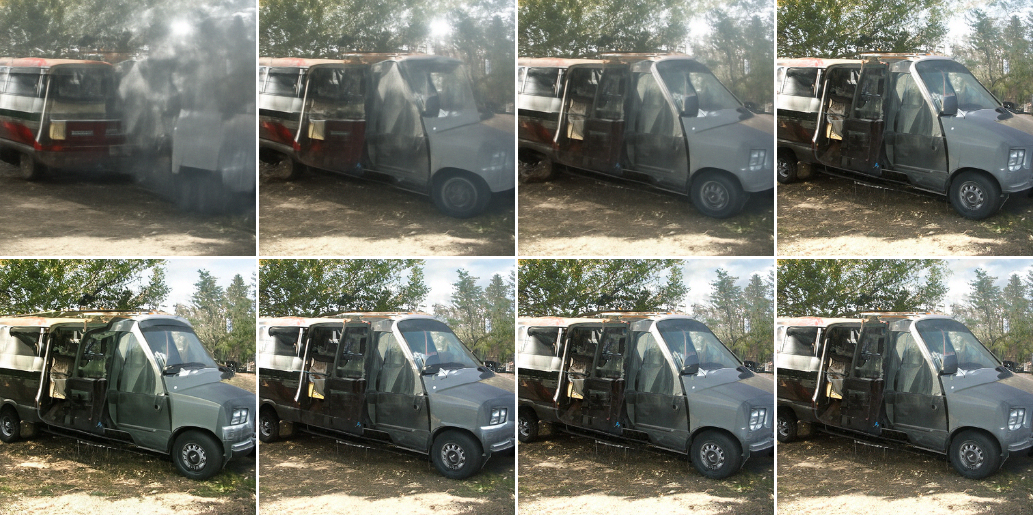}\\
	\end{minipage}
	\caption{\label{fig:appendix-conditional}
Samples by DDIM~\citep{song2020denoising} and DPM-Solver (ours) with 10, 15, 20, 100 number of function evaluations (NFE) with the same random seed, using the pre-trained DPMs on ImageNet 256$\times$256 with classifier guidance~\citep{dhariwal2021diffusion}.
}
\end{figure}

Fig.~\ref{fig:appendix-conditional} shows the conditional sample results by DDIM and DPM-Solver. We find that DPM-Solver with 15 NFE can generate comparable samples with DDIM with 100 NFE.

\subsection{Additional Samples}
\begin{figure}[t]
	\hspace{0.1\linewidth}
	\begin{minipage}{0.21\linewidth}
	\centering
	NFE = 10
	\end{minipage}
	\begin{minipage}{0.22\linewidth}
	\centering
	NFE = 12
	\end{minipage}
	\begin{minipage}{0.22\linewidth}
	\centering
	NFE = 15
	\end{minipage}
	\begin{minipage}{0.22\linewidth}
	\centering
	NFE = 20
	\end{minipage}
	\vspace{0.2cm}
	\\
	\centering
	\begin{minipage}{0.1\linewidth}
	\centering
    \small{DDIM}\\
    \centering
    \small{\citep{song2020denoising}}
    \\
    \vspace{2.5cm}
	\centering
    \small{DPM-Solver} \\ 
	\centering
    \small{(ours)}
	\end{minipage}
	\begin{minipage}{0.89\linewidth}
		\centering
			\includegraphics[width=0.24\linewidth]{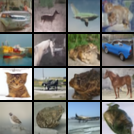}
			\includegraphics[width=0.24\linewidth]{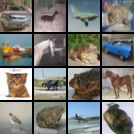}
			\includegraphics[width=0.24\linewidth]{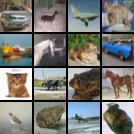}
			\includegraphics[width=0.24\linewidth]{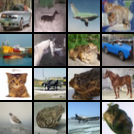} \\
            \vspace{0.5cm}
			\includegraphics[width=0.24\linewidth]{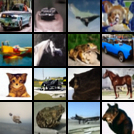}
			\includegraphics[width=0.24\linewidth]{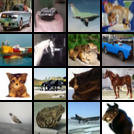}
			\includegraphics[width=0.24\linewidth]{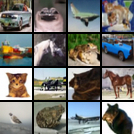}
			\includegraphics[width=0.24\linewidth]{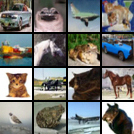} \\
	\end{minipage}
	\vspace{0.2cm}
	\caption{\label{fig:appendix-cifar10-discrete}
Random samples by DDIM~\citep{song2020denoising} (quadratic time steps) and DPM-Solver (ours) with 10, 12, 15, 20 number of function evaluations (NFE) with the same random seed, using the pre-trained discrete-time DPMs~\citep{ho2020denoising} on CIFAR-10.
}
\end{figure}

\begin{figure}[t]
	\hspace{0.1\linewidth}
	\begin{minipage}{0.21\linewidth}
	\centering
	NFE = 10
	\end{minipage}
	\begin{minipage}{0.22\linewidth}
	\centering
	NFE = 12
	\end{minipage}
	\begin{minipage}{0.22\linewidth}
	\centering
	NFE = 15
	\end{minipage}
	\begin{minipage}{0.22\linewidth}
	\centering
	NFE = 20
	\end{minipage}
	\vspace{0.2cm}
	\\
	\centering
	\begin{minipage}{0.1\linewidth}
	\centering
    \small{DDIM}\\
    \centering
    \small{\citep{song2020denoising}}
    \\
    \vspace{2.cm}
	\centering
    \small{DPM-Solver} \\ 
	\centering
    \small{(ours)}
	\end{minipage}
	\begin{minipage}{0.89\linewidth}
		\centering
			\includegraphics[width=0.24\linewidth]{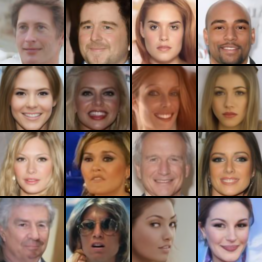}
			\includegraphics[width=0.24\linewidth]{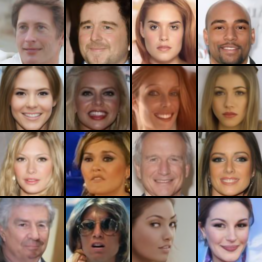}
			\includegraphics[width=0.24\linewidth]{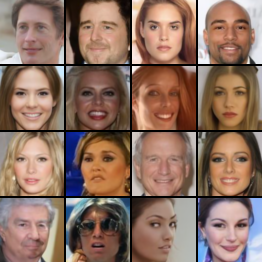}
			\includegraphics[width=0.24\linewidth]{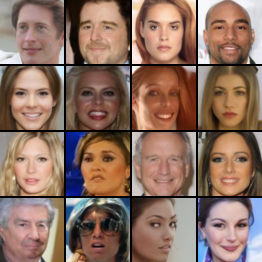} \\
            \vspace{0.5cm}
			\includegraphics[width=0.24\linewidth]{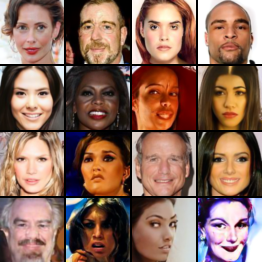}
			\includegraphics[width=0.24\linewidth]{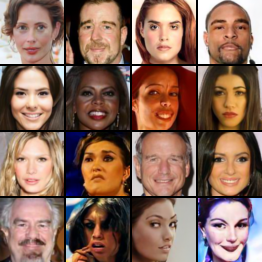}
			\includegraphics[width=0.24\linewidth]{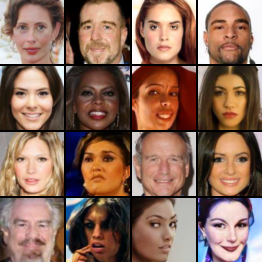}
			\includegraphics[width=0.24\linewidth]{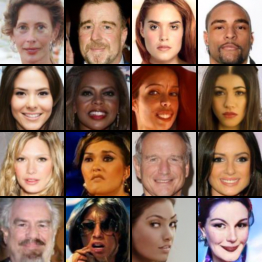} \\
	\end{minipage}
	\vspace{0.2cm}
	\caption{\label{fig:appendix-celeba-discrete}
Random samples by DDIM~\citep{song2020denoising} (quadratic time steps) and DPM-Solver (ours) with 10, 12, 15, 20 number of function evaluations (NFE) with the same random seed, using the pre-trained discrete-time DPMs~\citep{song2020denoising} on CelebA 64x64.
}
\end{figure}

\begin{figure}[t]
	\hspace{0.1\linewidth}
	\begin{minipage}{0.21\linewidth}
	\centering
	NFE = 10
	\end{minipage}
	\begin{minipage}{0.22\linewidth}
	\centering
	NFE = 12
	\end{minipage}
	\begin{minipage}{0.22\linewidth}
	\centering
	NFE = 15
	\end{minipage}
	\begin{minipage}{0.22\linewidth}
	\centering
	NFE = 20
	\end{minipage}
	\vspace{0.2cm}
	\\
	\centering
	\begin{minipage}{0.1\linewidth}
	\centering
    \small{DDIM}\\
    \centering
    \small{\citep{song2020denoising}}
    \\
    \vspace{2.cm}
	\centering
    \small{DPM-Solver} \\ 
	\centering
    \small{(ours)}
	\end{minipage}
	\begin{minipage}{0.89\linewidth}
		\centering
			\includegraphics[width=0.24\linewidth]{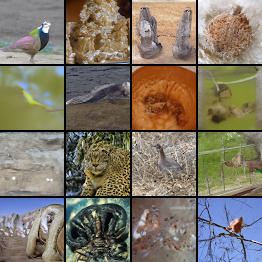}
			\includegraphics[width=0.24\linewidth]{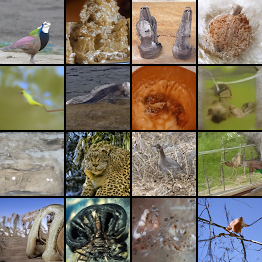}
			\includegraphics[width=0.24\linewidth]{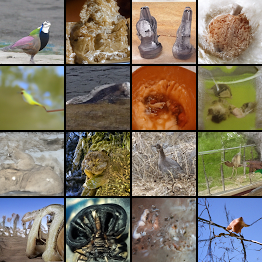}
			\includegraphics[width=0.24\linewidth]{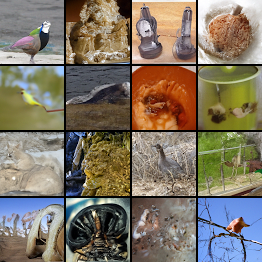} \\
            \vspace{0.5cm}
			\includegraphics[width=0.24\linewidth]{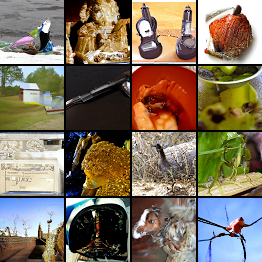}
			\includegraphics[width=0.24\linewidth]{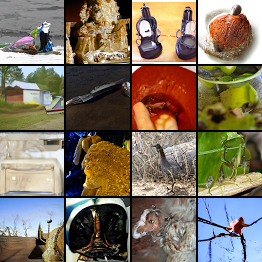}
			\includegraphics[width=0.24\linewidth]{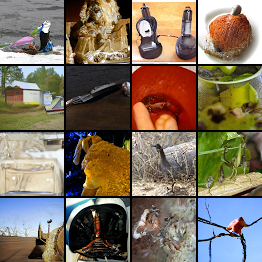}
			\includegraphics[width=0.24\linewidth]{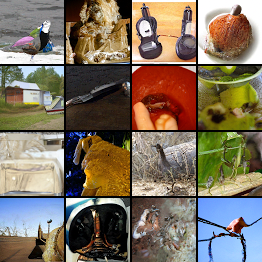} \\
	\end{minipage}
	\vspace{0.2cm}
	\caption{\label{fig:appendix-imagenet64-discrete}
Random samples by DDIM~\citep{song2020denoising} (uniform time steps) and DPM-Solver (ours) with 10, 12, 15, 20 number of function evaluations (NFE) with the same random seed, using the pre-trained discrete-time DPMs~\citep{nichol2021improved} on ImageNet 64x64.
}
\end{figure}

\begin{figure}[t]
	\hspace{0.1\linewidth}
	\begin{minipage}{0.21\linewidth}
	\centering
	NFE = 10
	\end{minipage}
	\begin{minipage}{0.22\linewidth}
	\centering
	NFE = 12
	\end{minipage}
	\begin{minipage}{0.22\linewidth}
	\centering
	NFE = 15
	\end{minipage}
	\begin{minipage}{0.22\linewidth}
	\centering
	NFE = 20
	\end{minipage}
	\vspace{0.2cm}
	\\
	\centering
	\begin{minipage}{0.1\linewidth}
	\centering
    \small{DDIM}\\
    \centering
    \small{\citep{song2020denoising}}
    \\
    \vspace{2.cm}
	\centering
    \small{DPM-Solver} \\ 
	\centering
    \small{(ours)}
	\end{minipage}
	\begin{minipage}{0.89\linewidth}
		\centering
			\includegraphics[width=0.24\linewidth]{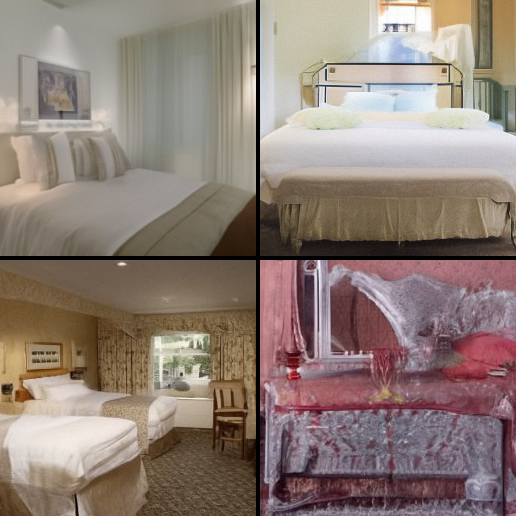}
			\includegraphics[width=0.24\linewidth]{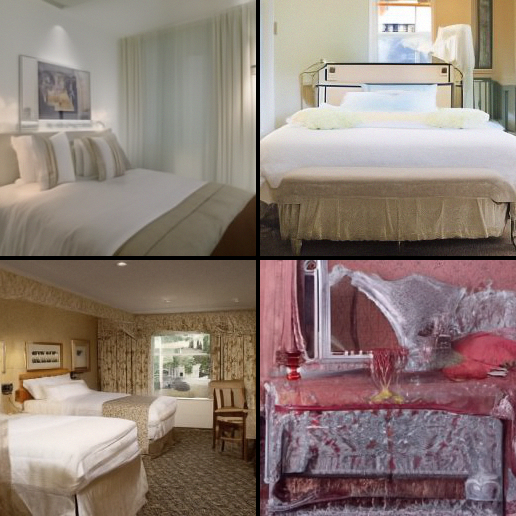}
			\includegraphics[width=0.24\linewidth]{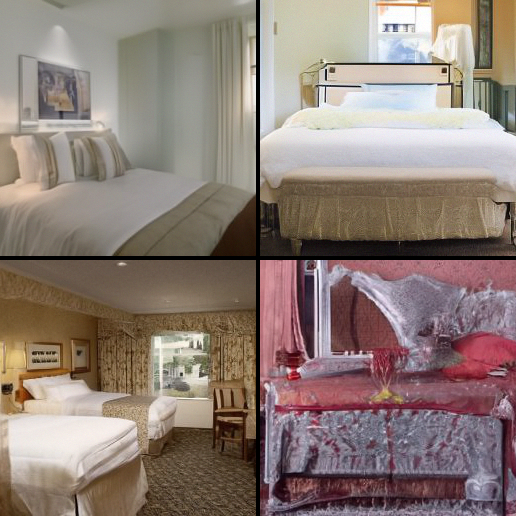}
			\includegraphics[width=0.24\linewidth]{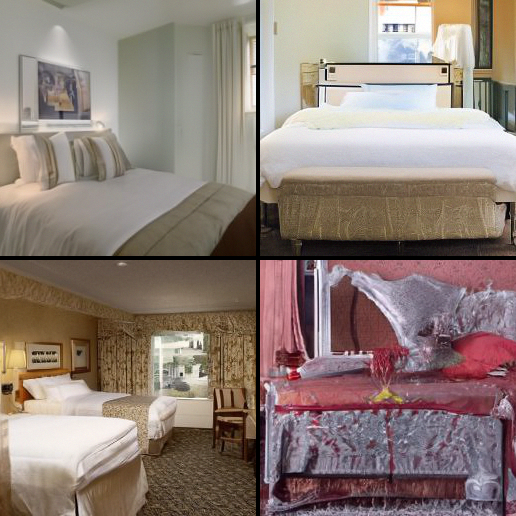} \\
            \vspace{0.5cm}
			\includegraphics[width=0.24\linewidth]{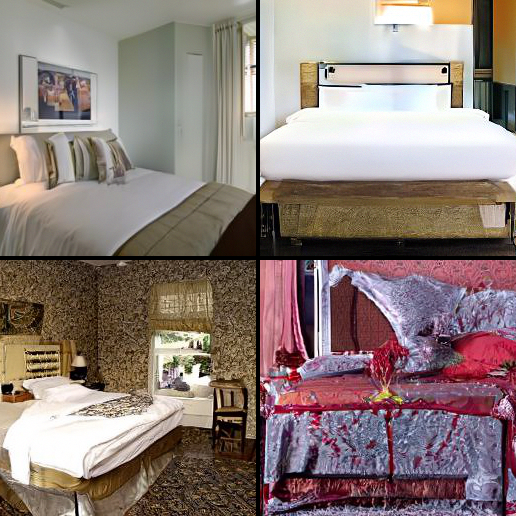}
			\includegraphics[width=0.24\linewidth]{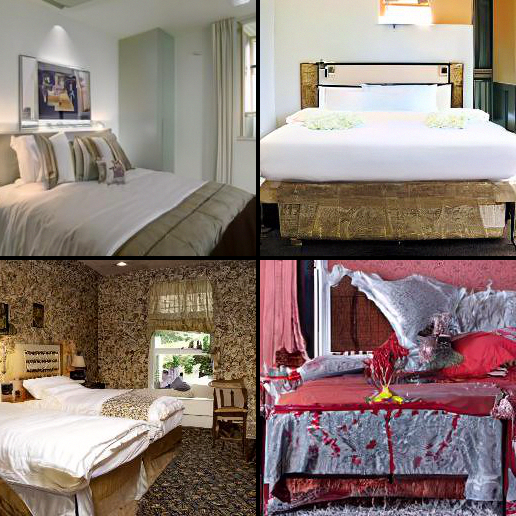}
			\includegraphics[width=0.24\linewidth]{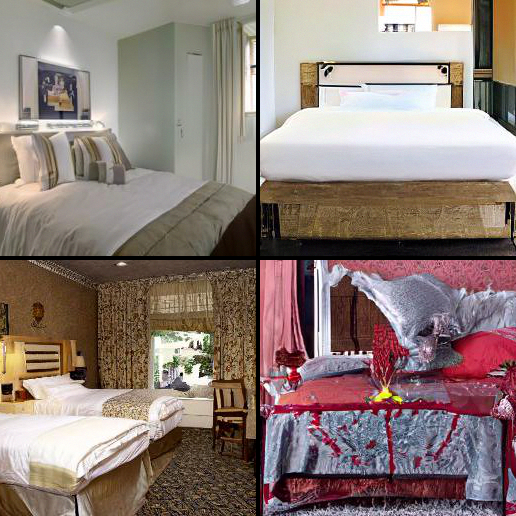}
			\includegraphics[width=0.24\linewidth]{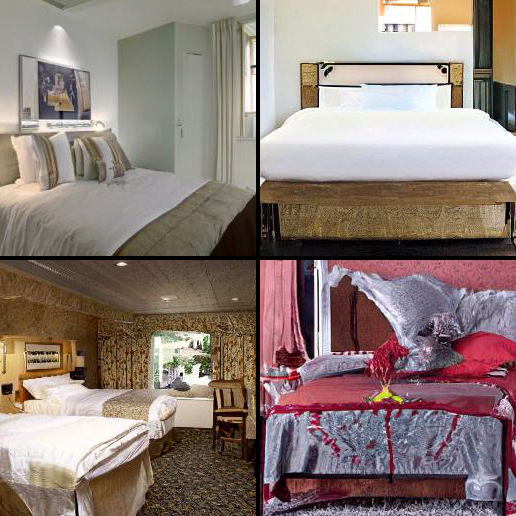} \\
	\end{minipage}
	\vspace{0.2cm}
	\caption{\label{fig:appendix-bedroom256-discrete}
Random samples by DDIM~\citep{song2020denoising} (uniform time steps) and DPM-Solver (ours) with 10, 12, 15, 20 number of function evaluations (NFE) with the same random seed, using the pre-trained discrete-time DPMs~\citep{dhariwal2021diffusion} on LSUN bedroom 256x256.
}
\end{figure}

\begin{figure}[t]
	\hspace{0.1\linewidth}
	\begin{minipage}{0.21\linewidth}
	\centering
	NFE = 10
	\end{minipage}
	\begin{minipage}{0.22\linewidth}
	\centering
	NFE = 12
	\end{minipage}
	\begin{minipage}{0.22\linewidth}
	\centering
	NFE = 15
	\end{minipage}
	\begin{minipage}{0.22\linewidth}
	\centering
	NFE = 20
	\end{minipage}
	\vspace{0.2cm}
	\\
	\centering
	\begin{minipage}{0.1\linewidth}
	\centering
    \small{DDIM}\\
    \centering
    \small{\citep{song2020denoising}}
    \\
    \vspace{2.cm}
	\centering
    \small{DPM-Solver} \\ 
	\centering
    \small{(ours)}
	\end{minipage}
	\begin{minipage}{0.89\linewidth}
		\centering
			\includegraphics[width=0.24\linewidth]{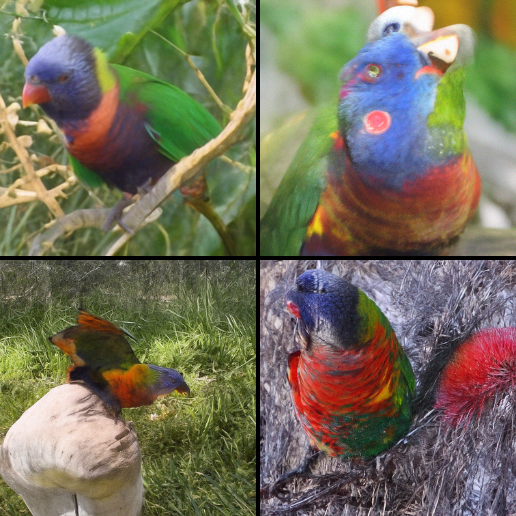}
			\includegraphics[width=0.24\linewidth]{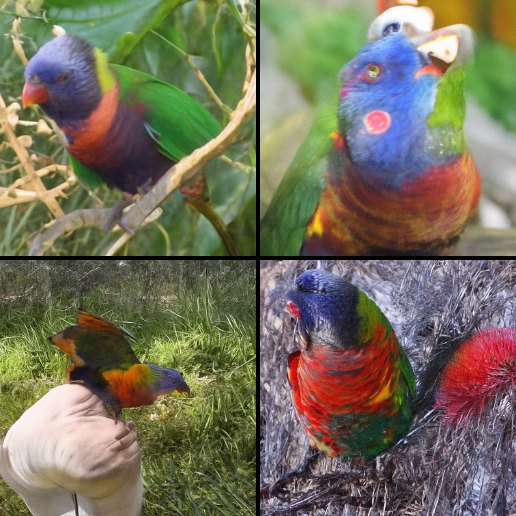}
			\includegraphics[width=0.24\linewidth]{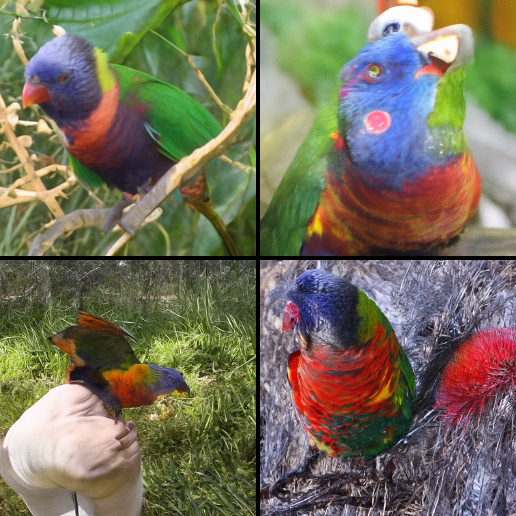}
			\includegraphics[width=0.24\linewidth]{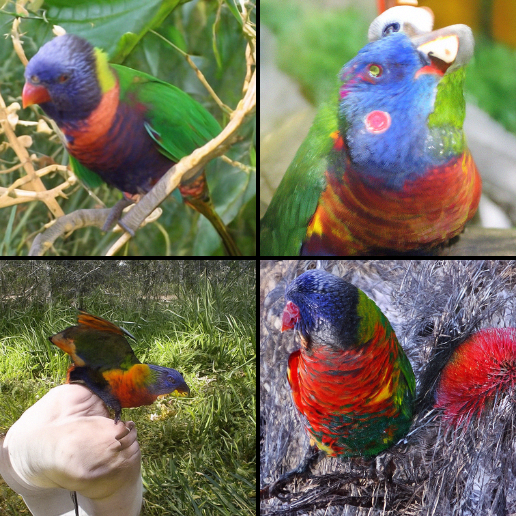} \\
            \vspace{0.5cm}
			\includegraphics[width=0.24\linewidth]{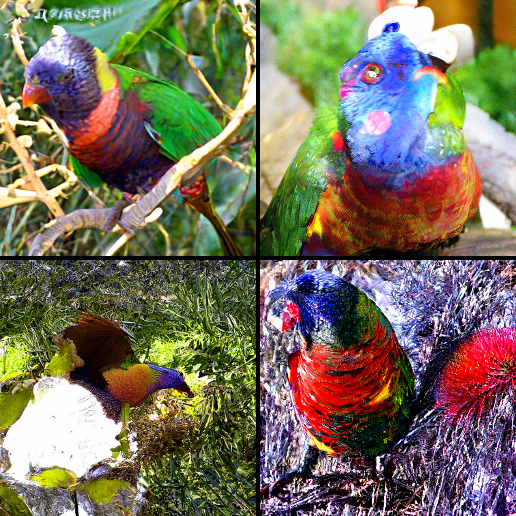}
			\includegraphics[width=0.24\linewidth]{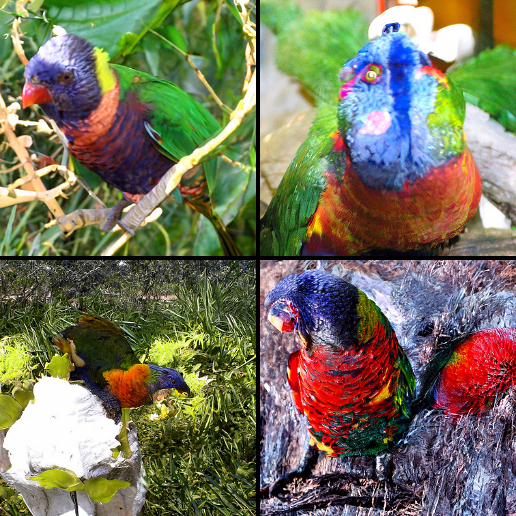}
			\includegraphics[width=0.24\linewidth]{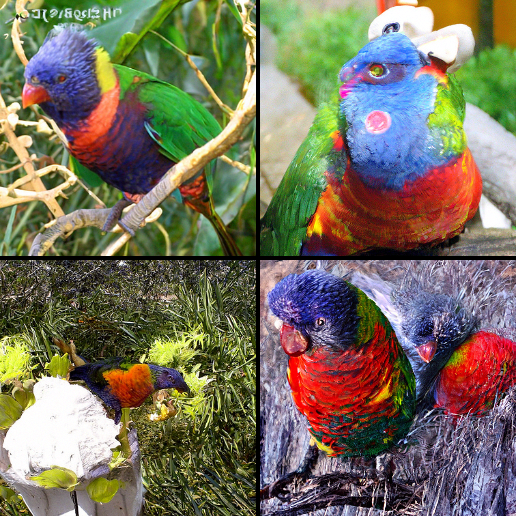}
			\includegraphics[width=0.24\linewidth]{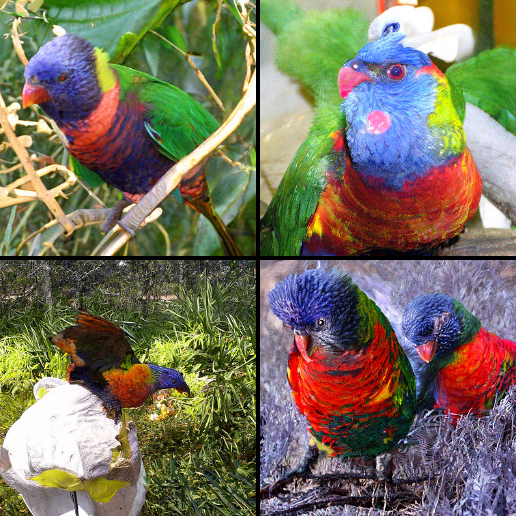} \\
	\end{minipage}
	\vspace{0.2cm}
	\caption{\label{fig:appendix-imagenet256-discrete}
Random class-conditional samples (class: 90, lorikeet) by DDIM~\citep{song2020denoising} (uniform time steps) and DPM-Solver (ours) with 10, 12, 15, 20 number of function evaluations (NFE) with the same random seed, using the pre-trained discrete-time DPMs~\citep{dhariwal2021diffusion} on ImageNet 256x256 with classifier guidance (classifier scale: 1.0).
}
\end{figure}

Additional sampling results on CIFAR-10, CelebA 64x64, ImageNet 64x64, LSUN bedroom 256x256~\citep{yu2015lsun}, ImageNet 256x256 are reported in
Figs.~\ref{fig:appendix-cifar10-discrete}-\ref{fig:appendix-imagenet256-discrete}.

\end{document}